\documentclass[10pt,twocolumn,letterpaper]{article}

\usepackage[pagenumbers]{main} 

\usepackage{graphicx}
\usepackage{amsmath}
\usepackage{amssymb}
\usepackage{booktabs}

\usepackage{amsmath,epsfig}
\usepackage[ruled, vlined, linesnumbered]{algorithm2e}%
\usepackage{enumitem,amsthm,amsmath,amsfonts,amssymb}
\usepackage{color}
\usepackage{xcolor}
\usepackage{multirow}
\usepackage{multicol}
\usepackage{makecell}
\usepackage{subcaption}
\usepackage{balance}
\usepackage{rotating}
\newcommand{\rulesep}{\unskip\ \vrule\ }

\def\sim{\texttt{sim}}
\def\thr{\texttt{Thr}}

\usepackage[colorlinks]{hyperref}
\usepackage[capitalize]{cleveref}
\crefname{section}{Sec.}{Secs.}
\Crefname{section}{Section}{Sections}
\Crefname{table}{Table}{Tables}
\crefname{table}{Tab.}{Tabs.}

\def\cvprPaperID{*****} 
\def\confName{CVPR}
\def\confYear{2022}

\begin{document}

\title{PseudoProp: Robust Pseudo-Label Generation for Semi-Supervised Object Detection in Autonomous Driving Systems}


\author{Shu Hu$^{1}$\thanks{Work done while interning at Bosch Center for Artificial Intelligence}, \  Chun-Hao Liu$^{2}$\thanks{Corresponding author}, \  Jayanta Dutta$^{2}$, \   Ming-Ching Chang$^{3}$, \  Siwei Lyu$^1$, \  Naveen Ramakrishnan$^4$ \\
$^{1}$University at Buffalo, SUNY, USA \ \ \  $^{2}$Bosch Center for Artificial Intelligence, USA \\ $^{3}$University at Albany, SUNY, USA \ \ \ $^{4}$Amazon, USA\\
{\tt\small \{shuhu, siweilyu\}@buffalo.edu}, \ \ \  {\tt\small \{Chun-Hao.Liu, Jayanta.Dutta\}@us.bosch.com} \\ {\tt\small mchang2@albany.edu}, \ \ \
{\tt\small rnaveen83@gmail.com}
}
\maketitle

\begin{abstract}
   Semi-supervised object detection methods are widely used in autonomous driving systems, where only a fraction of objects are labeled. To propagate information from the labeled objects to the unlabeled ones, pseudo-labels for unlabeled objects must be generated. Although pseudo-labels have proven to improve the performance of semi-supervised object detection significantly, the applications of image-based methods to video frames result in numerous miss or false detections using such generated pseudo-labels. In this paper, we propose a new approach, \textit{PseudoProp}, to generate robust pseudo-labels by leveraging motion continuity in video frames. Specifically, PseudoProp uses a novel bidirectional pseudo-label propagation approach to compensate for misdetection. A feature-based fusion technique is also used to suppress inference noise. 
Extensive experiments on the large-scale Cityscapes dataset demonstrate that our method outperforms the state-of-the-art semi-supervised object detection methods by 7.4\% on mAP$^{75}$.
\end{abstract}

\section{Introduction}
In autonomous driving system design and development, it is common to collect multiple video sequences and only label key frames to train a deep neural network (DNN) based object detector. However, the efficacy of the detector may be limited  by the size of the human-annotated dataset. Therefore, we have witnessed the \textit{tour de force} of modern DNNs with semi-supervised learning (SSL) in the past that have been applied to autonomous driving \cite{misra2015watch, chen2020naive}.
SSL uses available human-annotated data to guide the model training with unlabeled data. One dominant idea in SSL is pseudo-labeling, where pseudo-labels of unlabeled data are repeatedly generated by a pre-trained model. The model is then updated by training on a mixture of pseudo-labels and human-annotated data. Since the pre-trained model can generate highly confident pseudo-labels, SSL-based models can improve the performance of both image~\cite{sohn2020simple,  tang2021humble} and video~\cite{chen2020naive} object detection. Notwithstanding this tremendous success, pseudo-labels generated by conventional SSL-based object detection models from unlabeled data are not all reliable. Hence they cannot be directly applied to the training procedure of the detector network to improve performance~\cite{arazo2020pseudo, berthelot2019remixmatch}. 
In particular, misdetections and false detections can easily appear in the pseudo-labels, due to the performance bottleneck of the selected pre-trained object detector. 
In this paper, we propose to leverage motion cues to gather useful information among sequential frames for robust pseudo-label generation. Our hypothesis is that motion continuity can effectively improve the quality of pseudo-labels for the critical task of object detection in autonomous driving systems. 

\begin{figure}[t]
\vspace{-\intextsep}
\centering
\includegraphics[trim=1 1 1 1, clip,keepaspectratio, width=0.45\textwidth]{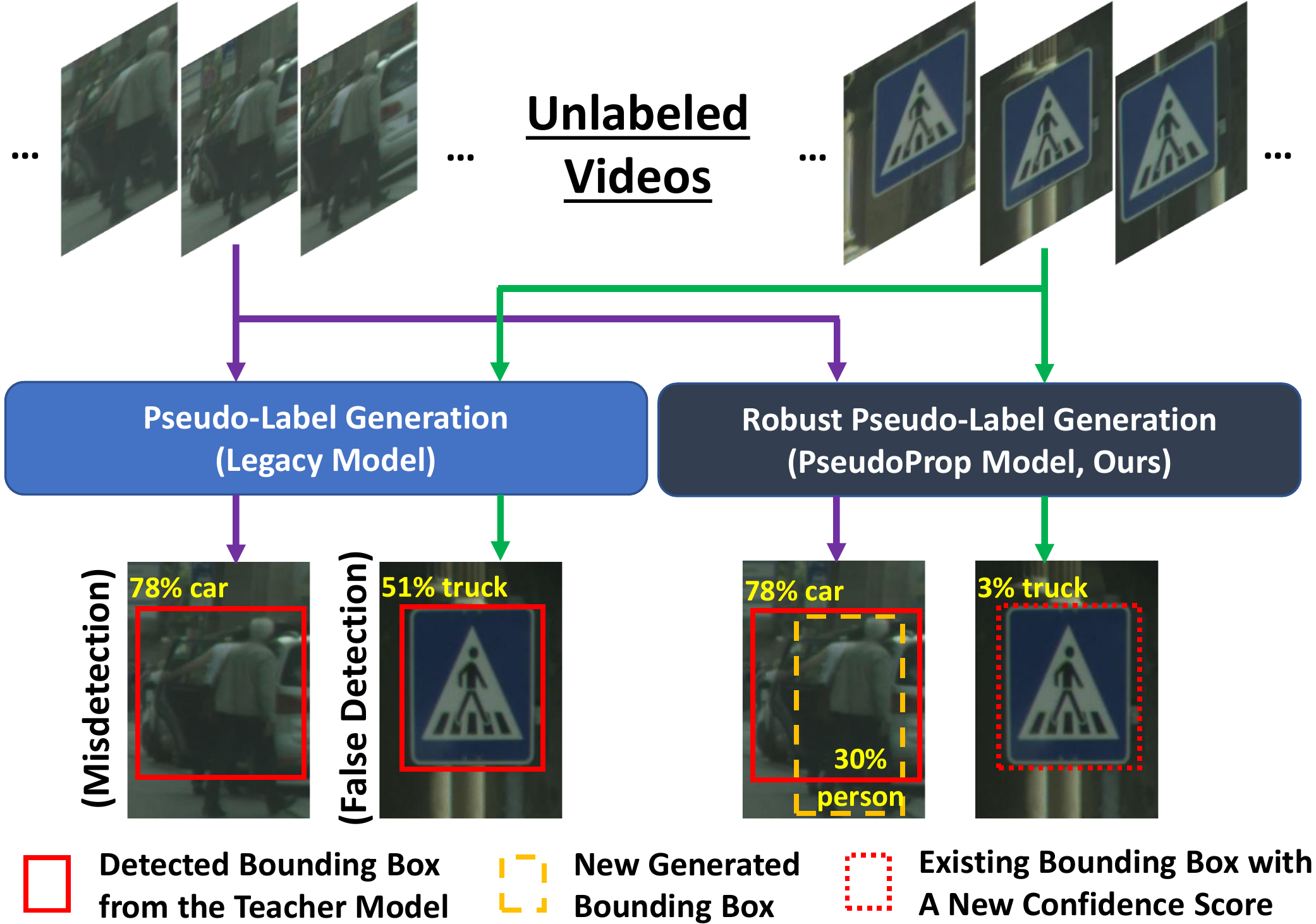}
\vspace{-0.2cm}
\caption{\it An illustrative example comparing the proposed PseudoProp model for robust pseudo-label generation and the legacy semi-supervised object detection, where both models leverage the teacher-student training framework. Images are from the Cityscapes dataset~\cite{Cordts2016Cityscapes}. 
}
\vspace{-0.8cm}
\label{fig:interpretation}
\end{figure}

Despite the idea being intuitive, motion cue is often overlooked in the design of SSL-based object detectors for autonomous driving.
Most existing SSL-based object detection methods worked on single images individually~\cite{rosenberg2005semi, tang2016large, sohn2020simple, tang2021humble}, 
thus the relationship among images is not considered thoroughly. A few object detection works~\cite{misra2015watch, chen2020naive} leverage SSL-based methods on videos to generate pseudo-labels during the training, where the original labeled data are mostly composed of sparse video frames~\cite{Cordts2016Cityscapes}. Each frame from the video can be viewed as an image, and then image-based SSL models can be applied for object detection.
However, such methods suffer from unwanted misdetections and false detections.


In this paper, we propose a novel effective algorithmic model by leveraging motion information for robust pseudo-label generation; our model can effectively improve SSL-based object detection for autonomous driving. Our model is named {\bf PseudoProp}, as it exploits motion as a unique property from the autonomous driving data to robustly propagate pseudo labels. 
Fig.~\ref{fig:interpretation} overviews PseudoProp and compares it against a legacy baseline. 
We adopt a teacher-student framework in PseudoProp, where a teacher model annotates pseudo-labels and a student model  learns and benefits from the pseudo-labels. 


We developed a similarity-aware weighted boxes fusion (SWBF) in PseudoProp based on a novel bidirectional pseudo-label propagation (BPLP) to make pseudo-labels more robust. 
BPLP can generate better pseudo-labels to ease the misdetection problems. On the other hand, BPLP might also generate too many redundant bounding boxes and inevitably introduce false positives due to the exhaustive forward and backward motion prediction. 
To this end, we propose an approach to reduce confidence scores of falsely transferred bounding boxes, based on the similarity between their extracted features. With this similarity check, we adapt the weighted boxes fusion (WBF)~\cite{solovyev2021weighted} originally designed for bounding boxes reduction. Fig.~\ref{fig:overview} explains how PseudoProp works in details.
PseudoProp can alleviate the misdetection problem and significantly reduce the confidence scores for falsely detected objects. Experiments are performed on the large-scale Cityscapes dataset~\cite{Cordts2016Cityscapes}, which demonstrates the effectiveness of PseudoProp on generating robust pseudo-labels for image-based SSL object detection. 

Note that we do not aim to develop a full-scale object tracking algorithm, where the goal is to find the motion trajectories for objects in the video. Instead, we focus on how best to associate motion objects across frames to generate pseudo-labels to improve SSL object detection. Our work is distinct from the conventional video object detection methods, which execute only one round of detection on all video frames. 

The main contributions of our work can be summarized as follows:
\begin{enumerate}
    \item We present a novel framework -- PseudoProp for robust pseudo-label generation for per-image object detection based on motion propagation and SSL.
    \item The proposed SWBF method based on the BPLP approach can solve the misdetection problem and significantly reduce the confidence scores of the false positives in the generated pseudo-labels. 
    \item Experiments on the Cityscapes dataset demonstrate the effectiveness of our model in generating robust pseudo-labels and boosting SSL object detection performance.
\end{enumerate}


\section{Related Work}

This section surveys relevant works of semi-supervised learning, pseudo-label generation, and video motion prediction.

\subsection{SSL with Pseudo-Label Generation}
Object detection is widely used in autonomous driving and video surveillance systems. 
Deep learning methods~\cite{redmon2016you, zhang2018single, dai2016r, grigorescu2020survey, wang2022gan, guo2021robust, guo2021eyes, hu2021exposing} have become {\em de facto} for object detection because of their dominant performance and scalability. However, training deep neural networks requires a large amount of annotated data, and to this end {\em Semi-Supervised Learning (SSL)} has growing popularity in generating or augmenting annotations for training powerful networks.

SSL has been widely applied to computer vision tasks such as object detection \cite{rosenberg2005semi, tang2021humble} and semantic segmentation \cite{papandreou2015weakly, chen2020naive, wang2021nir}. One important idea in this domain is pseudo-labeling \cite{arazo2020pseudo, berthelot2019remixmatch}. For object detection, the pseudo-labels are the bounding boxes of objects in unlabeled data repeatedly generated by a pre-trained model. 
Most SSL-based object detection methods focus on images. For example, STAC \cite{sohn2020simple} and Humble teacher \cite{tang2021humble}.
There are few existing works \cite{misra2015watch, chen2020naive} on object detection by leveraging SSL-based methods on videos to generate pseudo-labels on unlabeled data. In \cite{misra2015watch}, the authors assume that the training videos contain only sparsely labeled bounding boxes and apply a traditional detector (Exemplar-SVM) instead of deep learning-based models for object detection. This results in lower performance for the final model. In addition, the teacher-student framework in Naive-Student \cite{chen2020naive}, can be applied to object detection and semantic segmentation on videos. 
However, this work does not consider the relationship among frames in the same video. Therefore, the generated pseudo-labels may include many misdetections and false detections. 

\begin{figure*}[t!]
\centering
\includegraphics[width=1\textwidth]{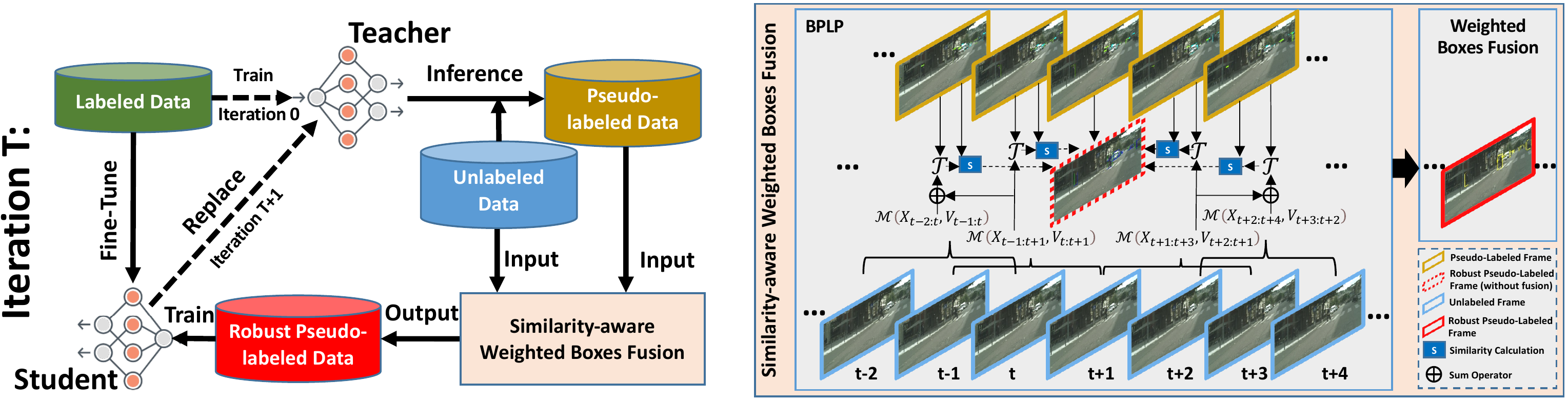}
\vspace{-1.8em}
\caption{\it Overview of our proposed PseudoProp model for robust pseudo-label generation in semi-supervised object detection. (Left) Teacher-student framework for semi-supervised learning. The dash lines represent operations (`Train' and `Replace') only working on specific iterations (Iteration 0 and Iteration $T$+1). (Right) Details of the proposed similarity-aware weighted boxes fusion (SWBF). The figure is better viewed in color.}
\vspace{-0.4cm}
\label{fig:overview}
\end{figure*}

\subsection{Video Motion}

The spatially-displaced convolution network (SDC-Net) in \cite{Reda_2018_ECCV} can predict future video frames based on a two-stage process of first estimating motion then predicting frames. A follow-up work~\cite{Zhu_2019_CVPR} is proposed to enhance semantic segmentation via ground truth label propagation, where the semantic labels are propagated through video motion built on top of the SDC-Net. The spatial-temporal algorithm of \cite{Chen_2018_CVPR} achieves fast object detection on videos through motion-assisted supervised learning. 
In~\cite{McKee2021multi}, the authors propose to generate pseudo-labels and adopt those to train a multi-object tracking model with unlabeled videos. Their pseudo bounding boxes are derived from hallucination of videos, where the videos are generated by motion transformations to simulate various effects. A deep neural tracker is trained with hard example mining. 

There is a large literature on video object tracking~\cite{McKee2021multi, bhat2020know, bergmann2019tracking}. However, a full-blown tracking algorithm is not required in this study, since generating pseudo-labels does not require creating accurate trajectories of objects in every video frame. Our method is more efficient and pertinent to the requirement of the SSL tasks. 




\section{Methodology}
The PseudoProp model contains two parts: (1) a teacher-student framework for training semi-supervised object detector ($\S$~\ref{sec:ts}), (2) Motion Prediction ($\S$~\ref{sec:mp}) and (3) the {\em similarity-aware weighted boxes fusion}  ($\S$~\ref{sec:SWBF}). Fig.~\ref{fig:overview} illustrates detailed components of the parts.



\subsection{Teacher-Student based SSL} 
\label{sec:ts}
The teacher-student framework \cite{zhao2020uncertainty, chen2020naive}, which starts with the idea of knowledge distillation \cite{hinton2015distilling} has been widely applied in SSL. 
In this paper, we use a state-of-the-art teacher-student architecture from \cite{chen2020naive} for video object detection, but our method introduces motion propagation to generate robust pseudo-labels. 

Given the labeled training data $\mathcal{D}_{L}=\{(\widetilde{X}_i, \widetilde{Y}_i)\}_{i=1}^n$, where $n$ is the size of labeled data. $\widetilde{X}_i$ denotes a video frame, and $\widetilde{Y}_i$ is the corresponding human annotations (a set of bounding boxes) of $\widetilde{X}_i$. 
Let $\mathcal{D}_{U}=\{X_i\}_{i=1}^m$ be an unlabeled dataset, where $m$ is the size of unlabeled data. $\mathcal{D}_{U}$ is extracted from multiple video sequences with no human annotation. The human-annotated dataset $\mathcal{D}_{L}$ is exploited to train a teacher network $\theta_1$ by using loss $\mathcal{L}$ for object detection, where $\mathcal{L}$ consists of conventional classification and regression losses for bounding box prediction. Therefore, we have
\begin{equation*}
    \begin{aligned}
    \theta_1^* = \underset{\theta_1}{\arg\min}\frac{1}{n}\sum_{(\widetilde{X}_i, \widetilde{Y}_i)\in \mathcal{D}_{L}}\mathcal{L}(\widetilde{Y}_i, f_{\theta_1}(\widetilde{X}_i)),
    \end{aligned}
\end{equation*}
where $\theta_1^*$ is the trained teacher network with a  prediction function $f$. 

We apply $\theta_1^*$ to generate (or update) the pseudo-labels for all unlabeled data in $\mathcal{D}_{U}$. Therefore, $Y_i = f_{\theta_1^*}(X_i)$, where $Y_i$ is a set of pseudo-labels (bounding boxes) of the unlabeled data $X_i$. Next, we propose the {\em similarity-aware weighted boxes fusion} (SWBF) based on a motion prediction model and a noise-resistant pseudo-label fusion model, to enhance the quality of the generated pseudo-labels. This can be represented as $\overline{Y}_i = \mbox{SWBF} (Y_i), \forall X_i\in \mathcal{D}_{U}$,
where $\overline{Y}_i$ is a set of high-quality pseudo-labels after performing SWBF on $Y_i$.

A student network is subsequently trained with the pseudo-labeled frames using the same loss function $\mathcal{L}$. Thus, we have
\begin{equation*}
    \begin{aligned}
    \theta_2^* = 
\arg\min_{\theta_2}
\frac{1}{m}\sum_{X_i\in \mathcal{D}_{U}}\mathcal{L} 
\left(
\overline{Y}_i, f_{\theta_2}(X_i) 
\right).
    \end{aligned}
\end{equation*}
Since the pseudo-labeled data are noisy, the trained student network cannot achieve high performance yet. The student network is next fine-tuned on $\mathcal{D}_{L}$ before evaluated on the validation or test dataset. This way,
\begin{equation*}
    \begin{aligned}
    \theta_2^{**} = \underset{\theta_2^*}{\arg\min}\frac{1}{n}\sum_{(\widetilde{X}_i, \widetilde{Y}_i)\in \mathcal{D}_{L}}\mathcal{L}(\widetilde{Y}_i, f_{\theta_2^*}(\widetilde{X}_i)).
    \end{aligned}
\end{equation*}
Finally, we replace the teacher $f_{\theta_1^*}$ with the student $f_{\theta_2^{**}}$ and iterate the procedure again until termination.


\subsection{Motion Prediction} 
\label{sec:mp}

To estimate motion from unlabeled video frames, we adopt the SDC-Net~\cite{Reda_2018_ECCV} to predict the motion vector $(du,dv)$ on each pixel $(u,v)$ per frame $X_t$ at time $t$. SDC-Net is proposed to predict a video frame $X_{t+1}$ based on past frame observations as well as their estimated optical flow. 
%
%
It can be trained easily using consecutive frames without providing any manual labels. A later work~\cite{Zhu_2019_CVPR} was proposed to improve the SDC-Net by using video frame reconstruction instead of frame prediction, {\em i.e.}, to apply bi-directional frames to reconstruct the current frame. The predicted frame $\hat X_{t+1}$ and its corresponding predicted pseudo-labels $\hat Y_{t+1}$ can be formulated as:
\begin{equation}
\begin{aligned}
&\hat X_{t+1} = \mathcal{B}(\mathcal{M}(X_{t-\tau:t+1}, V_{t-\tau+1:t+1}), X_t),\\
&\hat Y_{t+1} = \mathcal{T}(\mathcal{M}(X_{t-\tau:t+1}, V_{t-\tau+1:t+1}), Y_t),
\end{aligned}
\label{eq:frame_reconstruction2}
\end{equation}
where $X_{t-\tau:t+1}$ are frames from time $t-\tau$ to $t+1$, $V_{t-\tau+1:t+1}$ are the corresponding optical flows from time $t-\tau+1$ to $t+1$, $\mathcal{M}$ is a convolutional neural network (CNN) to predict per-pixel motion vector $(du,dv)$ on $X_t$, $\mathcal{B}$ is a bilinear sampling operation to interpolate the motion-translated frame into the predicted frame, and $\mathcal{T}$ is a floor operation for deriving pseudo-labels from motion prediction.
We adopt the pre-trained optical flow estimation model  FlowNet2~\cite{IMKDB17} to generate $V$, and this video frame reconstruction approach is used for $\mathcal{M}$. 
We select $\tau=1$ throughout all experiments unless specified otherwise. 
Once the motion vectors on all pixels are available, we use $\mathcal{T}$ to predict $(u,v)$ in $Y_t$ as $(\lfloor u+du \rfloor, \lfloor v+dv \rfloor)$ in $\hat Y_{t+1}$, where $\lfloor\cdot\rfloor$ is the floor operation.





\subsection{Similarity-aware Weighted Boxes Fusion (SWBF)}
\label{sec:SWBF}

In this section, we first propose a bidirectional pseudo-label propagation method to generate candidate pseudo-labels according to the motion predictions. Then we propose a robust fusion method to generate final pseudo-labels.

{\bf Bidirectional Pseudo-Label Propagation (BPLP)}. 
Since the predicted pseudo-labels from the teacher model may contain many false negatives, {\em e.g.}, humans are misdetected as in Fig.~\ref{fig:interpretation}, we apply the motion prediction in Eq.~\eqref{eq:frame_reconstruction2} to propagate pseudo-label prediction. However, such motion prediction can only predict frames and labels in one time step. To make the predicted pseudo-labels more robust at time $t+1$, we propose the {\em bidirectional pseudo-label propagation} to generate pseudo-label proposals across frames, via interpolations from both forward propagation from existing labels and time-reversed backward propagation.
We also apply different propagation lengths $k\in \mathbb{Z}^+$. Specifically,
\begin{equation}
\begin{aligned}
\overline{Y}_{t+1} =  Y_{t+1}\cup \hat Y_{t+1}, \ \ \  \hat Y_{t+1}=\bigcup_{i\in K} \hat Y^{i}_{t+1},
\end{aligned}
\label{eq:motion_update_0}
\vspace{-0.1cm}
\end{equation}
\begin{equation}
\begin{aligned}
&\hat Y^{i}_{t+1} = \mathcal{T}\Big(\sum_{j\in J}\mathcal{M}(X_{t-j:t-j+2}, V_{t+1-j:o}), Y_{t+1-i}\Big),\\
\end{aligned}
\label{eq:motion_update}
\end{equation}
where $K=\{\pm 1,\cdots,\pm(k$-$1), \pm k  \}$, 
$i\in K$, as well as \\
$J=\left\{\begin{matrix}
 \{1,\cdots,i\},& \mbox{if} \ i>0 \\ 
 \{i,\cdots,-1\},& \mbox{if} \ i<0
\end{matrix}\right.$, 
%
$o=\left\{\begin{matrix}
 t+2-j,& \mbox{if} \ i>0 \\ 
 t-j,& \mbox{if} \ i<0
\end{matrix}\right.$,
%
where $\pm$ indicates forward and backward propagation.
In Eq.~\eqref{eq:motion_update_0}, $Y_{t+1}$ is the pseudo-label set of the unlabeled frame $X_{t+1}$ from the teacher model prediction. 
$\hat Y_{t+1}$ is a set containing pseudo-labels from the past and future frames after using motion propagation from Eq.~\eqref{eq:motion_update}. 
$\hat Y^{i}_{t+1}$ is the pseudo-label set from $Y_{t+1-i}$. We also show details on calculating $\hat Y_{t+1}$ in the Appendix. Next, we calculate $\overline{Y}_{t+1}$ for frame $X_{t+1}$ by taking the union of $Y_{t+1}$ and $\hat Y_{t+1}$. 

\begin{figure}
\centering
\includegraphics[width=0.45\textwidth]{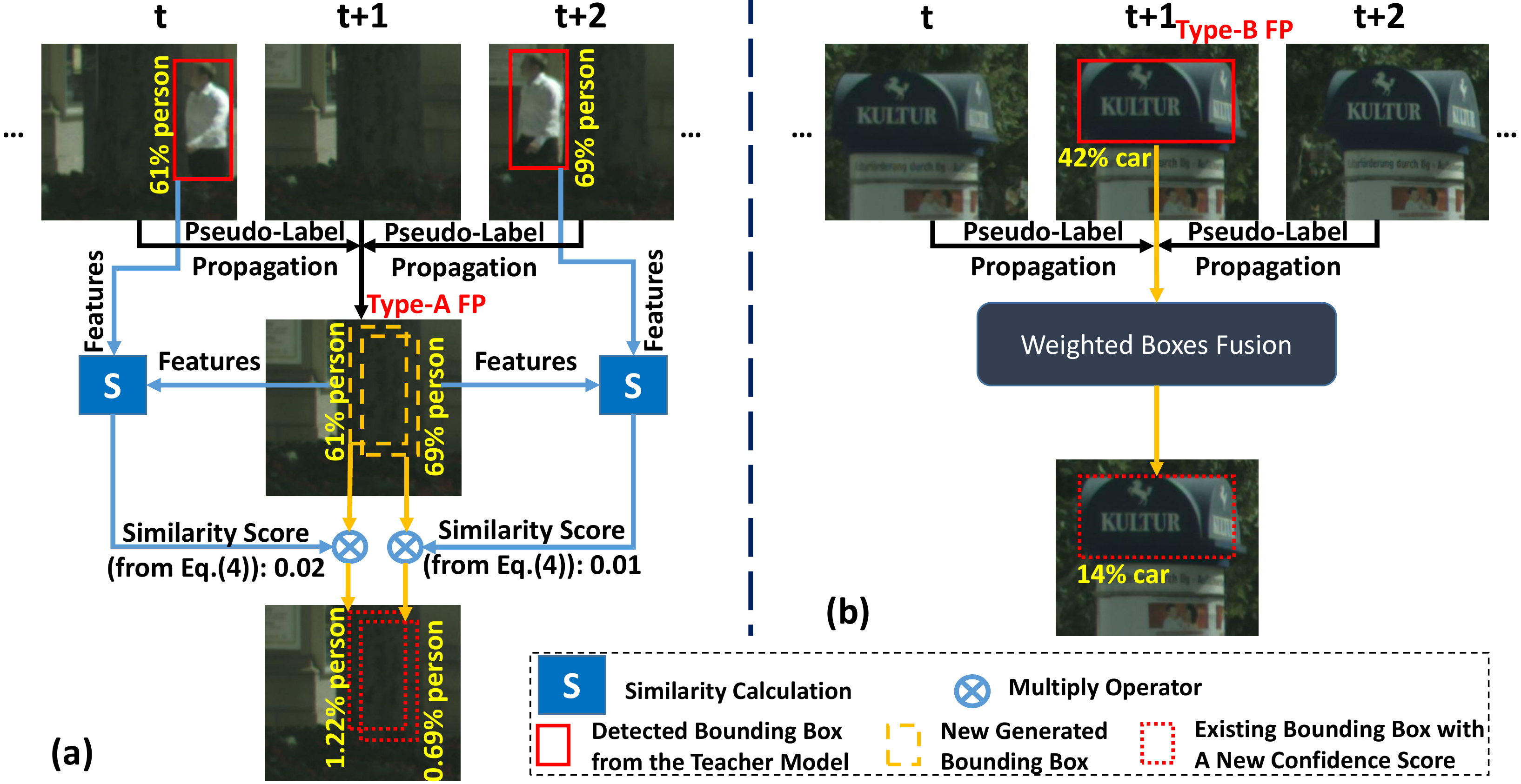}
\vspace{-0.1cm}
\caption{\it Examples of (a) Type-A FP and (b) Type-B FPs produced from BPLP that can be solved by the proposed SWBF method.}
\vspace{-0.3cm}
\label{fig:swfb}
\end{figure}

{\bf Robust Fusion}. BPLP with different settings of $k$ can generate many candidates of pseudo-labels, which may induce additional false positives (FP), which we categorize into two types. 
For \textit{Type-A FP}, refer to an example in Fig.~\ref{fig:swfb}(a), where a person is detected at time $t$ and $t+2$, but not detected at $t+1$ due to occlusion by a tree. In this case, BPLP can generate two bounding boxes at $t+1$ however with low confidence scores caused by the occlusion. 
For \textit{Type-B FP}, refer to Fig.~\ref{fig:swfb}(b), where a billboard is mistakenly detected as a truck at time $t+1$ with a high confidence score. 
Even worse, the number of candidate pseudo-labels (bounding boxes) increases as the value of $k$ increases.
Thus, many redundant bounding boxes $\overline{Y}_{t+1}$ can be predicted in frame $X_{t+1}$. 

We propose a similarity-based approach to reduce the confidence scores of these false pseudo-label predictions. 
We define $Y_{t+1-i}:=\{(L_{t+1-i}^z, P_{t+1-i}^z, S_{t+1-i}^z)\}_{z=1}^{|Y_{t+1-i}|}$, where $L_{t+1-i}^z$, $P_{t+1-i}^z$, $S_{t+1-i}^z$ are the class, positions, confidence score of the $z$-th bounding box in $Y_{t+1-i}$, respectively, and $|\cdot|$ denotes the number of bounding boxes in the set.
Similarly, we define $\hat Y_{t+1}^i:=\{(\hat L_{t+1}^{i,z}, \hat P_{t+1}^{i,z}, \hat S_{t+1}^{i,z})\}_{z=1}^{|\hat Y_{t+1}^i|}$. Note that $L_{t+1-i}^z = \hat L_{t+1}^{i,z}, \forall z$, because (1) we do not change the bounding box class during the propagation and (2) $\hat P_{t+1}^{i,z}$ (assume it is inside the frame) can be obtained from $P_{t+1-i}^z$ by applying $\mathcal{T}$ from Eq.~\eqref{eq:motion_update}. 
Recall that previously $S_{t+1-i}^z= \hat S_{t+1}^{i,z}, \forall z$, which results in Type-A FPs. Therefore, we introduce a similarity function $\sim (\cdot)$ based on $\hat P_{t+1}^{i,z}$ and $P_{t+1-i}^z$ to estimate the bounding box confidence score when transitioned from $S_{t+1-i}^z$ to $\hat S_{t+1}^{i,z}$. To calculate this similarity, we first crop images at frame $X_{t+1-i}$ and $X_{t+1}$ according to the positions $P_{t+1-i}^z$ and $\hat P_{t+1}^{i,z}$, respectively. Then we use a pre-trained neural network to extract the high-level feature representatives from the cropped images. The similarity is obtained by comparing these two high level feature representatives:
\begin{equation}
    \begin{aligned}
    \hat S_{t+1}^{i,z}= S_{t+1-i}^z \cdot \sim 
    \left(
    \mathcal{C}(P_{t+1}^{i,z}),\mathcal{C}(P_{t+1-i}^z)
    \right),
    \end{aligned}
\label{eq:sim_score}
\end{equation}
where $\mathcal{C}(\cdot)$ is a function extracting the high-level feature representatives from the cropped images based on the box positions, and $\sim(\cdot)$ is a similarity function.
The reason we adopt a feature-based approach for similarity calculation is that we prefer assigning similar scores to objects within the same class before and after pseudo-label propagation. The use of such similarity scores can effectively reduce Type-A FPs. 
Fig.~\ref{fig:swfb}(a) shows an illustrative example.

Although this similarity screening can reduce the confidence score for some Type-A FPs, it can not address Type-B FPs. To further filter out redundant prediction boxes, we adapt weighted boxes fusion (WBF)~\cite{solovyev2021weighted}. WBF also reduces the confidence scores of the Type-B FP boxes by averaging the localization and confidence scores of predictions from all sources (previous, current, and future frames) of the same object. Fig.~\ref{fig:swfb}(b) shows an illustrative example. 

Before using WBF, we spilt $\overline{Y}_{t+1}$ into $d$ parts according to the bounding boxes classes, where $d$ is the total number of classes in $\overline{Y}_{t+1}$. We define  $\overline{Y}_{t+1,c} \subseteq \overline{Y}_{t+1}$ as a subset for the $c$-th class. For each subset, i.e. $\overline{Y}_{t+1,c}$, we briefly introduce the fusion procedures as follows:
\begin{enumerate}
    \item We first divide bounding boxes from $\overline{Y}_{t+1,c}$ into different clusters. For each cluster, the intersection over union (IoU) of each two bounding boxes should be greater than a user-defined threshold $\thr$~(in our experiments, $\thr=0.5$  is close to an optimal threshold). 
    \item For boxes in each cluster $r$, we calculate their average confidence score $C_r$, and 
the weighted average for the positions using $C_r=\frac{1}{B}\sum_{l=1}^B C_r^l$ and $P_r =\frac{\sum_{l=1}^B C_r^l\cdot P_r^l}{\sum_{l=1}^B C_r^l}$,
where $B$ is the total number of boxes in the cluster $r$. $C_r^l$ and $P_r^l$ are the confidence score and the position of the $l$-th box in the cluster $r$, respectively.
    \item The above two steps can reduce the redundant bounding boxes. However, it cannot solve the Type-B FP problems. To reduce the confidence score of false detected boxes, we re-scale $C_r$ by 
\begin{equation}
    C_r = C_r \cdot \frac{\min(B,|K|+1)}{|K|+1},
\end{equation} 
where $|K|$ is the size of the set $K$. 
The interpretation is that, if a small number of sources can provide pseudo-labels on an object, this detection is most likely a false detection. An illustrative example is shown in Figure \ref{fig:swfb}(b).

\item Finally, $\overline{Y}_{t+1,c}$ only contains the averaged bounding box information $(c, P_r, C_r)$ from each cluster. 
\end{enumerate}

Therefore, the final $\overline{Y}_{t+1}$ only contains the updated $\overline{Y}_{t+1,c}$ from each class. The details of this fusion method can be found in Appendix. 
The pseudo-code of our proposed SWBF method for $\overline{Y}_{t+1}$ is described in Algorithm \ref{Alg0}.

\begin{algorithm}[ht]
    \caption{Similarity-aware Weighted Boxes Fusion (SWBF) for $\overline{Y}_{t+1}$.}\label{Alg0}
    \SetAlgoLined
    \KwIn{$k$, $d$, \thr, $X$, $Y$ (from the teacher model prediction).}
    \KwOut{The new pseudo-labels $\overline{Y}_{t+1}$ on $X_{t+1}$} 

    \For{$i\in\{\pm 1,\cdots,\pm(k-1), \pm k  \}$}{
    Create or update $\hat Y_{t+1}$ based on 
    Eq.(\ref{eq:motion_update_0}), Eq.(\ref{eq:motion_update}),
    and 
    Eq.(\ref{eq:sim_score}).
    }
    
    Create $\overline{Y}_{t+1}$ based on 
    Eq.(\ref{eq:motion_update_0}).
    
    \For{$c={1,\cdots,d}$}{
    $\overline{Y}_{t+1,c} \leftarrow $ $\mbox{WBF}$($\overline{Y}_{t+1,c} $, \thr) .
    }
    
    Update $\overline{Y}_{t+1}$ based on $\overline{Y}_{t+1,c}, \forall c$.
    
    \Return{$\overline{Y}_{t+1}$}

\end{algorithm}
\section{Experiments}
We evaluate PseudoProp for SSL-based object detection in autonomous driving applications. PseudoProp can be applied in any video dataset of autonomous driving with sequential frames. Our experiments are performed on the popular large-scale Cityscapes dataset~\cite{Cordts2016Cityscapes}, as it fits our scenario well. Due to space limitations, we only present significant results and leave additional results in the supplementary material. 

\subsection{Experimental Settings}

{\bf Datasets and Evaluation Metrics}. 
The Cityscapes dataset \cite{Cordts2016Cityscapes} contains diverse street-views recorded from 50 cities in Germany. 
We use the annotated $2,975$ training images as our training set and the annotated validation $500$ images as our test set. Each image is selected from the $20$-th frame of a $30$-frame video snippet. Therefore, the training video set contains $2,975$ videos.
For each training video, we estimate motion vectors and randomly select 3 frames (excluding the frames already in the training set) without replacement as the pseudo-labeled frames. 
To generate robust pseudo-labels for the student model, we first use the pseudo-labels of the selected frames with confidence scores $> 0.4$ produced from the teacher model. Note that the threshold of $0.4$ yields nearly the best performance in our experiments, and is also widely used in the literature~\cite{tan2020efficientdet}. 
This procedure is important to suppress noisy labels as in~\cite{zou2021pseudoseg}. 
We next apply SWBF on these noisy labels and obtain the robust pseudo-labels. We then use these frames with robust pseudo-labels to construct the pseudo-labeled  sets with 1$\times$, 2$\times$, $3\times$ sizes of the original training set. We report the mean average precision (mAP), mAP with IoU $0.5$ (mAP$^{50}$), and mAP with IoU $0.75$ (mAP$^{75}$)~\cite{lin2014microsoft} as the object detection evaluation results. 


{\bf Computing Infrastructure}.
The algorithm in this paper is implemented with Python 3.6, and it is trained and tested on an Intel(R) Xeon(R) Gold 6150 CPU @2.70GHz with 128GB RAM, and one NVIDIA Tesla V100 GPU with 32GB VRAM. The TensorFlow version is 2.5.0 for EfficientDet-D1 (object detection). The PyTorch version is 1.7.0 for SDC-Net (motion prediction). 

{\bf Teacher and Student Models}. We train a deep neural network object detector as our initial teacher model. We adopt the EfficientDet-D1~\cite{tan2020efficientdet} as the teacher network, with backbone pre-trained on ImageNet~\cite{russakovsky2015imagenet} and the whole network fine-tuned on Cityscapes with batch size 8. 
The maximum number of epochs is $180$. Random image horizontal flip and scaling are applied as our data augmentation strategy. We also adopt the stochastic gradient descent optimizer and a cosine decay learning rate scheduler in the training loop. The learning rate is set to $0.08$, after $1$ epoch warmup with an initial learning rate $0.008$.
After fine-tuning, we obtained 0.355 mAP$^{50}$ performance for Cityscapes on the test set, which is close to the state-of-the-art performance~\cite{michaelis2019benchmarking, chen2018driving}. 

EfficientDet-D1 is also adopted as our student network. We train it with the pseudo-labeled data with a maximal $180$ number of epochs. After that, we fine-tune the student model on the training dataset. The maximum number of epochs is also $180$.
Note that SWBF is a {\em post-processing} method to generate pseudo-labels. Any object detector producing detection boxes can be integrated with our method to take advantage of the improved pseudo-labels. Since the pseudo-labels generation is only used in the training procedure, the inference speed of PseudoProp is the same as EfficientDet-D1.

\textbf{Motion Prediction and Feature Extraction}. A pre-trained SDC-Net~\cite{Reda_2018_ECCV} is used to predict pseudo-labels 
according to the motion vectors presented in Eq.~\eqref{eq:frame_reconstruction2}. 
For calculating the similarity in Eq.~\eqref{eq:sim_score} between two cropped images, we use EfficientNet-B1~\cite{tan2019efficientnet}, which is also the backbone of EfficientDet-D1 for feature extraction. Cosine similarity is used to calculate the value of $\sim (\cdot)$. Feature values are normalized into $[0,1]$ to ensure $\sim (\cdot)\in[0,1]$. Other similarity functions can also be applied. 


\begin{figure*}[t!]
\centering
\includegraphics[width=1\textwidth]{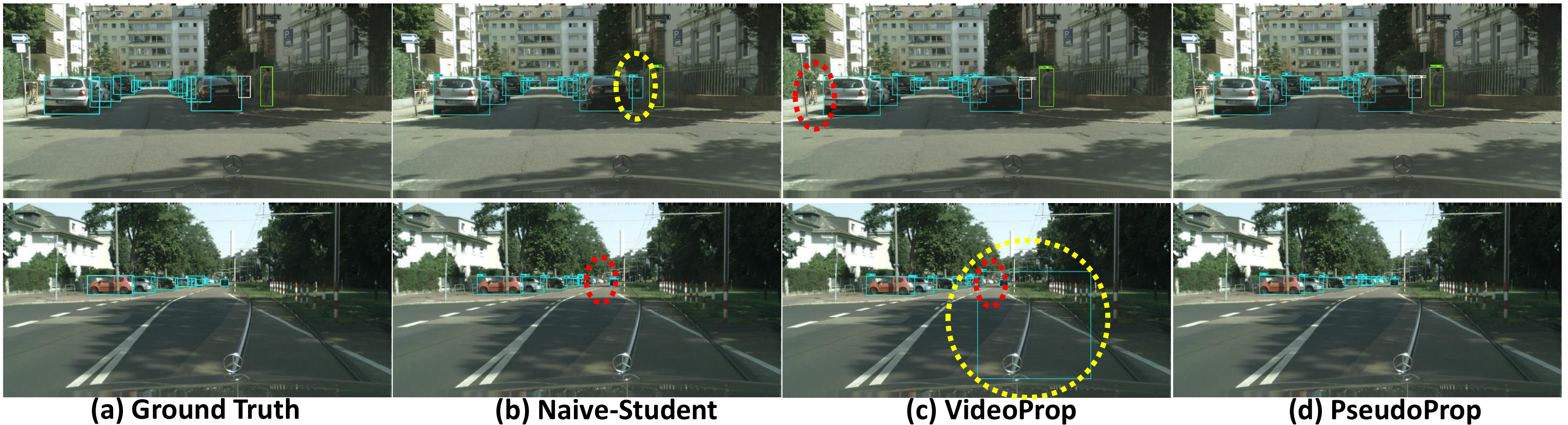}
\vspace{-7mm}
\caption{\it Visual comparison among the (a) ground truth, (b) Naive-Student, (c) VideoProp, and (d) the proposed PseudoProp in the Cityscapes evaluation. Red and yellow dotted ovals indicate the objects  misdetected and falsely detected, respectively. 
}
\label{fig:visual_results}
\vspace{-0.3cm}
\end{figure*}

{\bf Comparisons}. 
We compare PseudoProp with two existing SSL object detectors based on pseudo-label generation. 

\begin{itemize}
    \item {\bf Naive-Student}~\cite{chen2020naive}: The original Naive-Student model uses a teacher-student model with a test-time augmentation. 
    However, their test-time augmentation does not fit directly to object detection. 
    We also use the same selected frames and pseudo-labels with confidence scores higher than $0.4$ to construct the pseudo-labeled datasets. 
    Naive-Student can be viewed as a variant of PseudoProp without SWBF.
    
    \item {\bf VideoProp}~\cite{Zhu_2019_CVPR}: It was originally proposed to improve the semantic segmentation. Here we use it to increase the pseudo-labeled data size via their label propagation. Since this method can only generate pseud-labels according to the ground truth (GT), we follow the approach in their original paper by only considering the 19-th, 20-th, 21-th, and 22-th frames in each video. Specifically, 
    for each training video, we use the 20-th, 21-th frames, and GT labels (from 20-th frame) to predict the labels in the 19-th frame. Similarly, we use the 19-th, 20-th frames, and GT labels to predict the labels in the 21-th frames. We also reconstruct the 21-th frame and combined it with the 19-th and 20-th frames to predict pseudo-labels for the 22-th frame.
    The evaluation pseudo-labeled set is thus constructed for experimental comparison.
\end{itemize}

 
Since PseudoProp is a pure image-based object detector, we also compare the inference performance with state-of-the-art supervised object detection models SSD and DSPNet~\cite{chen2018driving} as baselines.

\begin{table}[t]
\centering
\setlength\tabcolsep{0.1pt}
\scriptsize{
\begin{tabular}{|c|c|c|ccc|}
\hline
\multirow{2}{*}{Ratio} & \multirow{2}{*}{Models} & \multirow{2}{*}{\makecell{Training\\Methods}} & \multicolumn{3}{c|}{Test Set} \\ \cline{4-6} 
                  &                   &                   &    mAP (\%)   &  mAP$^{50}$ (\%)     &   mAP$^{75}$ (\%)      \\ \hline\hline
\multirow{8}{*}{1$\times$} &                  EfficientDet-D1 & s &19.0 & 35.5 &17.2      \\  
                  &                   SSD  & s&-&36.7 &-      \\ \ 
                  &                   DSPNet  & s &-&36.9&-      \\ 
                  &                    VideoProp & ss&21.7&40.3 &19.9       \\ 
                  &                   \makecell{Naive-Student (iteration 1)} &ss &20.8 & 39.0 &18.8        \\  
                  &                  \makecell{Naive-Student (iteration 2)} &ss &22.2 &40.8 &20.3       \\ \cline{2-6} 
                  &                  \makecell{PseudoProp (iteration 1, ours)} &ss &21.6 &40.4 & 19.9    \\ 
                  &                   \makecell{PseudoProp (iteration 2, ours)} &ss&\textbf{\makecell{22.6 (+1.8\%)}}&~\textbf{\makecell{41.4 (+1.5\%)}}&~\textbf{\makecell{20.9 (+3.0\%)}}       \\ \hline\hline
\multirow{5}{*}{2$\times$} &                  VideoProp& ss&21.9 &43.0&19.6     \\ 
                                    &Naive-Student (iteration 1)& ss& 21.2 &38.9 &19.6     \\ 
                                    &Naive-Student* (iteration 1)& ss& 22.8 &43.3 &19.8       \\ \cline{2-6} 
                                   &PseudoProp (iteration 1, ours)& ss&21.7&41.0&20.2       \\  
                                    &PseudoProp* (iteration 1, ours)& ss&\textbf{\makecell{23.2 (+1.8\%)}}&~\textbf{\makecell{44.4 (+2.5\%)}}&~\textbf{\makecell{20.9 (+5.6\%)}}      \\ \hline\hline
\multirow{5}{*}{3$\times$} &                  VideoProp& ss&22.3&42.0 &19.8      \\ 
                                    &Naive-Student (iteration 1)& ss&21.0 &39.7 &18.7     \\ 
                                    &Naive-Student* (iteration 1)& ss& 23.1 &43.2 &21.5       \\ \cline{2-6} 
                                   &PseudoProp (iteration 1, ours)& ss&21.7 &40.0 &   19.8     \\  
                                    &PseudoProp* (iteration 1, ours)& ss&\textbf{\makecell{23.1 (+0\%)}}&~\textbf{\makecell{43.9  (+1.6\%)}}& ~\textbf{\makecell{23.1 (+7.4\%)}}      \\ \hline
\end{tabular}
\vspace{-2mm}
}
\caption{ \it Comparison of mAP, mAP$^{50}$, and mAP$^{75}$ of different object detection models in the Cityscapes evaluation when using 1$\times$, 2$\times$, and 3$\times$ pseudo-labeled data and $k$=1. The improved percentage is also reported when compared with the best baseline result under the same conditions.  ``-" represents that no performance is provided in the existing works. ``s" stands for supervised and ``ss"  for semi-supervised training. All ``ss"-based models are implemented based on the EfficientDet-D1. Best results are shown in bold.}
\label{tab:general_performance}
\vspace{-0.4cm}
\end{table}
\subsection{Results} 

{\bf General Performance}. We set $2$ iterations for the Naive-Student and PseudoProp for the setting of 1$\times$ pseudo-label size. We also test model performance on the size of pseudo-labeled data with 2$\times$ and 3$\times$ settings. Since VideoProp can also generate pseudo-labeled data according to the GT labeled data, we proposed to combine 1$\times$ pseudo-labeled data from PseudoProp and 1$\times$ (2$\times$) pseudo-labeled data from VideoProp as a new 2$\times$ (3$\times$) pseudo-labeled data. We then train the student models on these new datasets and name them PseudoProp*. A similar approach is applied to the Naive-Student model (Naive-Student*). The performance of all models is shown in Table~\ref{tab:general_performance}. Observe that PseudoProp and PseudoProp* achieve superior performance in all settings. They achieve larger performance gains in mAP$^{75}$, which implies our generated bounding boxes are more accurate than the others.
Fig.~\ref{fig:visual_results} shows qualitative comparison results.

{\bf Discussions}.
We next discuss observations and analysis of our experiments.
First, SSL models outperform supervised learning-based models. This is because SSL models use not only the original labeling but also high-quality pseudo-labeled data for training. 
Second, PseudoProp improves the pseudo-label quality of the ordinary Naive-Student thanks to the SWBF. Comparing VideoProp and PseudoProp (iteration 2) in the 1$\times$ setting, we find the improved performance of the motion-based model in the teacher-student architecture. PseudoProp is more general and flexible than VideoProp, as VideoProp only generates pseudo-labels near the GT. 
Third, performance for all SSL-based models can be improved by increasing the pseudo-labeled data size. 
Fourth, PseudoProp* achieves the best performance, as the most high-quality pseudo-labeled data propagated from the GT are used. 
SWBF generates more random pseudo-labels, which increases the diversity of the data. When the generated data ratio increases 2$\times$ to 3$\times$, PseudoProp performance decreases slightly, and a reason is that more noisy data were used for training. 
Finally, the inference time of PseudoProp is no different than any teacher model. 
\subsection{Ablation Study}

\begin{figure}[t]
\begin{subfigure}[t]{0.495\linewidth}
    \includegraphics[width=\linewidth]{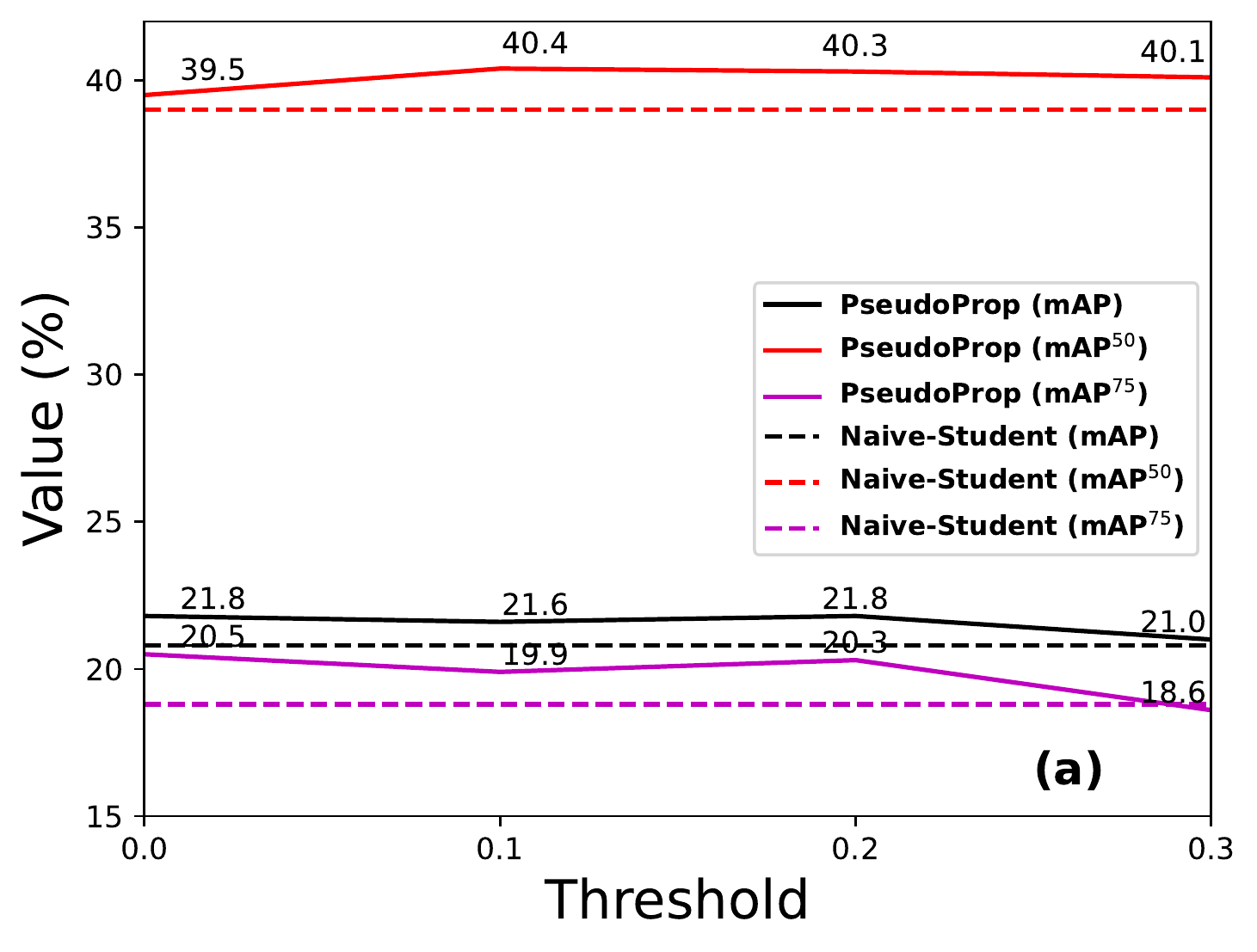}
\end{subfigure}%
    \hfill%
\begin{subfigure}[t]{0.5\linewidth}
    \includegraphics[width=\linewidth]{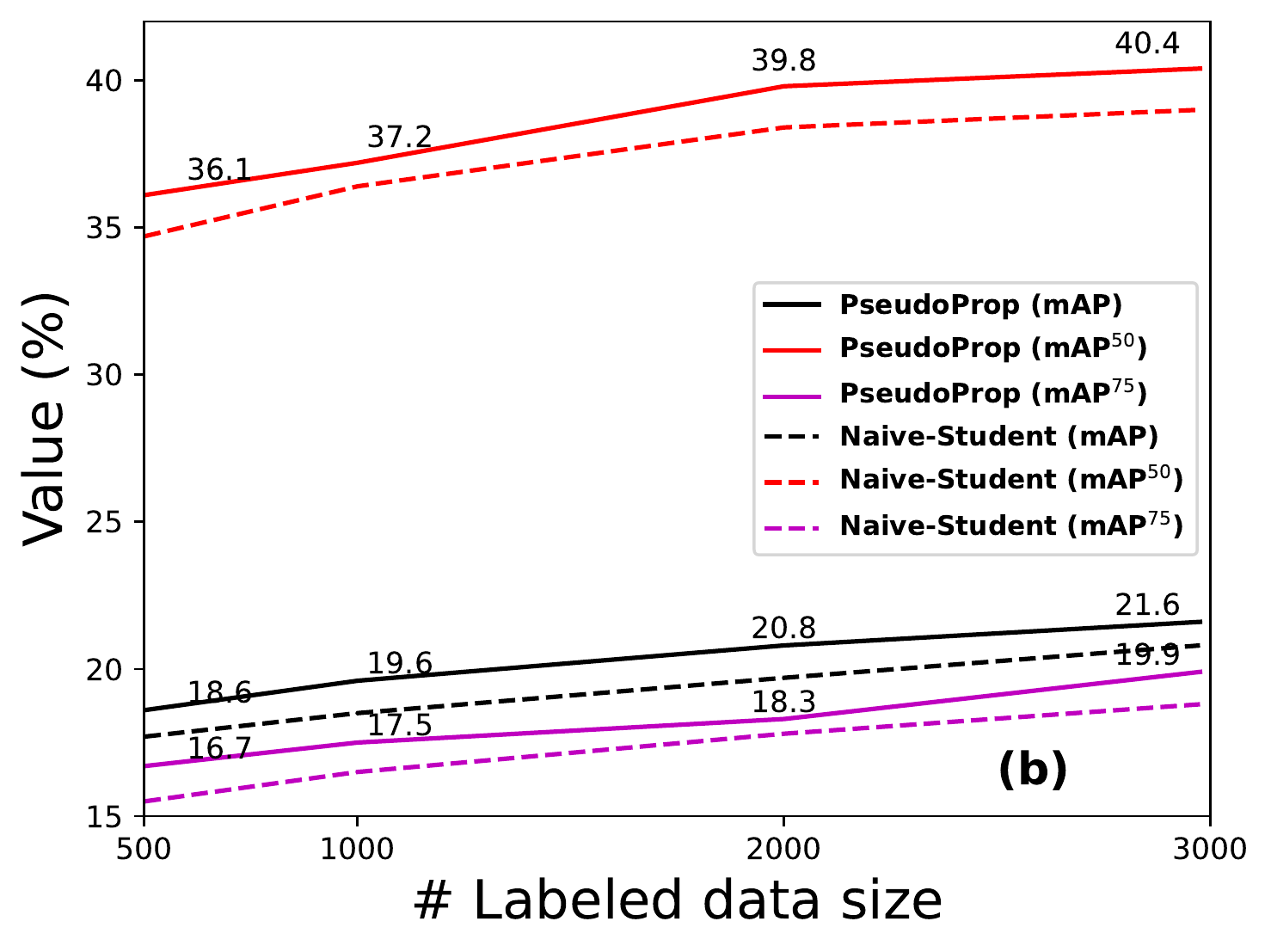}
\end{subfigure}
\vspace{-2em}
\caption{\it Comparison of mAP, mAP$^{50}$, and mAP$^{75}$ of Naive-Student and PseudoProp models on the Cityscapes dataset with $1$ iteration when using 1$\times$ pseudo-labeled data, $k=1$, and (a) different thresholds and (b) different size of labeled data. 
}
\vspace{-1em}
\label{fig:Thr_and_sizeoftrain}
\end{figure}

{\bf Performance on Different Score Thresholds}. After using SWBF to generate the robust pseudo-labels, we set a threshold to remove noisy pseudo-labels based on their confidence scores before passing them to the student model. We compare Naive-Student and PseudoProp using 4 different thresholds $\{0, 0.1, 0.2, 0.3\}$ and report the results in Fig.~\ref{fig:Thr_and_sizeoftrain}(a). 
Observe that even without setting the threshold, PseudoProp outperforms Naive-Student on all evaluation metrics. This means SWBF generates more reliable pseudo-labels. Furthermore, PseudoProp can be more robust by tuning the threshold {\em w.r.t.} the observed mAP$^{50}$ performance. 

{\bf Performance on Low-Data
Regime} (training with only a small amount of labeled data). 
We explore the model performance with different amounts of training data together with a {\em fixed} amount of pseudo-labeled data. This way, we can understand how many labeled data points are needed to fine-tune our model that can effectively speed up the whole training process. Therefore, we randomly extract small sets of labeled data with different sizes such as $500$, $1000$, $2000$ from the original training set. Then we compare Naive-Student and PseudoProp models on each of these sets when performing the fine-tuning of the student model. We show the comparison results in Fig.~\ref{fig:Thr_and_sizeoftrain}(b). Note that PseudoProp outperforms Naive-Student in all evaluation metrics, even in the low-data regime.

\begin{figure}[t]
\begin{subfigure}[t]{0.495\linewidth}
    \includegraphics[width=\linewidth]{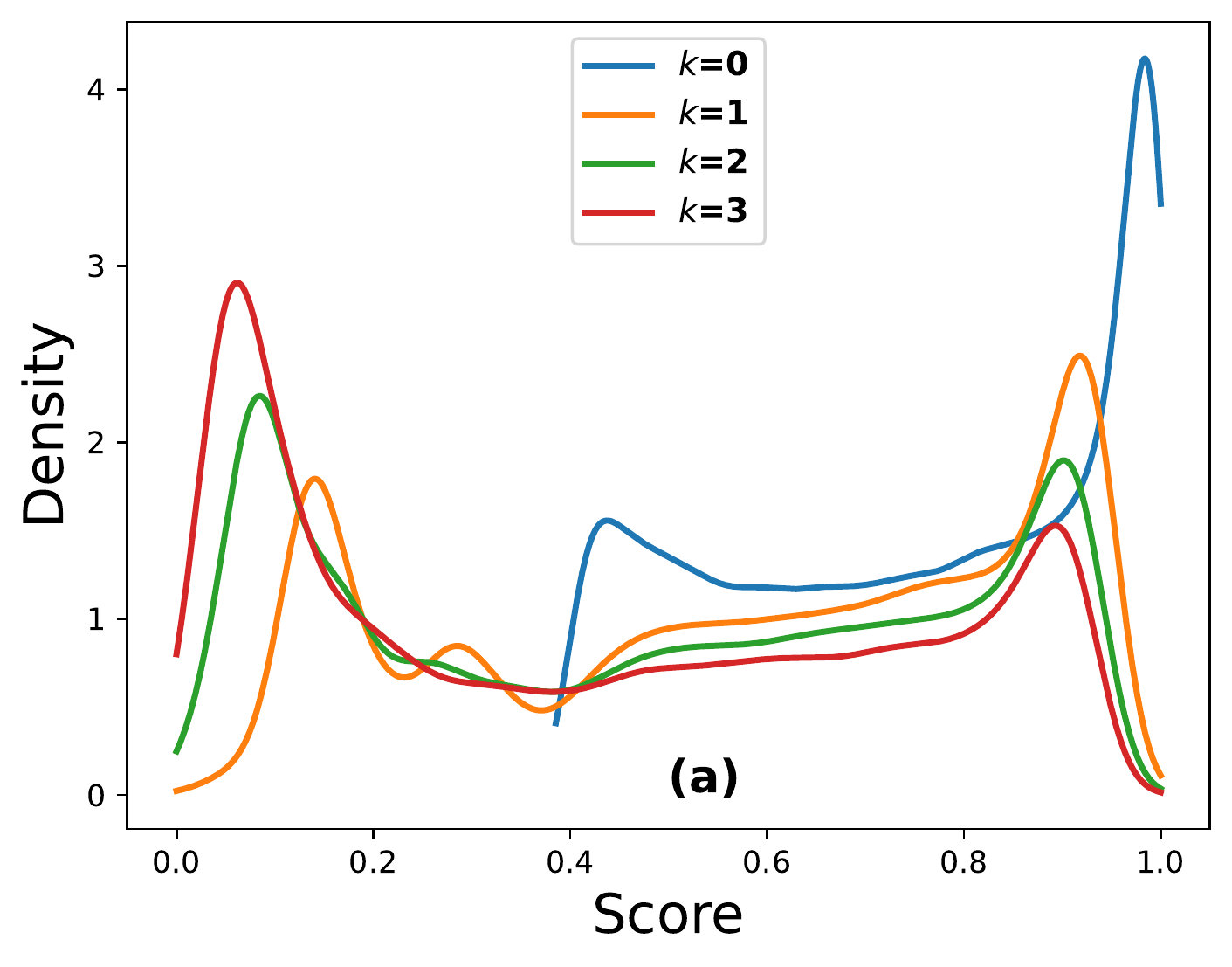}
\end{subfigure}%
    \hfill%
\begin{subfigure}[t]{0.5\linewidth}
    \includegraphics[width=\linewidth]{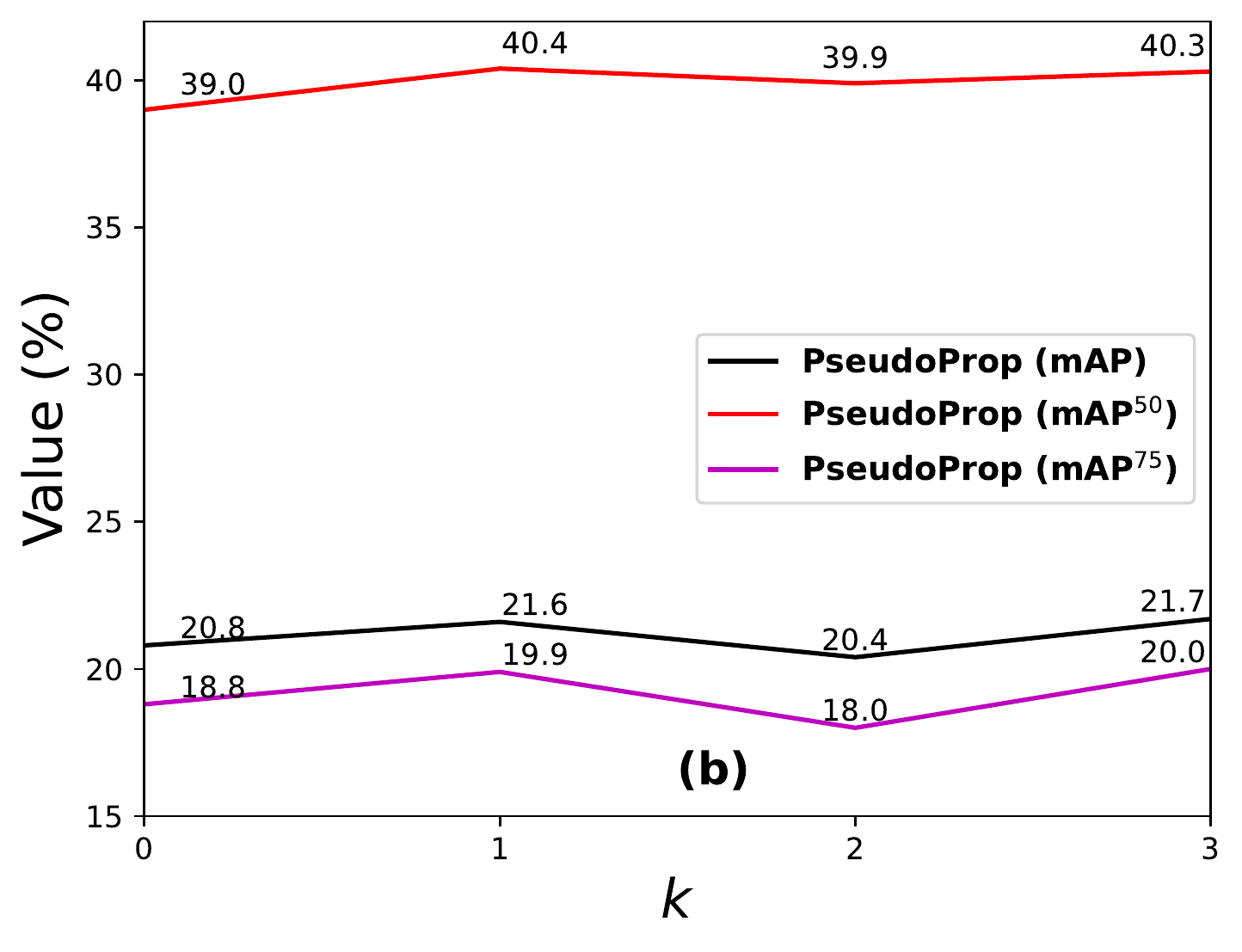}
\end{subfigure}
\vspace{-2em}
\caption{\it (a) Confidence score distributions with different $k$. (b) Comparison of mAP, mAP$^{50}$, and mAP$^{75}$ of Naive-Student ($k$=0) and PseudoProp models ($k$=1,2,3) on the Cityscapes dataset with $1$ iteration, 1$\times$ pseudo-labeled data, and threshold $0.1$.}
\vspace{-2em}
\label{fig:diff_k}
\end{figure}

{\bf Performance on Different Motion Propagation Length $k$}. 
We compare the model performance with different $k$ values, which control the BPLP time steps.
First, in Figure \ref{fig:diff_k}(a), we explore the pseudo-label confidence score distributions with different $k$. Note that $k$=0 represents the Naive-Student. As discussed before, we select pseudo-labels with confidence scores larger than $0.4$ to remove the noisy labels. Therefore, the density curve starts from $0.4$ for $k$=0. In addition, since we use a similarity function in Eq.(\ref{eq:sim_score}) and different $k$ for the fusion method to reduce the false positives, we should expect that our confidence scores for pseudo-labels are smaller than the original ones. Fig. \ref{fig:diff_k}(a) verifies this phenomenon in that the curves shift to the left as $k$ increases. The more neighboring frames we use to propagate pseudo-labels, the more likely the fused final confidence score would decrease. In other words, the confidence score for non-robust pseudo-labels would be suppressed further.

Observe in Fig.~\ref{fig:diff_k}(b) that PseudoProp outperforms Naive-Student model ($k$=0) in mAP$^{50}$ by setting $k>0$. We also observe that increasing $k$ does not always increase performance. This can be explained by the fact that larger $k$ also induces more noise (false positives) from neighboring frames. 
Nevertheless, our method achieves the best performance in mAP and mAP$^{75}$ when setting $k$=3. Since both metrics consider higher requirements for predicting accurate bounding box positions, the proposed SWBF yields better estimation for bounding box coordinates when fusing with a larger $k$. The advantage of accurate bounding box coordinates outweighs the introduced noise. 
\begin{table}[t]
\centering
\setlength\tabcolsep{0.5pt}
\begin{tabular}{c|ccc}
\hline
\multirow{2}{*}{Fusion Methods} & \multicolumn{3}{c}{Test Set} \\ \cline{2-4} 
                  &   mAP (\%)  &  mAP$^{50}$ (\%)     &   mAP$^{75}$ (\%)      \\ \hline
         NMS& 21.0 &39.7 &19.1   \\ 
         NMW& 21.0 &39.8 &19.1   \\ 
         SNMS&21.2 &39.8 &19.3    \\ 
         WBF& 21.0 & 39.6 & 19.1    \\ \hline
         SWBF (ours)&\textbf{21.6 (+1.9\%)} &\textbf{ 40.4 (+1.5\%)}&\textbf{19.9 (+3.1\%)}  \\ \hline
\end{tabular}
\caption{\it Comparison of mAP, mAP$^{50}$, and mAP$^{75}$ for PseudoProp model with different fusion methods in the Cityscapes evaluation. The hyperparameters are $1$ iteration, 1$\times$ pseudo-labeled data, $k=1$, and threshold $0.1$. Best results are shown in bold.}
\label{tab:diff_fusion}
\vspace{-5mm}
\end{table}


{\bf Performance on Different Fusion Methods}. A similarity-based WBF approach is used in PseudoProp for bounding boxes fusion. The following ablation study verifies that our fusion module is indeed better than the state-of-the-art fusion methods~\cite{solovyev2021weighted}, 
including non-maximum suppression (NMS) \cite{neubeck2006efficient}, soft-NMS (SNMS) \cite{bodla2017soft}, non-maximum weighed (NMW) \cite{zhou2017cad}, and the original WBF \cite{solovyev2021weighted}. 
We replace our fusion module with each of the comparison methods and then evaluate performance on the test dataset. Table~\ref{tab:diff_fusion} shows results of this experiment. It is clear that our proposed SWBF method achieves the best performance. Note the performance of the original WBF is very close to other state-of-the-art methods. By adding the proposed similarity modification on top of WBF, SWBF outperforms all comparison methods.

\section{Conclusion}

In this paper, we develop the PseudoProp model to generate robust pseudo-labels that can effectively improve semi-supervised, per-image based object detection. We propose a BPLP method to resolve the misdetection problem in the pseudo-labels. In addition, the proposed similarity approach combined with the WBF method can effectively suppress the confidence scores of the falsely detected bounding boxes. Experimental evaluations on the Cityscapes dataset demonstrate that PseudoProp can improve not only traditional teacher-student based but also motion-based semi-supervised models. Our generated pseudo-labels are reliable for model training, which is validated qualitatively and quantitatively. 

{\bf Future Works.}
There are several important future directions to improve PseudoProp. First, we would consider jointly training an end-to-end model for object detection and motion prediction in our semi-supervised framework. Second, motion prediction error can be taken into account, in which we assume perfect prediction in this work. Third, since the fusion mechanism naturally benefits from soft labels, we believe that incorporating soft labels into PseudoProp will further improve its performance. Fourth, further evaluation can be performed on quantifying how the feature extraction module affects PseudoProp performance.




\pagebreak

{\small
\bibliographystyle{ieee_fullname}
\bibliography{main}
}

\newpage

\appendix
\begin{center}
{\bf \Large  Appendix}
\end{center}
\bigskip 

This appendix provides supplementary details of the proposed method and additional results aside from the main paper.

\section{Self-Consistency of Motion Prediction}

We analyze the self-consistency of motion prediction by validating the accuracy of the estimated motion vectors. 
We first predict the bounding boxes $\hat Y_{t+1}$ on $X_{t+1}$ using SDC-Net~\cite{Reda_2018_ECCV}, given the current ground truth bounding box $Y_t$, $X_t$, and $X_{t-1}$. 
We then reconstruct $\check{Y}_t$ by using reversed motion prediction from $\hat Y_{t+1}$, $X_{t+1}$, and $X_{t+2}$. 
Finally, we measure the IoU between $Y_t$ and $\check{Y}_t$ as the self-consistency estimation. 
We randomly select $100$ images from the Cityscapes dataset~\cite{Cordts2016Cityscapes} and measure such IoU performance. A total of $2,167$ bounding boxes are measured, and the mean of all the measured IoUs is $0.81$. 
We can see from this result that the SDC-Net motion estimation consistency is indeed high. 

\begin{figure}[t]
\begin{subfigure}[t]{0.495\linewidth}
    \includegraphics[width=\linewidth]{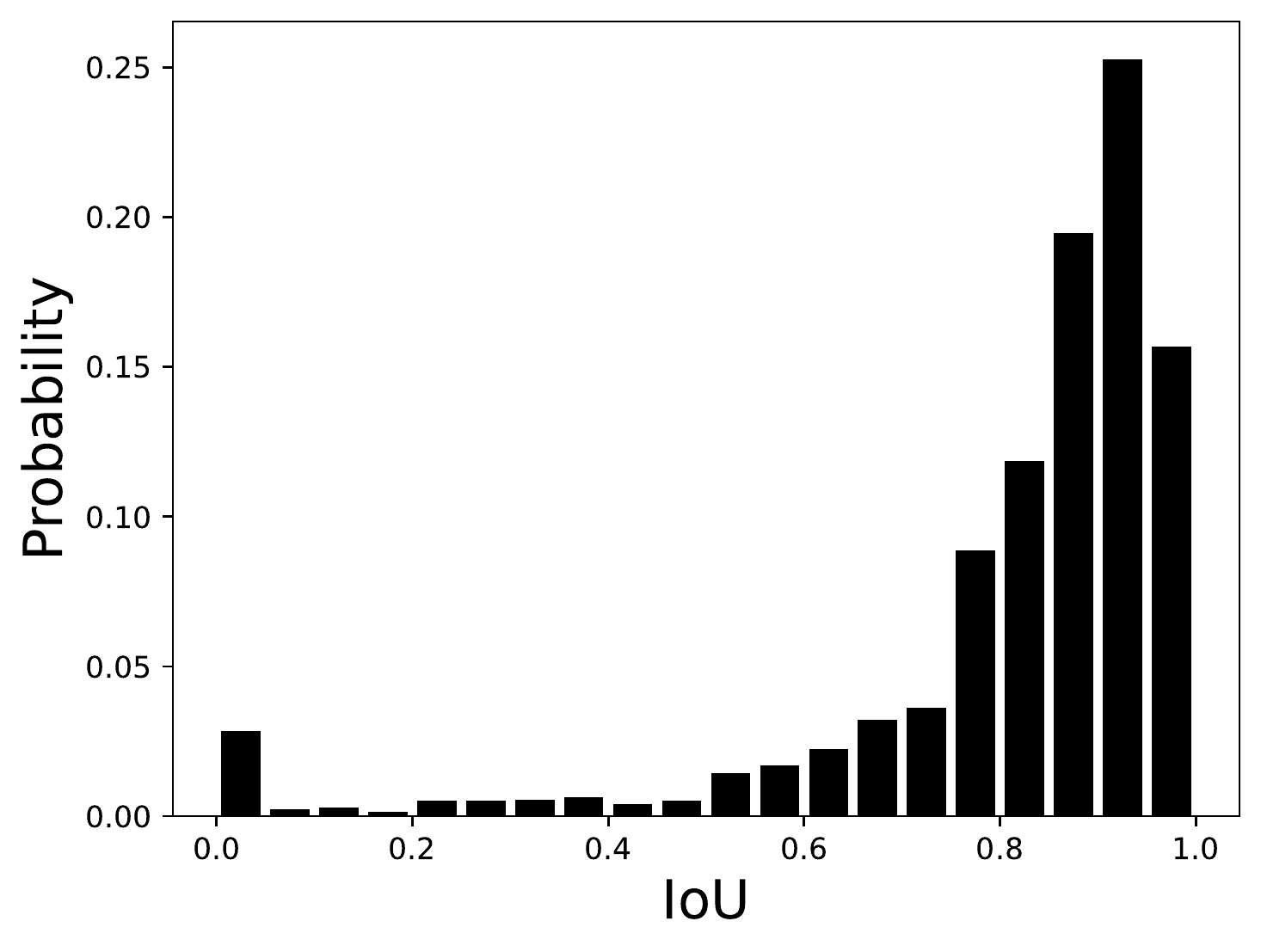}
    \caption{}
\end{subfigure}%
    \hfill%
\begin{subfigure}[t]{0.495\linewidth}
    \includegraphics[width=\linewidth]{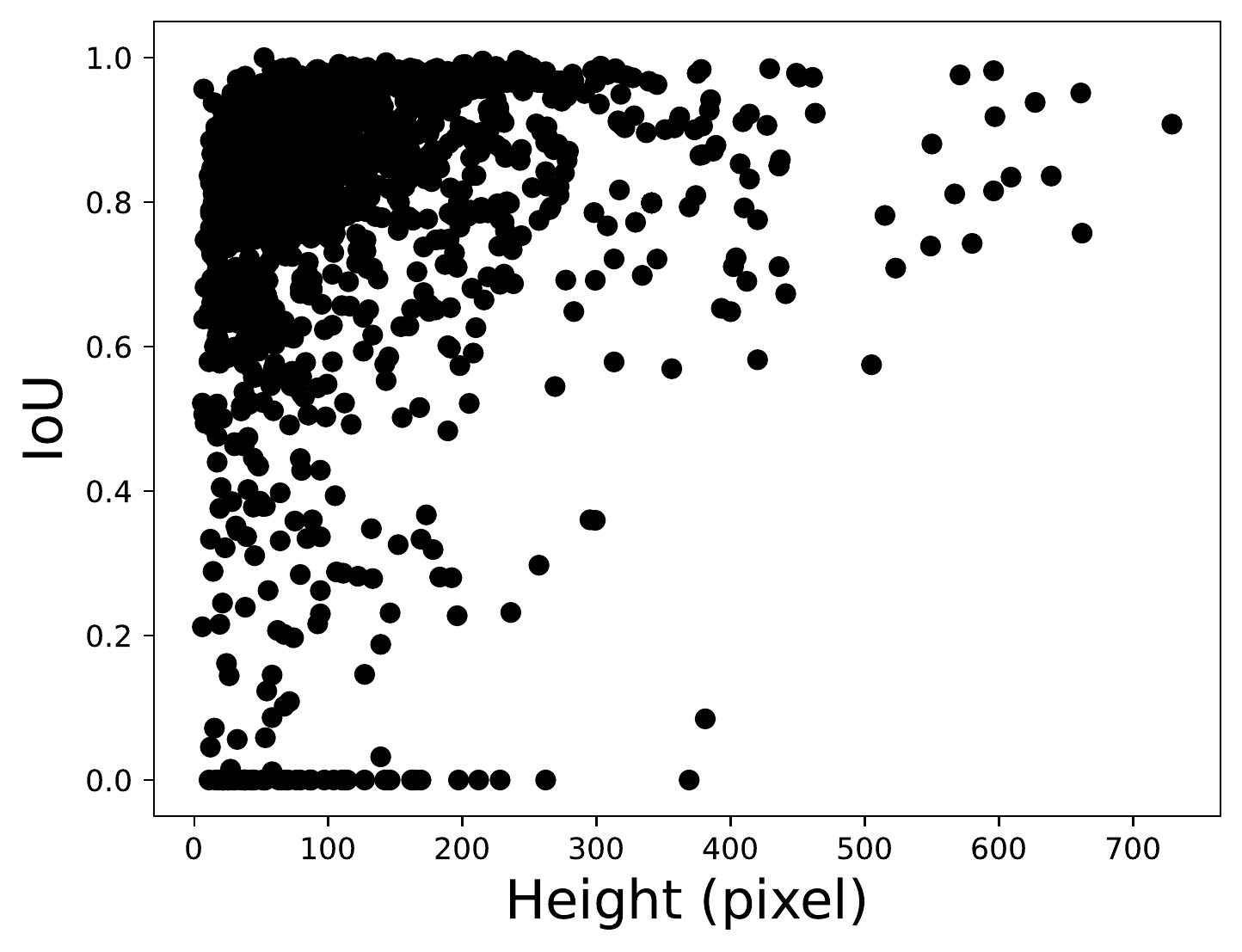}
    \caption{}
\end{subfigure}
\begin{subfigure}[t]{\linewidth}
    \includegraphics[width=\linewidth]{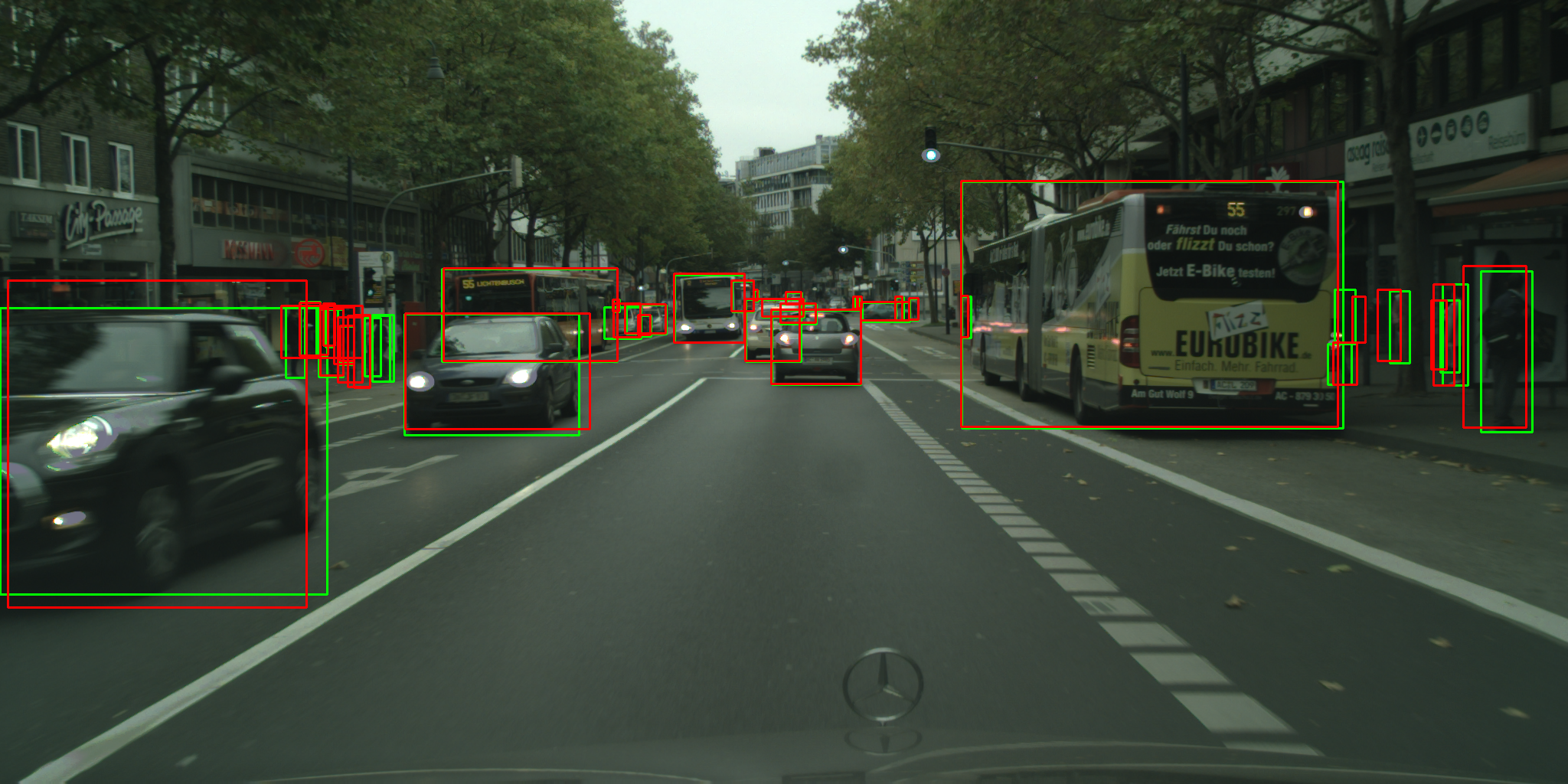}
    \caption{}
\end{subfigure}%
\caption{\small \it {\bf Self-consistency test.}
(a) The probability mass function for IoU using pre-trained SDC-Net. (b) Scatter plot of object height versus IoU.
(c) A visualization example, where green depicts the ground truth  boxes, and red depicts the reconstructed bounding boxes.
}
\label{fig:sdc_pmf_and_sdc_scatter}
\end{figure}

Fig.~\ref{fig:sdc_pmf_and_sdc_scatter}(a) shows the probability mass function of the measured IoUs from the above self-consistency test on the $100$ random Cityscapes images. There are a few IoU $=0$ cases, which is mainly due to: 
(1) The predicted bounding boxes are outside video frames, where the original boxes are near frame boundary: with probability $\text{Pr} (\text{Out} \mid \text{IoU}=0) = 25\%$. 
(2) Small objects are more error-prone to reconstruction: with probability $\text{Pr} (\text{Height} \leq 45 \mid \text{IoU}=0) = 46\%$, where the average height for all objects is $96$ pixels.

The scatter plot in Fig.~\ref{fig:sdc_pmf_and_sdc_scatter}(b) shows the relationship between the object height and IoU for this self-consistency test, with Pearson correlation coefficient $0.07$ (little or no relationship). In other words, the IoU is not biased toward either tall or short objects. 
A similar observation is also found for object area versus IoU. 
Table~\ref{table:sdc_scatter} lists the per-class average IoU from the self-consistency test.
Observe in this table that two specific types of vehicles, namely bus and truck, are with higher IoU. 
This may be due to the slow motion of buses and trucks, which is easier to estimate (in contrast, other vehicle types tend to move faster).
Another potential reason is that buses and trucks do not often appear in groups unlike people and cars. 
The grouping for objects makes motion estimation difficult due to potential occlusions, and Fig.~\ref{fig:sdc_pmf_and_sdc_scatter}(c) shows one example. On the left-hand side of this figure, motion deviation is large for the group of people. Also, observe that all three buses are with good bounding box reconstruction.

\begin{table}[t]
\centering
\scriptsize{
\begin{tabular}{c | c | c | c | c | c | c | c}
\hline
Class & car & person & truck & rider & motorcycle & bicycle & bus\\
\hline
IoU   & $0.84$ & $0.76$ & $0.89$ & $0.78$ & $0.76$ & $0.78$ & $0.92$\\
\hline
\end{tabular}
\caption{\small \it Per-class average IoU from the self-consistency test. Only $7$ classes are shown, as there are no train instance from the $100$ randomly sampled images.
}
\label{table:sdc_scatter}
}
\end{table}

%
%


%

\section{Additional Details of PseudoProp}

In this section, we will provide an example to explain the procedure of BPLP and the details of WBF.

\subsection{An Example Explaining $\widehat{Y}_{t+1}$}

This section explains the bidirectional pseudo-label propagation (BPLP) on the frame $X_{t+1}$ to generate $\hat Y_{t+1}$ by setting $k=2$ in Eq.(2) of the main paper. 

Given $k=2$, then $K=\{-2,-1,1,2\}$. Thus we should do motion propagation from $Y_{t+3}$ (for $i=-2$),  $Y_{t+2}$ (for $i=-1$), $Y_{t}$ (for $i=1$), and $Y_{t-1}$ (for $i=2$) to $Y_{t+1}$, respectively.

For $Y_{t+3}$, the motion vector should be the combination of $\mathcal{M}(X_{t+2:t+4}, V_{t+3:t+2})$ (for $j=-2$) and $\mathcal{M}(X_{t+1:t+3}, V_{t+2:t+1})$ (for $j=-1$). Therefore, we obtain 
\begin{equation*}\small
\begin{aligned}
    &\hat Y_{t+1}^{-2}\\
    &=\mathcal{T}\big(\mathcal{M}(X_{t+2:t+4}, V_{t+3:t+2})+\mathcal{M}(X_{t+1:t+3}, V_{t+2:t+1}), Y_{t+3}. \big)
\end{aligned}
\end{equation*}

For $Y_{t+2}$, the motion vector should be $\mathcal{M}(X_{t+1:t+3}, V_{t+2:t+1})$ (for $j=-1$). Therefore, we have
\begin{equation*}\small
\begin{aligned}
    \hat Y_{t+1}^{-1}=\mathcal{T}\big(\mathcal{M}(X_{t+1:t+3}, V_{t+2:t+1}), Y_{t+2} \big).
\end{aligned}
\end{equation*}

For $Y_{t}$, the motion vector should be $\mathcal{M}(X_{t-1:t+1}, V_{t:t+1})$ (for $j=1$). Therefore, we have
\begin{equation*}\small
\begin{aligned}
    \hat Y_{t+1}^{1}=\mathcal{T}\big(\mathcal{M}(X_{t-1:t+1}, V_{t:t+1}), Y_{t} \big).
\end{aligned}
\end{equation*}

For $Y_{t-1}$, the motion vector should be the combination of $\mathcal{M}(X_{t-2:t}, V_{t-1:t})$ (for $j=2$) and $\mathcal{M}(X_{t-1:t+1}, V_{t:t+1})$ (for $j=1$). Therefore, we have
\begin{equation*}\small
\begin{aligned}
    &\hat Y_{t+1}^{2}=\mathcal{T}\big(\mathcal{M}(X_{t-2:t}, V_{t-1:t})+\mathcal{M}(X_{t-1:t+1}, V_{t:t+1}), Y_{t-1} \big).
\end{aligned}
\end{equation*}

Hence the final $\hat Y_{t+1}$ should be
\begin{equation*}\small
\begin{aligned}
    \hat Y_{t+1} = \hat Y_{t+1}^{-2} \cup \hat Y_{t+1}^{-1} \cup \hat Y_{t+1}^{1} \cup \hat Y_{t+1}^{2}.
\end{aligned}
\end{equation*}


\subsection{The Weighted Box Fusion (WBF)}

This section explains details of the weighted box fusion (WBF), and the following procedure is organized from the content of the original paper~\cite{solovyev2021weighted}.

\begin{enumerate}
    \item First, bounding boxes in $\overline{Y}_{t+1,c}$ are sorted and saved in a descending order list $B$ according to their confidence scores. 
    \item Define two lists $L=\varnothing$ and $F=\varnothing$ for box clusters and fused boxes, respectively. Each position in the list $L$ can contain a set of boxes, which form a cluster. Each position in $F$ contains one box, which is the fused box from the corresponding cluster in $L$.
    \item Iterate through boxes in $B$ and try to find a matching box in the list $F$. The matching should satisfy that IoU is greater than a user-defined threshold \thr.
    \item If a match box is not found, add the current box from $B$ to the end of list $L$ and $F$ as new elements and proceed to the next box in $B$.
    \item If a match is found, add this box to the list $L$ at cluster $r$ corresponding to the matching box in list $F$ 
    \item For boxes in each cluster $r$, we calculate their average confidence score $C_r$, and regard their individual confidence score as a weight for their positions and do the weighted average for the positions as follows.
    \begin{equation*}
        C_r=\frac{1}{T}\sum_{l=1}^T C_r^l,\ \ \ P_r =\frac{\sum_{l=1}^T C_r^l\cdot P_r^l}{\sum_{l=1}^T C_r^l},
    \end{equation*}  
    where $T$ is the total number of boxes in the cluster $r$. $C_r^l$ and $P_r^l$ are the confidence scores and the position of the $l$-th box in the cluster $r$, respectively. 
    \item Re-scale $C_r$ by $C_r = C_r \cdot \frac{\min(T,|K|+1)}{|K|+1}$, where $|K|$ is the size of the set $K$ from Eq. (3). Finally, $\overline{Y}_{t+1,c}$ only contains the average bounding box information $(c, P_r, C_r)$ from each cluster. 
\end{enumerate}

\begin{table*}[th!]
\centering
\scriptsize{
\begin{tabular}{|c|c|c|c|c|c|c|c|c|c|c|c|c|}
\hline
                 Models &\makecell{Pseudo-labled\\Data Ratio}& mAP  & mAP$^{50}$& mAP$^{75}$ & bicycle & bus & car & motorcycle & person & rider & train & truck\\ \hline
                 EfficientDet-D1 & - & 19.0 & 35.5 & 17.2 & 29.4 & 45.4 & 53.6 & 22.8 & 32.2 & 36.7 & 38.9 & 25.4  \\ \hline
                 SSD & - & -  & 36.7 & -  & 30.1 & 47.5 & 60.2 & 26.9 & 36.3 & 37.2 & 28.8 & 26.6  \\ \hline
                 DSPNet & - & -  & 36.9 & -  & 30.0 & 49.3 & 59.1 & 24.6 & 34.9 & 37.7 & 30.4 & 29.4  \\ \hline
\multirow{3}{*}{VideoProp} & 1$\times$ & 21.7 & 40.3 & 19.9 & 32.4 & 52.8 & 59.4 & 26.5 & 35.1 & 39.7 & 42.4 & 33.9  \\ \cline{2-13} 
                  & 2$\times$ & 21.9 & 43.0 & 19.6 & 32.1 & 55.0 & 60.8 & 27.0 & 36.1 & 42.6 & 56.3 & 33.7  \\ \cline{2-13} 
                  & 3$\times$ & 22.3 & 42.0 & 19.8 & 34.1 & 55.1 & 60.3 & 24.4 & 37.6 & 41.5 & 48.4 & 34.7 \\ \hline
\multirow{3}{*}{Naive-Student (iteration 1)} & 1$\times$ & 20.8 & 39.0 & 18.8 & 29.3 & 51.0 & 55.6 & 25.3 & 33.8 & 36.8 & 50.0 & 30.5 \\ \cline{2-13} 
                  & 2$\times$ & 21.2 & 38.9 & 19.6 & 31.1 & 49.7 & 55.5 & 23.4 & 33.9 & 37.7 & 48.3 & 31.8 \\ \cline{2-13} 
                  & 3$\times$ & 21.0 & 39.7 & 18.7 & 29.9 & 50.7 & 56.0 & 26.5 & 34.3 & 38.0 & 52.0 & 30.0 \\ \hline
\multirow{3}{*}{PseudoProp (iteration 1)} & 1$\times$ & 21.6 & 40.4 & 19.9 & 30.9 & 50.3 & 56.3 & 24.5 & 34.9 & 37.5 & 56.4 & 32.2 \\ \cline{2-13} 
                  & 2$\times$ & 21.7 & 41.0 & 20.2 & 30.3& 52.2 &55.9& 25.6 &34.4  &38.2  & 59.6 & 31.6 \\ \cline{2-13} 
                  & 3$\times$ & 21.7 &40.0 &19.8  &31.2 & 50.4 &57.0& 25.4 & 35.8 & 38.4 & 49.3 & 32.3 \\ \hline
\end{tabular}
\vspace{-3mm}
\caption{\small \it Comparison of mAP (\%), mAP$^{50}$ (\%), and mAP$^{75}$ (\%) of different object detection baseline models on the Cityscapes test dataset. For semi-supervised models, we test different pseudo-labeled data ratio. The mAP$^{50}$ (\%) performance for each class is also reported.}
\label{tab:all_results1}
}
\vspace{-2mm}
\end{table*}

\begin{table*}[th!]
\centering
\scriptsize{
\begin{tabular}{|c|c|c|c|c|c|c|c|c|c|c|c|c|}
\hline
                Thresholds  & $k$ & mAP  & mAP$^{50}$& mAP$^{75}$ & bicycle & bus & car & motorcycle & person & rider & train & truck  \\ \hline
\multirow{3}{*}{0} & 1 & 21.8 & 39.5 & 20.5 & 31.0 & 50.0 & 56.1 & 26.2 & 34.2 & 38.1 & 49.4 & 31.1 \\ \cline{2-13} 
                  & 2 & 20.4 & 39.9 & 18.0 & 29.5 & 49.7 & 55.3 & 24.9 & 33.6 & 37.1 & 57.6 & 31.2 \\ \cline{2-13} 
                  & 3 & 21.7 & 40.3 & 20.0 & 30.5 & 51.3 & 55.8 & 26.0 & 33.4 & 37.2 & 57.9 & 30.8 \\ \hline
\multirow{3}{*}{0.1} & 1 & 21.6 & 40.4 & 19.9 & 30.9 & 50.3 & 56.3 & 24.5 & 34.9 & 37.5 & 56.4 & 32.2 \\ \cline{2-13} 
                  & 2 & 21.3 & 39.6 & 19.4 & 30.6 & 51.8 & 55.3 & 25.1 & 34.3 & 38.0 & 52.1 & 29.4 \\ \cline{2-13} 
                  & 3 & 20.8 & 40.1 & 18.9 & 30.7 & 50.9 & 55.4 & 24.1 & 34.5 & 37.8 & 56.0 & 31.1 \\ \hline
\multirow{3}{*}{0.2} & 1 & 21.8 & 40.3 & 20.3 & 29.5 & 51.9 & 56.2 & 24.8 & 33.8 & 37.4 & 58.4 & 30.2 \\ \cline{2-13} 
                  & 2 & 20.6 & 39.1 & 18.6 & 31.0 & 49.2 & 55.3 & 23.4 & 33.7 & 37.5 & 55.1 & 27.9 \\ \cline{2-13} 
                  & 3 & 20.5 & 39.5 & 18.4 & 31.0 & 48.5 & 55.1 & 24.5 & 33.9 & 37.2 & 54.7 & 31.2 \\ \hline
\multirow{3}{*}{0.3} & 1 & 21.0 & 40.1 & 18.6 & 31.7 & 48.6 & 56.5 & 22.2 & 34.0 & 37.1 & 58.5 & 32.3 \\ \cline{2-13} 
                  & 2 & 20.7 & 39.2 & 18.0 & 30.7 & 48.0 & 55.5 & 23.8 & 33.9 & 37.3 & 55.1 & 29.5 \\ \cline{2-13} 
                  & 3 & 20.7 & 39.3 & 19.7 & 30.1 & 48.3 & 55.4 & 21.2 & 33.8 & 36.9 & 56.4 & 32.4 \\ \hline
\end{tabular}
\caption{\small \it Comparison of mAP (\%), mAP$^{50}$ (\%), and mAP$^{75}$ (\%) of the PseudoProp model on the Cityscapes test dataset when using different thresholds and different $k$ values. The mAP$^{50}$ (\%) performance for each class is also reported.}
\label{tab:all_results2}
}
\end{table*}

\begin{table*}[th!]
\centering
\scriptsize{
\begin{tabular}{|c|c|c|c|c|c|c|c|c|c|c|c|}
\hline
Fusion Methods & mAP  & mAP$^{50}$& mAP$^{75}$ & bicycle & bus & car & motorcycle & person & rider & train & truck  \\ \hline
NMS & 21.0 & 39.7 & 19.1 & 30.0 & 51.1 & 55.3 & 24.6 & 34.3 & 37.3 & 54.5 & 30.8 \\ \hline
NMW & 21.0 & 39.8 & 19.1 & 29.1 & 50.0 & 55.2 & 24.9 & 34.3 & 36.0 & 56.5 & 32.3 \\ \hline
SNMS & 21.2 & 39.8 & 19.3 & 30.2 & 50.7 & 55.1 & 24.6 & 33.2 & 36.5 & 57.7 & 30.1 \\ \hline
WBF & 21.0 & 39.6 & 19.1 & 30.6 & 49.4 & 55.3 & 24.6 & 34.0 & 37.0 & 55.9 & 30.1 \\ \hline
SWBF & 21.6 & 40.4 & 19.9 & 30.9 & 50.3 & 56.3 & 24.5 & 34.9 & 37.5 & 56.4 & 32.2 \\ \hline
\end{tabular}
\caption{\small \it Comparison of mAP (\%), mAP$^{50}$ (\%), and mAP$^{75}$ (\%) of the PseudoProp model on the Cityscapes test dataset when using different fusion methods. The mAP$^{50}$ (\%) performance for each class is also reported.}
\label{tab:all_results3}
}
\end{table*}

\section{Additional Experimental Results}

In this section, we will provide more experimental results.

\begin{table*}[th!]
\centering
\scriptsize{
\begin{tabular}{|c|c|c|c|c|c|c|c|c|c|c|c|c|}
\hline
                  Methods  & \makecell{Labled\\Data Size}& mAP  & mAP$^{50}$& mAP$^{75}$ & bicycle & bus & car & motorcycle & person & rider & train & truck  \\ \hline
\multirow{3}{*}{Naive-Student (iteration 1)} & 2000 & 20.8 & 39.8 & 18.3 & 30.0 & 49.4 & 55.5 & 25.0 & 35.1 & 38.1 & 56.2 & 29.5 \\ \cline{2-13} 
                  & 1000 & 18.5 & 36.4 & 16.5 & 29.3 & 47.3 & 53.8 & 24.1 & 33.0 & 34.8 & 41.3 & 27.5 \\ \cline{2-13} 
                  & 500 & 17.7 & 34.7 & 15.5 & 28.8 & 45.7 & 53.6 & 21.4 & 32.5 & 34.7 & 37.6 & 23.4 \\ \hline
\multirow{3}{*}{PseudoProp (iteration 1)} & 2000 & 20.8 & 39.8 & 18.3 & 30.0 & 49.4 & 55.5 & 25.0 & 35.1 & 38.1 & 56.2 & 29.5 \\ \cline{2-13} 
                  & 1000 & 19.6 & 37.2 & 17.5 & 28.2 & 47.5 & 54.4 & 23.8 & 33.3 & 36.1 & 42.2 & 32.3 \\ \cline{2-13} 
                  & 500 & 18.6 & 36.1 & 16.7 & 28.5 & 49.9 &54.2  & 22.1 & 33.2 & 34.4 & 36.4 & 30.1 \\ \hline
\end{tabular}
\caption{\small \it Comparison of mAP (\%), mAP$^{50}$ (\%), and mAP$^{75}$ (\%) of the Naive-Student and PseudoProp models on the Cityscapes test dataset when using different small labeled data size. The mAP$^{50}$ (\%) performance for each class is also reported.}
\label{tab:all_results4}
}
\end{table*}

\begin{table*}[th!]
\centering
\scriptsize{
\begin{tabular}{|c|c|c|c|c|c|c|c|c|c|c|c|}
\hline
 Methods& mAP  & mAP$^{50}$& mAP$^{75}$ & bicycle & bus & car & motorcycle & person & rider & train & truck \\ \hline
Naive-Student (iteration 2) & 22.2 & 40.8 & 20.3 & 30.9 & 50.6 & 56.7 & 25.7 & 36.1 & 38.1 & 55.5 & 32.7 \\ \hline
PseudoProp (iteration 2) & 22.6 & 41.4 & 20.9 & 32.9 & 50.0 & 58.2 & 24.7 & 36.9 & 39.5 & 55.7 & 33.6 \\ \hline
\end{tabular}
\caption{\small \it Comparison of mAP (\%), mAP$^{50}$ (\%), and mAP$^{75}$ (\%) of the Naive-Student and PseudoProp models on the Cityscapes test dataset at iteration 2. The mAP$^{50}$ (\%) performance for each class is also reported.}
\label{tab:all_results5}
}
\end{table*}

\begin{table*}[th!]
\centering
\scriptsize{
\begin{tabular}{|c|c|c|c|c|c|c|c|c|c|c|c|c|}
\hline
                 Models &\makecell{Pseudo-labled\\Data Ratio} & mAP  & mAP$^{50}$& mAP$^{75}$ & bicycle & bus & car & motorcycle & person & rider & train & truck  \\ \hline
\multirow{2}{*}{Naive-Student* (iteration 1)} & 2$\times$ & 22.8 & 43.3 & 19.8 & 34.0 & 54.6 & 60.5 & 26.1 & 38.0 & 41.2 & 56.6 & 35.6 \\ \cline{2-13} 
                  & 3$\times$ & 23.1 & 43.2 & 21.5 & 33.3 &  54.1&  60.5& 28.3 & 38.5 & 41.2 & 51.8 & 38.2 \\ \hline
\multirow{2}{*}{PseudoProp* (iteration 1)} & 2$\times$ & 23.2 & 44.4 & 20.9 & 34.7 & 50.8 & 60.8 & 31.4 & 38.3 & 41.4 & 62.1 & 35.6 \\ \cline{2-13} 
                  & 3$\times$ & 23.1 & 43.9 & 21.3 & 34.2 & 55.2 & 61.3 & 30.8 & 39.0 & 41.7 & 53.4 & 35.5 \\ \hline
\end{tabular}
\caption{\small \it Comparison of mAP (\%), mAP$^{50}$ (\%), and mAP$^{75}$ (\%) of the Naive-Student* and PseudoProp* models on the Cityscapes test dataset at iteration 1 when using different pseudo-labled data ratio. The mAP$^{50}$ (\%) performance for each class is also reported.}
\label{tab:all_results6}
}
\end{table*}

\subsection{The Details of Model Performance}

We show the details of model performance under different settings and also report the mAP$^{50}$ performance on each class in Table \ref{tab:all_results1}, \ref{tab:all_results2}, \ref{tab:all_results3}, \ref{tab:all_results4}, \ref{tab:all_results5}, and \ref{tab:all_results6}. 

From Table \ref{tab:all_results1}, we can find our method can get the best performance when using 1$\times$ pseudo-labeled data. However, when we increase pseudo-labeled data, the VideoProp method has better performance. The reason is that the generated pseudo-labels from the VideoProp method are very close to the GT labels. Therefore, the pseudo-labeled data has high quality.  But this method can only generate pseudo-labels near the GT. Our model is more flexible and general than the VideoProp. On the other hand, if we compare the model performance in the `train' class, it is clear that our method has high performance in the rare class when using 1$\times$ and $2\times$ pseudo-labeled data. For Table \ref{tab:all_results2}, we can find the mAP, mAP$^{50}$, and mAP$^{75}$ performance of PseudoProp method can achieve the best when we set $k=1$. For Table \ref{tab:all_results3}, we can find the SWBF fusion method outperforms other methods. Specifically, when we compare WBF and SWBF, it is clear that applying the similarity method to the WBF method can improve the model performance. For Table \ref{tab:all_results4}, when we decrease the labeled data size, the performance gap between Naive-Student and our PseudoProp will become large. This means the generated pseudo-labels from our model are more reliable. For Table \ref{tab:all_results5}, comparing Naive-Student and PseudoProp, we can find the proposed SWBF method can be well adapted to the teacher-student semi-supervised learning framework. For Table \ref{tab:all_results6}, when we increase pseudo-labeled data size, both model performances  will be decreased. The reason is that more pseudo-labeled data indicates more noise will be inserted and used in the training procedure. However, we can find our method can also get the best performance in mAP$^{50}$.

\subsection{Additional Visual Results}

\begin{figure*}[t!]
\captionsetup[subfigure]{justification=centering}
\centering
        \begin{subfigure}[b]{0.49\textwidth}
                \includegraphics[width=\linewidth]{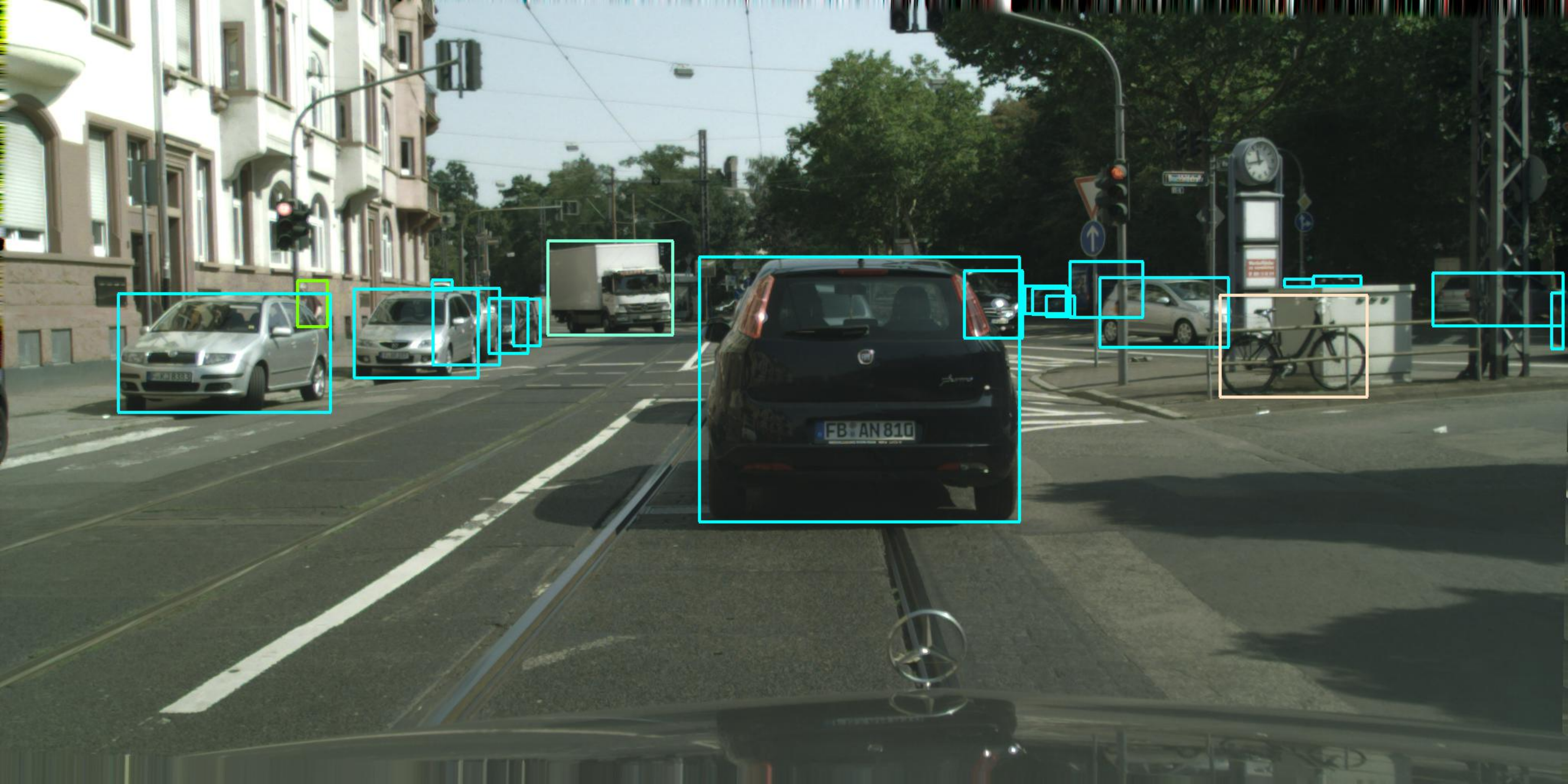}
                \caption*{Ground Truth}
        \end{subfigure}%
        \rulesep
        \begin{subfigure}[b]{0.49\textwidth}
                \includegraphics[width=\linewidth]{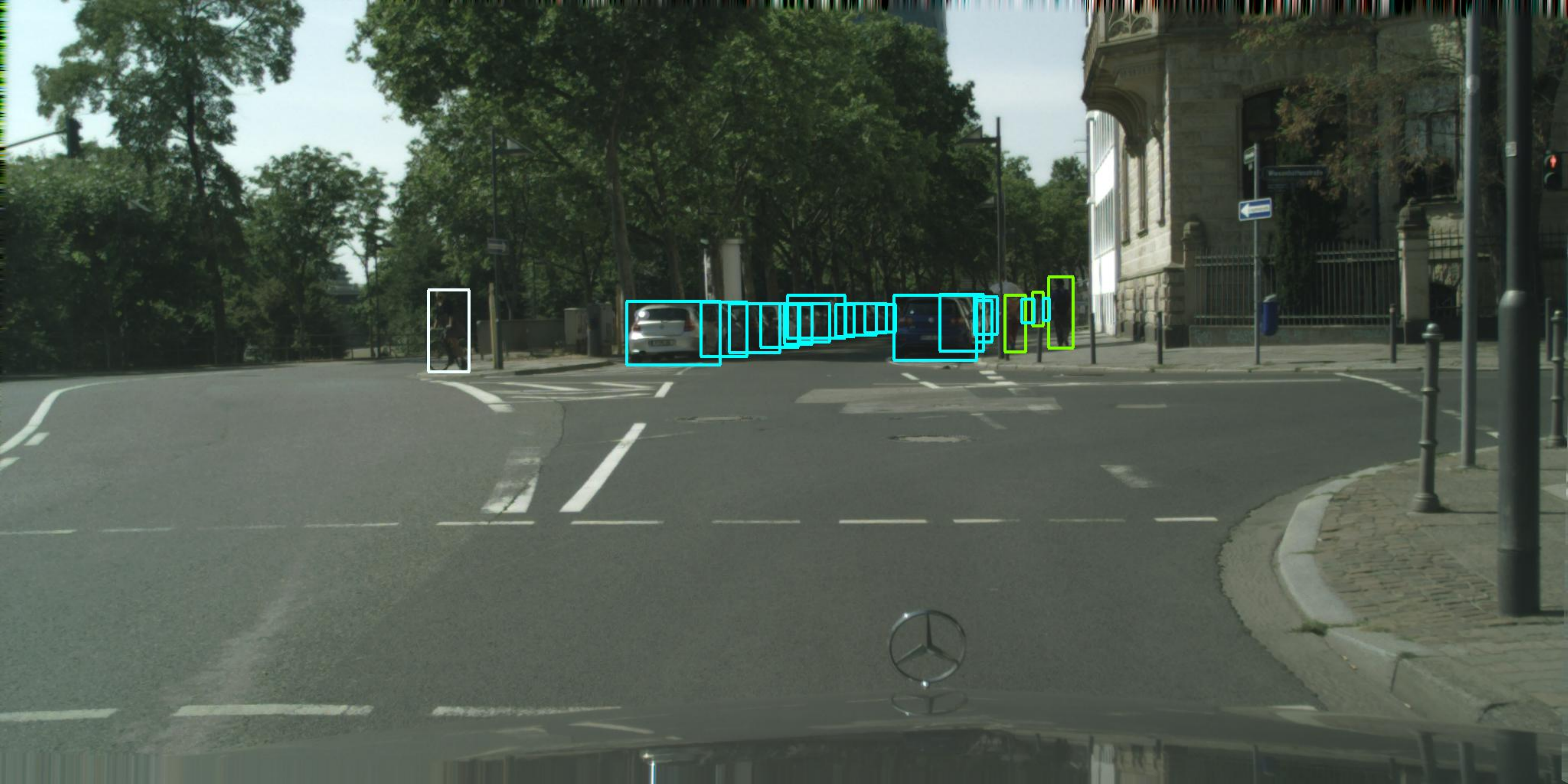}
                \caption*{Ground Truth}
        \end{subfigure}%
        
        \begin{subfigure}[b]{0.49\textwidth}
                \includegraphics[width=\linewidth]{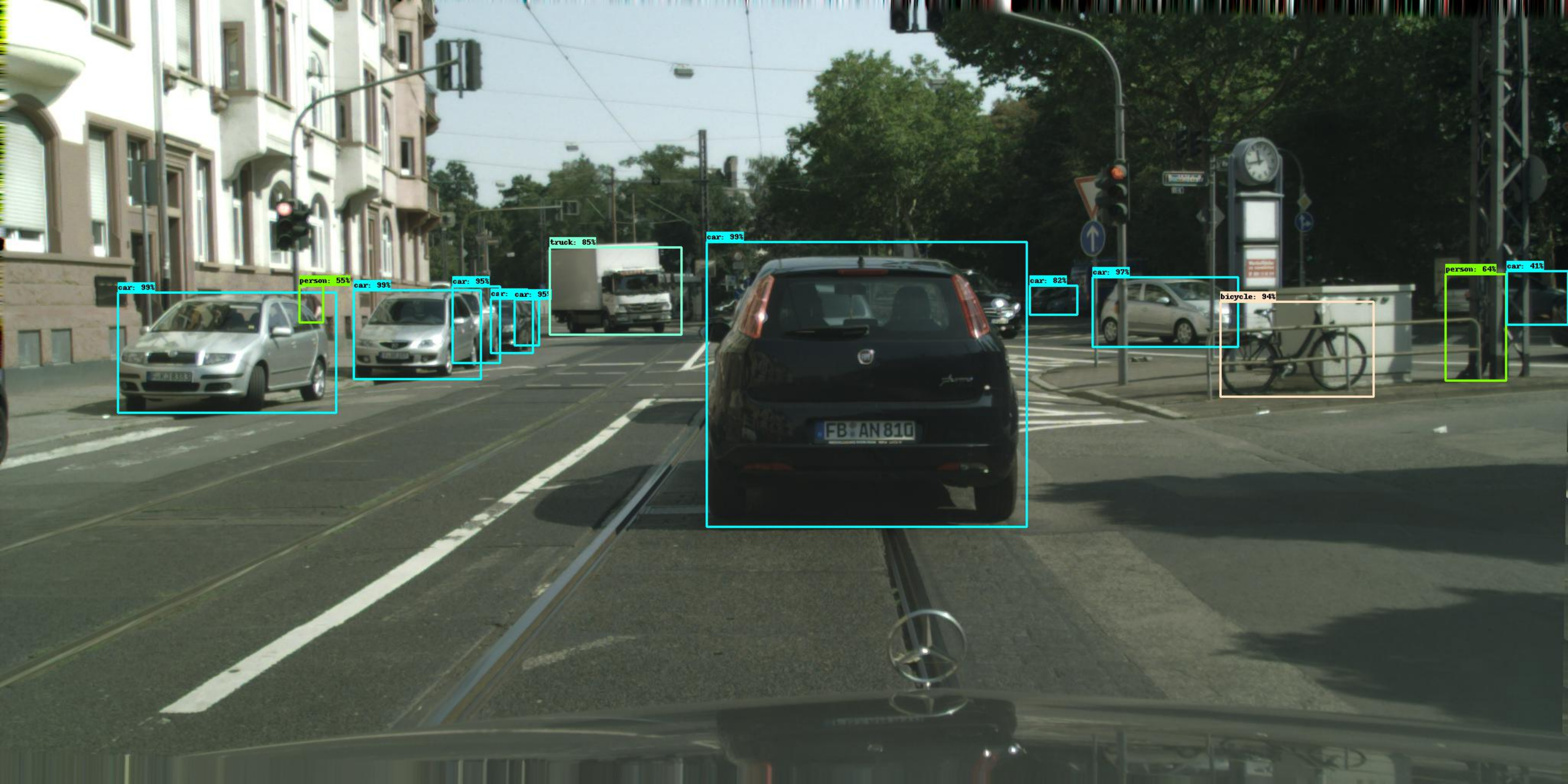}
                \caption*{Naive-Student}
        \end{subfigure}%
        \rulesep
        \begin{subfigure}[b]{0.49\textwidth}
                \includegraphics[width=\linewidth]{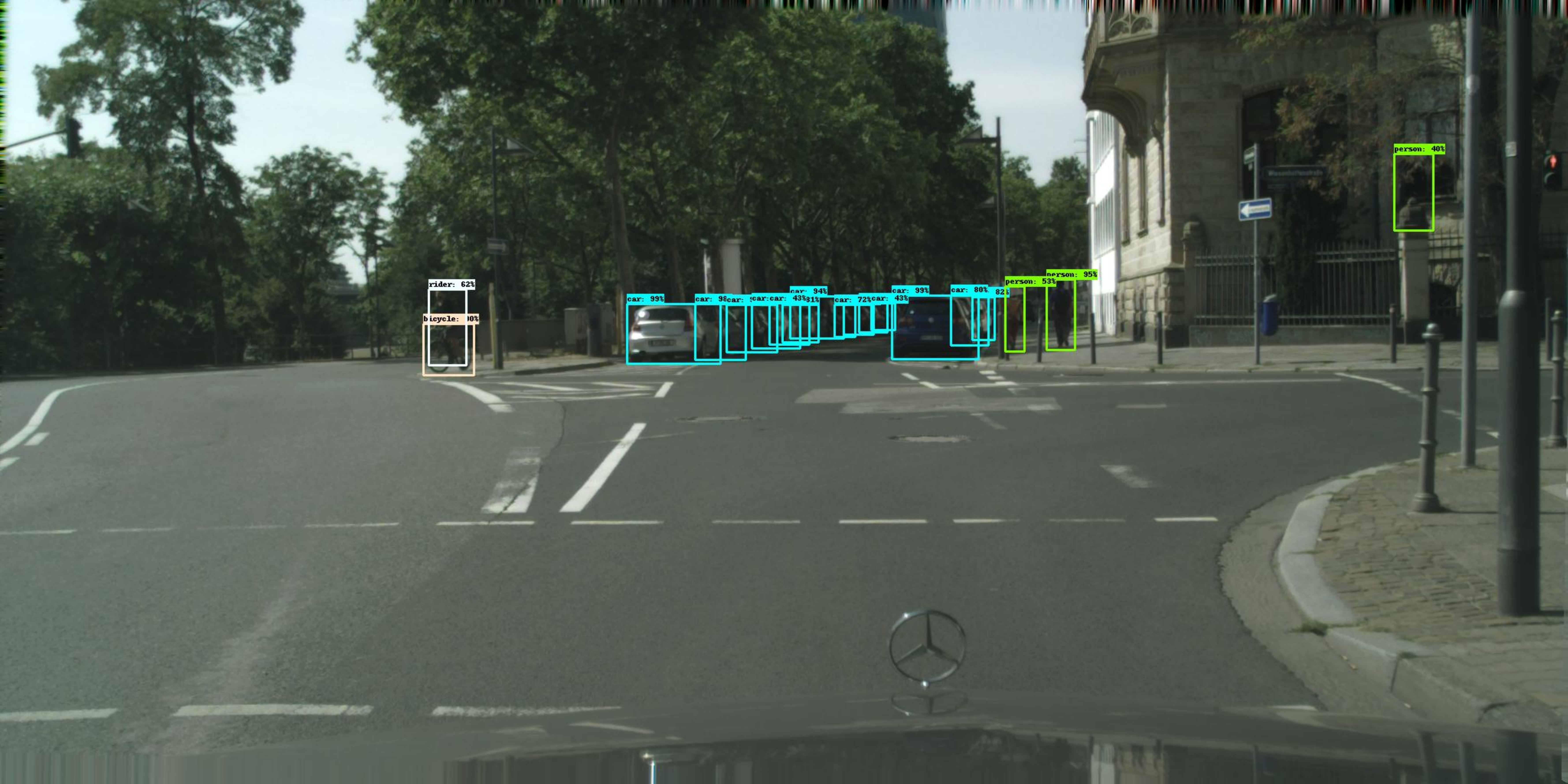}
                \caption*{Naive-Student}
        \end{subfigure}%
        
        \begin{subfigure}[b]{0.49\textwidth}
                \includegraphics[width=\linewidth]{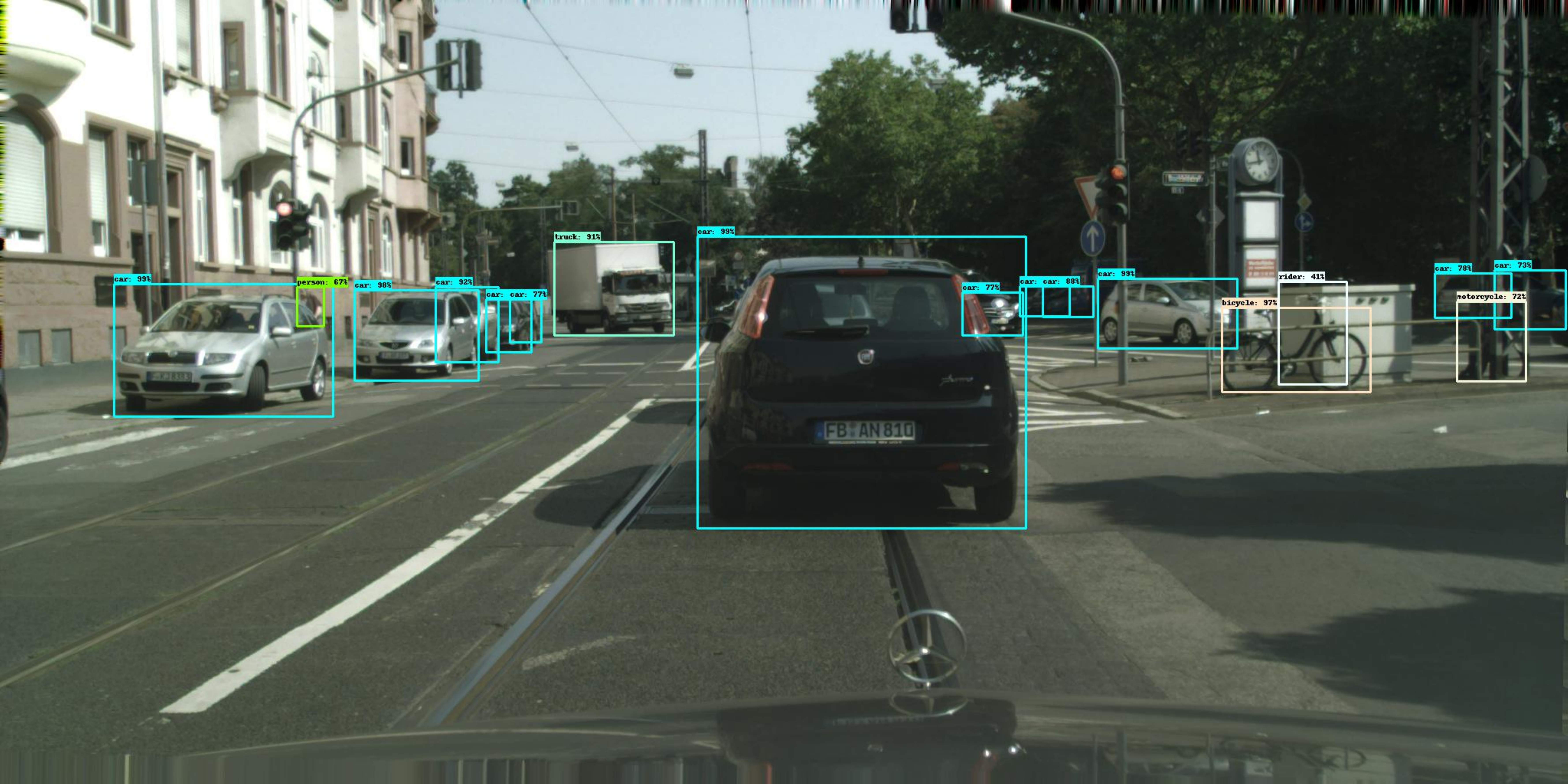}
                \caption*{VideoProp}
        \end{subfigure}%
        \rulesep
        \begin{subfigure}[b]{0.49\textwidth}
                \includegraphics[width=\linewidth]{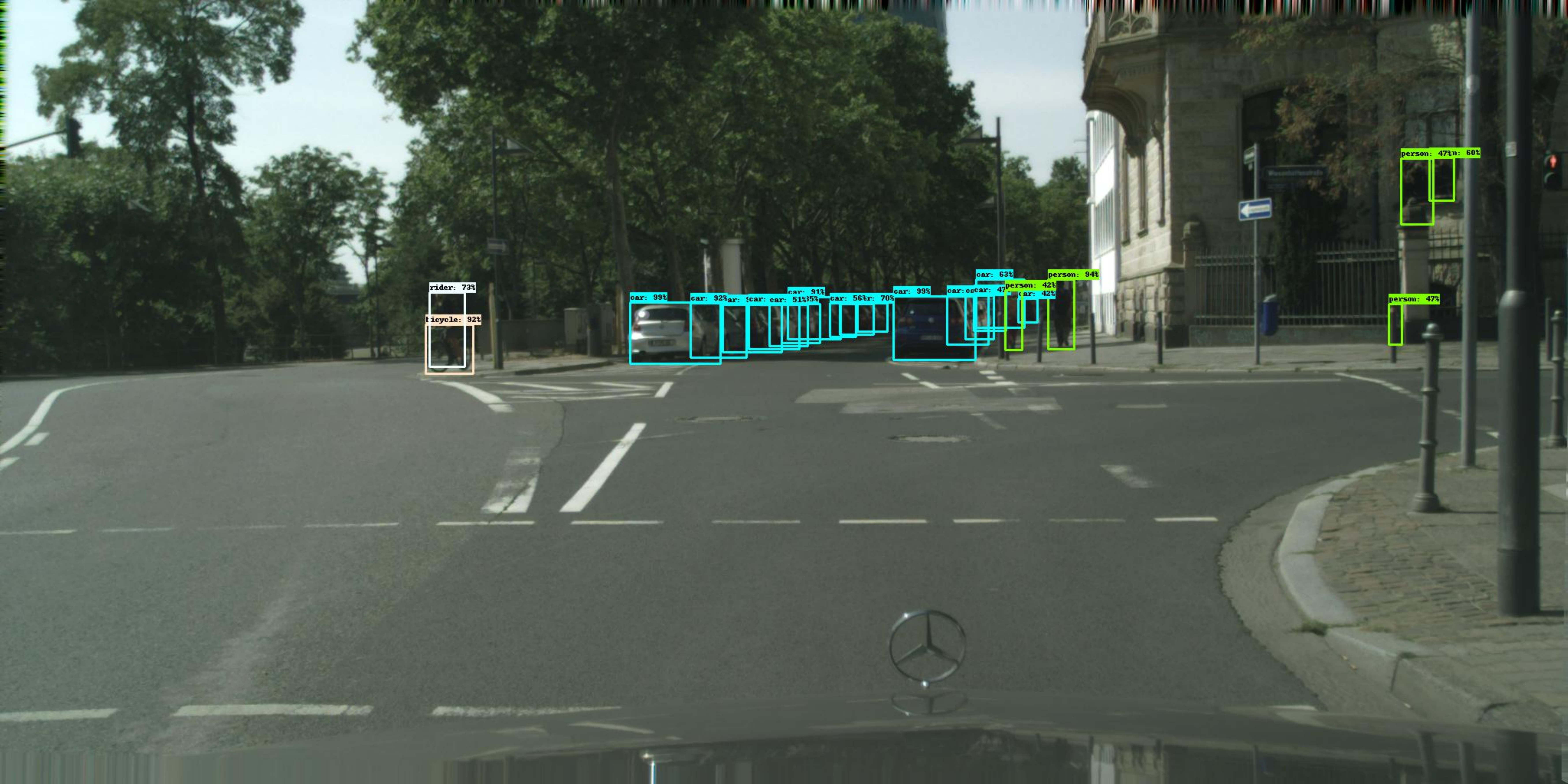}
                \caption*{VideoProp}
        \end{subfigure}%
        
        \begin{subfigure}[b]{0.49\textwidth}
                \includegraphics[width=\linewidth]{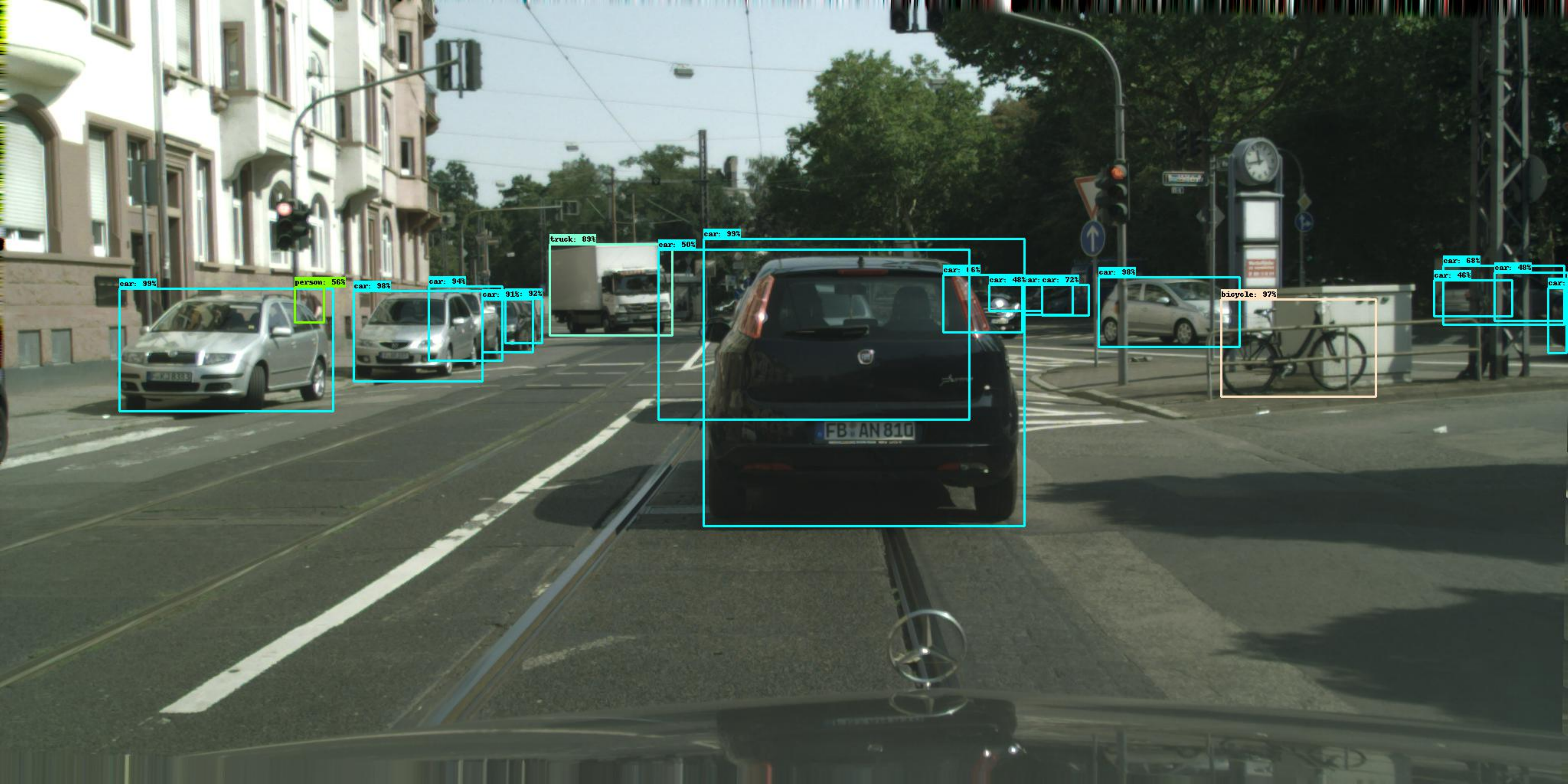}
                \caption*{PseudoProp}
        \end{subfigure}%
        \rulesep
        \begin{subfigure}[b]{0.49\textwidth}
                \includegraphics[width=\linewidth]{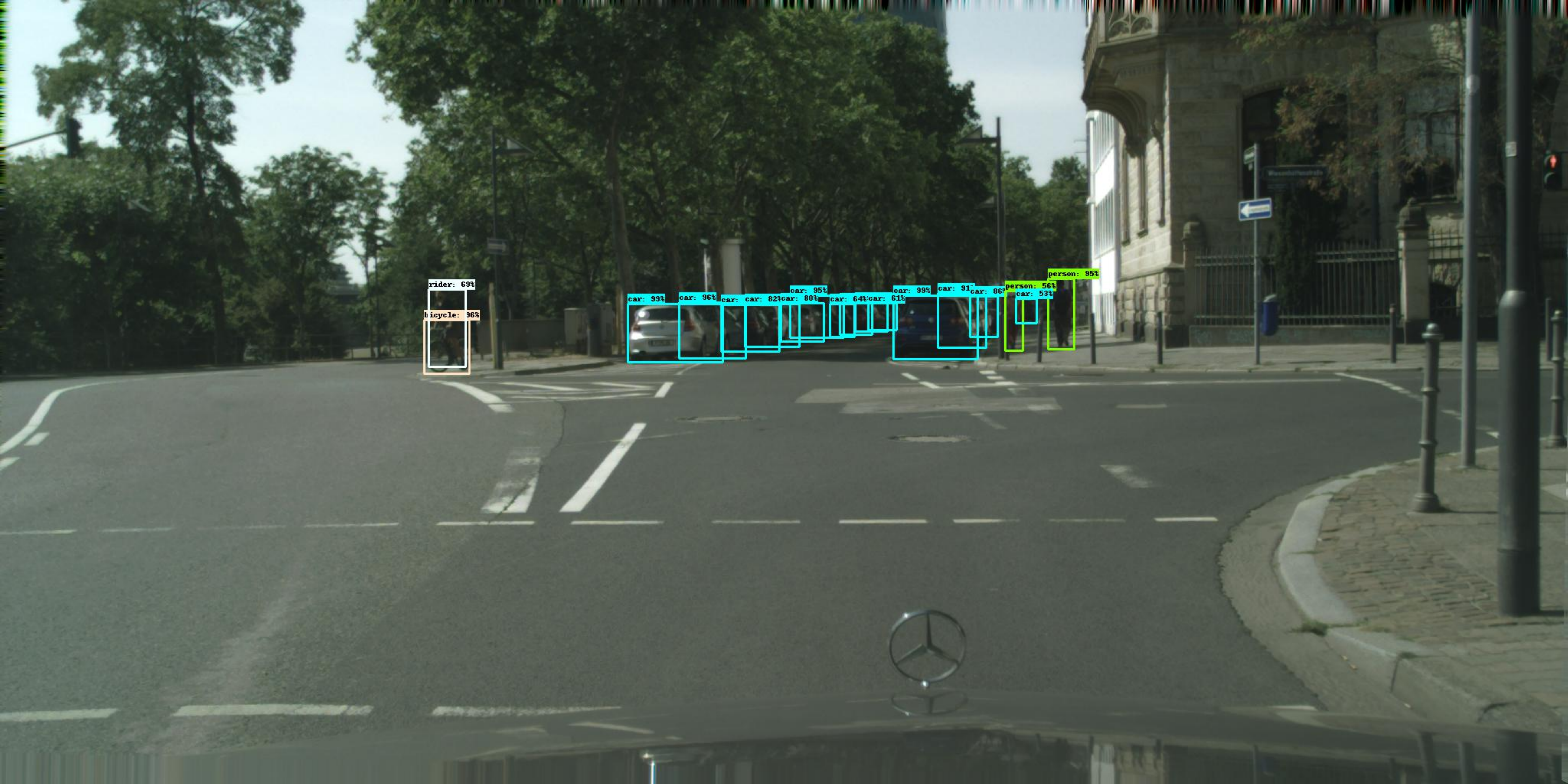}
                \caption*{PseudoProp}
        \end{subfigure}%
\caption{\small \it Visual comparison for the ground truth, Naive-Student, VideoProp, and our proposed PseudoProp on Cityscapes.}\label{fig: image_results1}
\end{figure*}

\begin{figure*}[t!]
\captionsetup[subfigure]{justification=centering}
\centering
        \begin{subfigure}[b]{0.49\textwidth}
                \includegraphics[width=\linewidth]{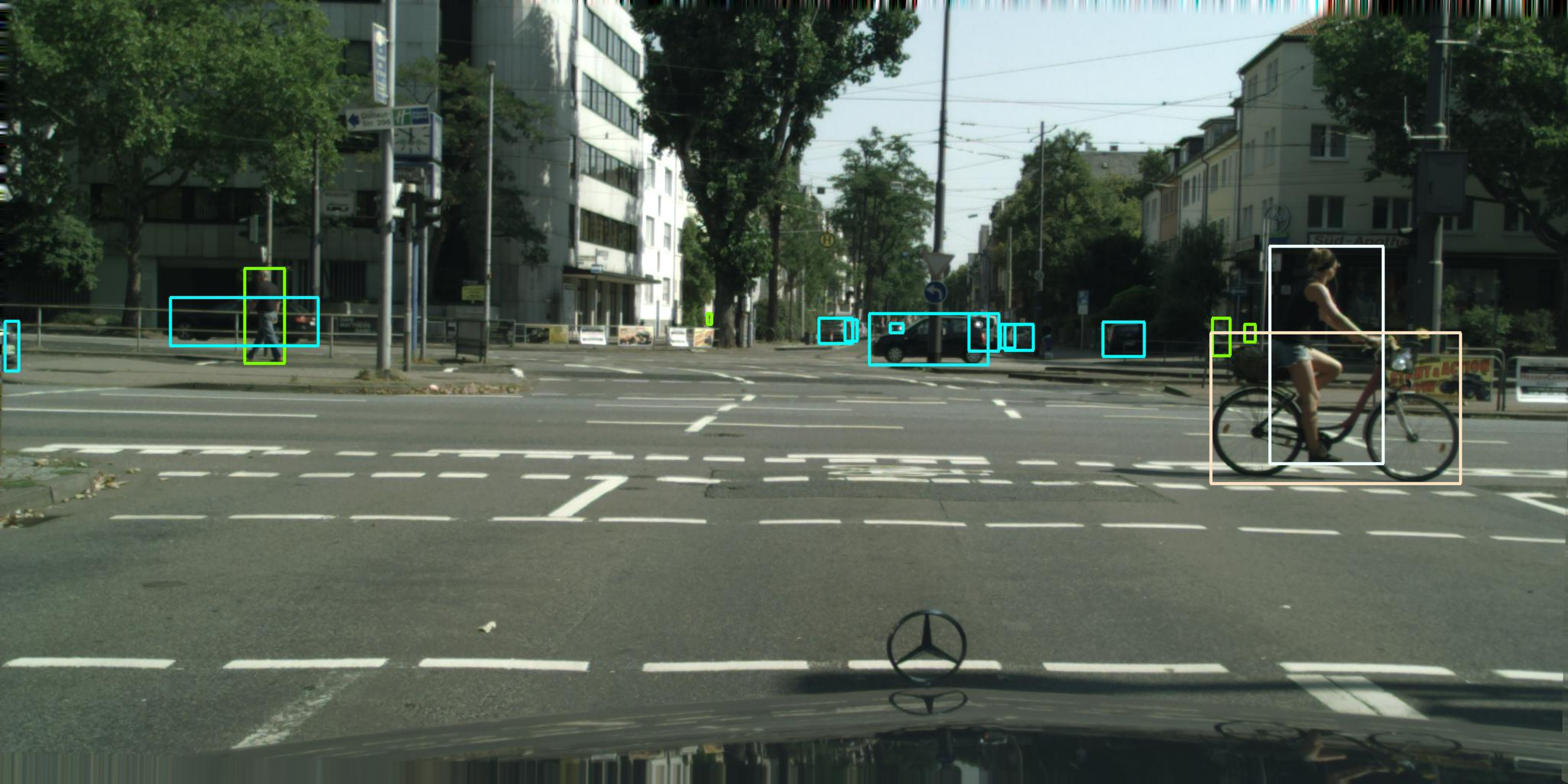}
                \caption*{Ground Truth}
        \end{subfigure}%
        \rulesep
        \begin{subfigure}[b]{0.49\textwidth}
                \includegraphics[width=\linewidth]{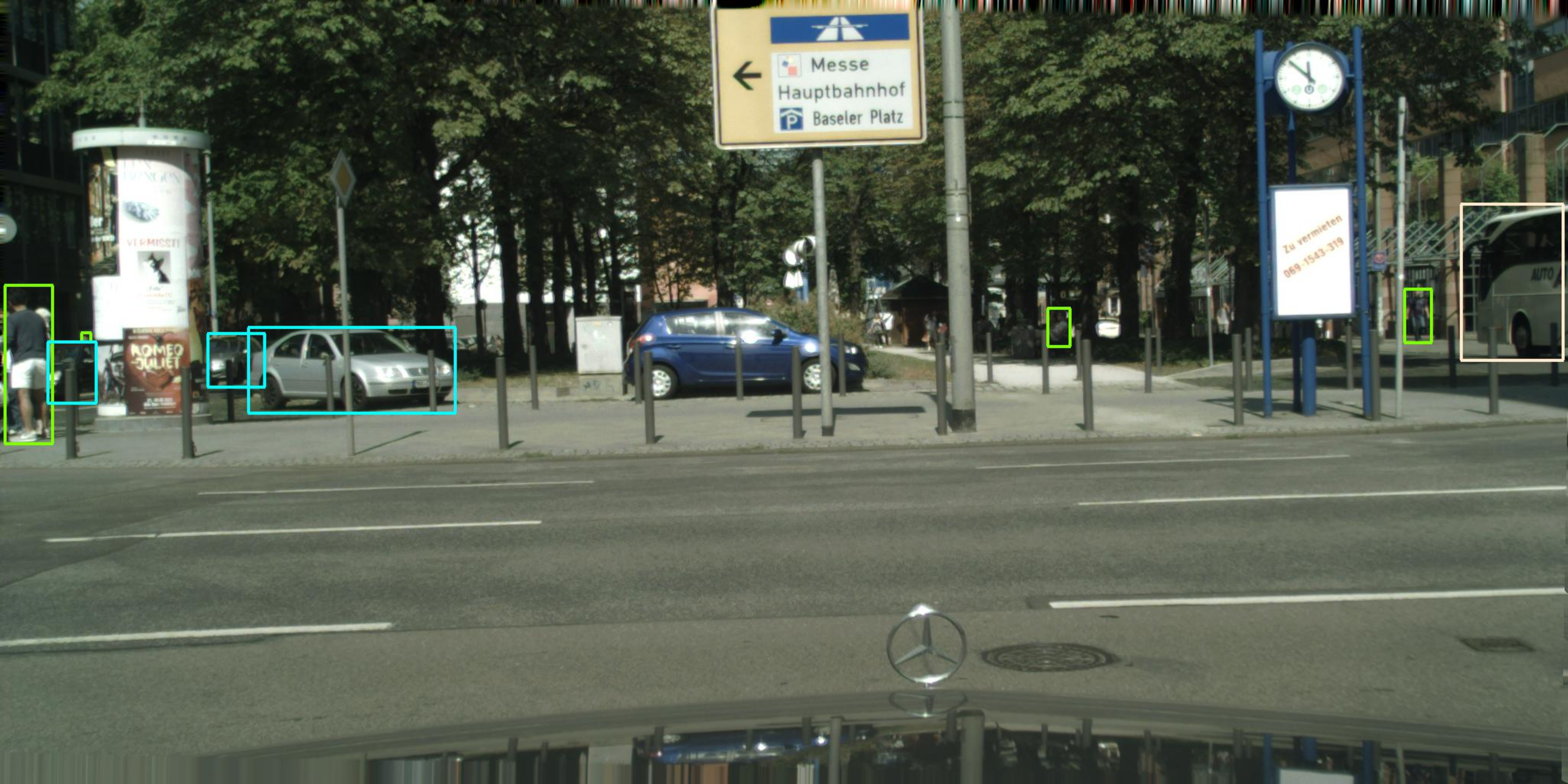}
                \caption*{Ground Truth}
        \end{subfigure}%
        
        \begin{subfigure}[b]{0.49\textwidth}
                \includegraphics[width=\linewidth]{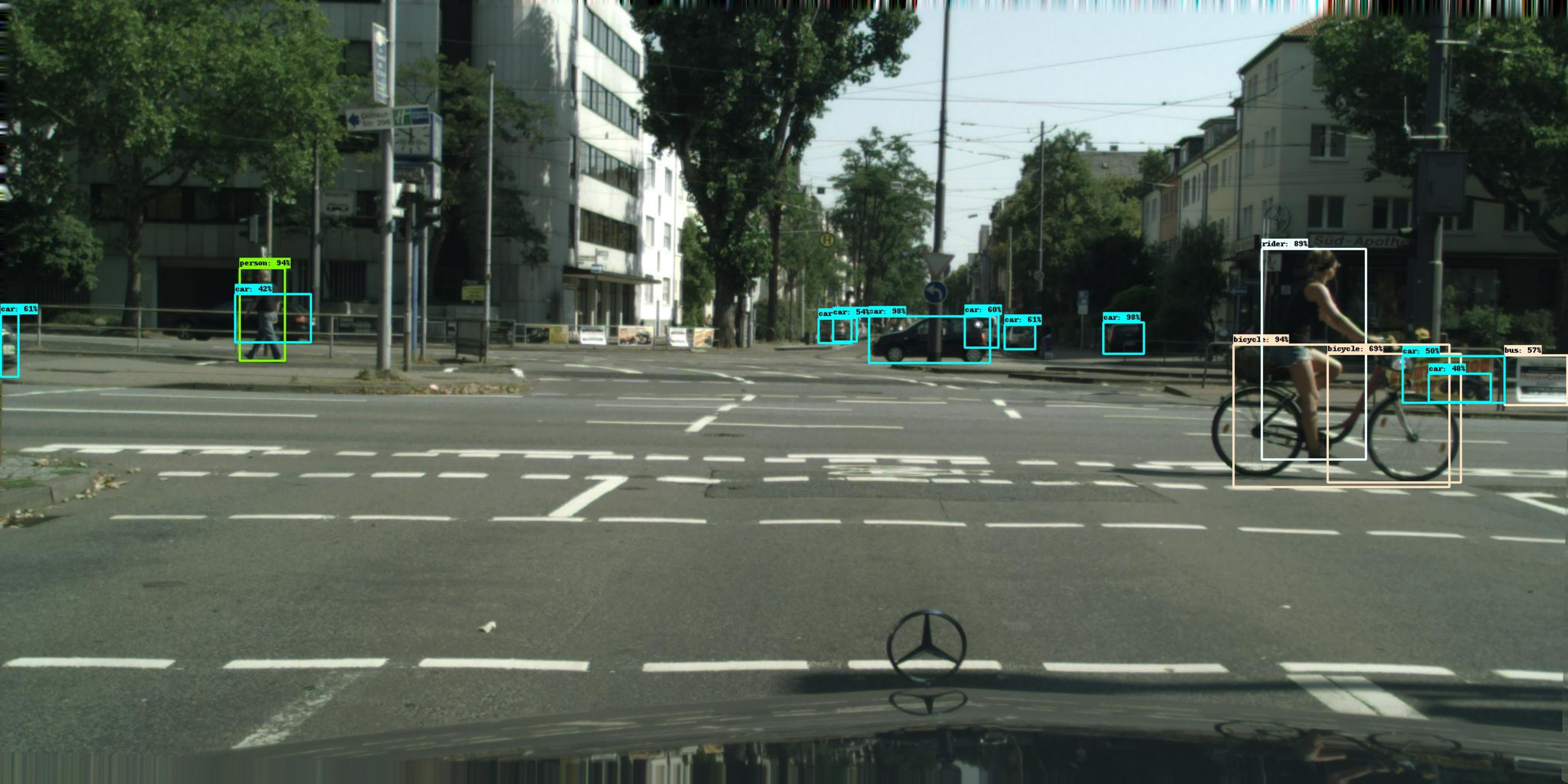}
                \caption*{Naive-Student}
        \end{subfigure}%
        \rulesep
        \begin{subfigure}[b]{0.49\textwidth}
                \includegraphics[width=\linewidth]{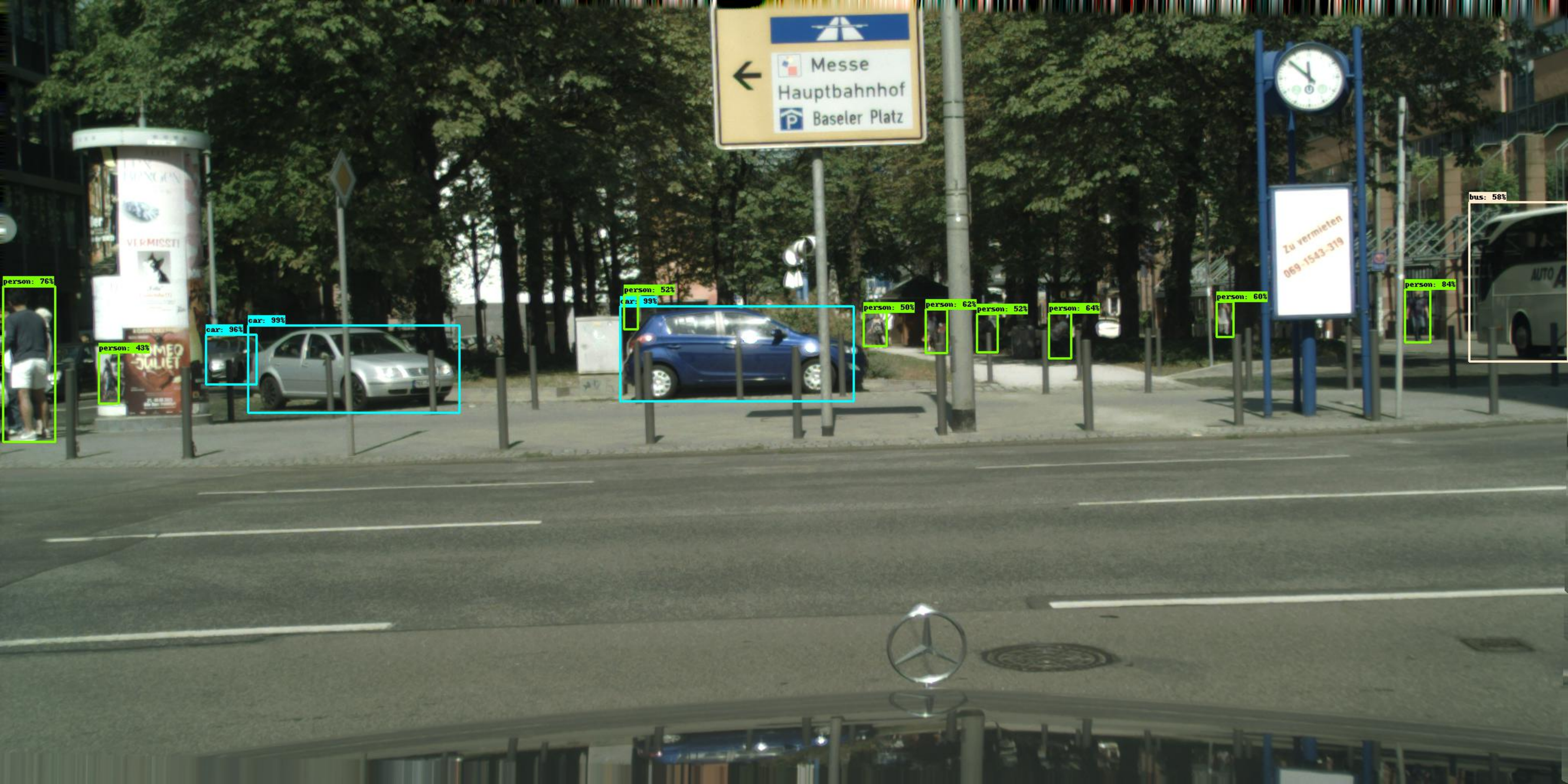}
                \caption*{Naive-Student}
        \end{subfigure}%
        
        \begin{subfigure}[b]{0.49\textwidth}
                \includegraphics[width=\linewidth]{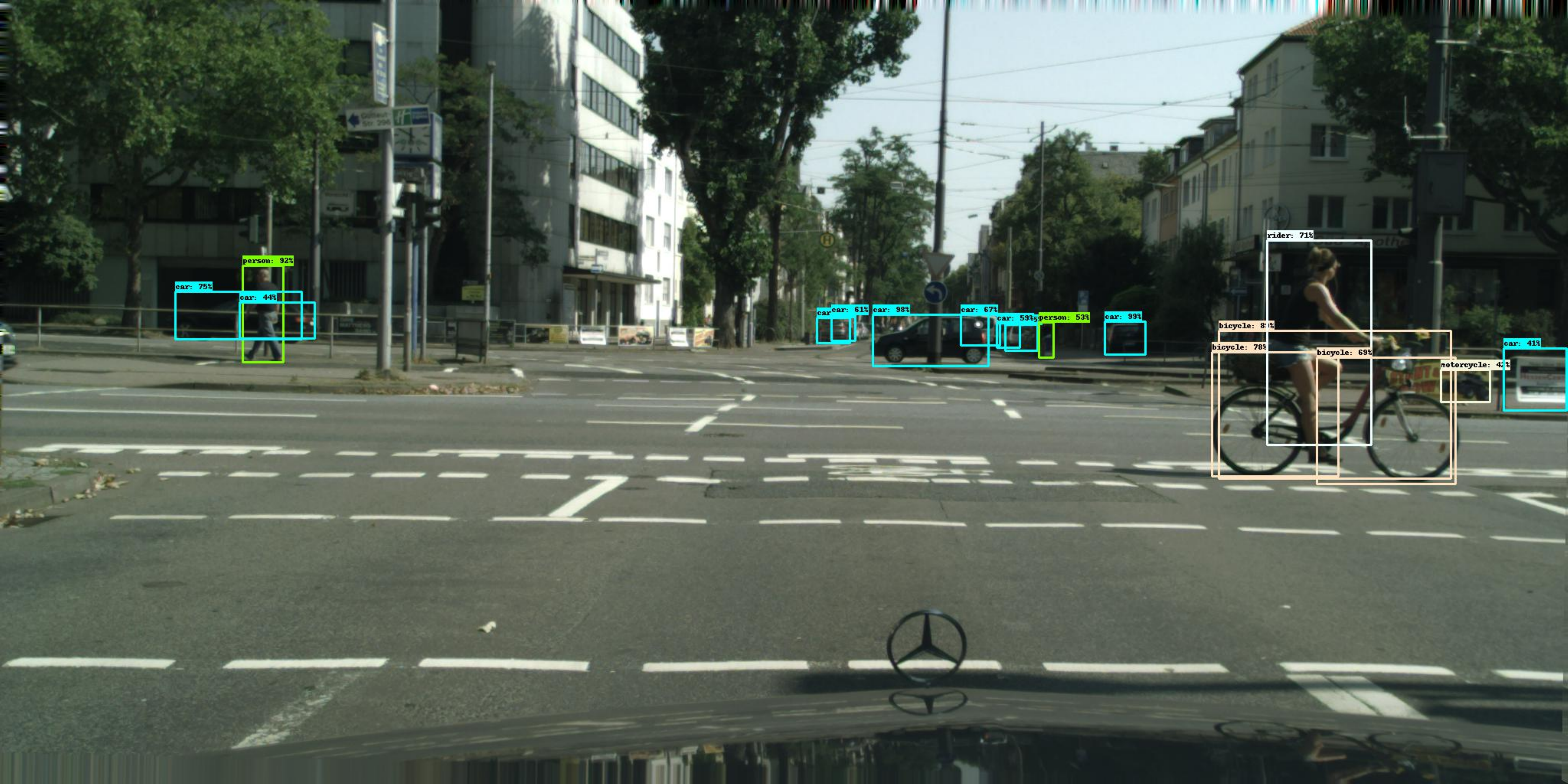}
                \caption*{VideoProp}
        \end{subfigure}%
        \rulesep
        \begin{subfigure}[b]{0.49\textwidth}
                \includegraphics[width=\linewidth]{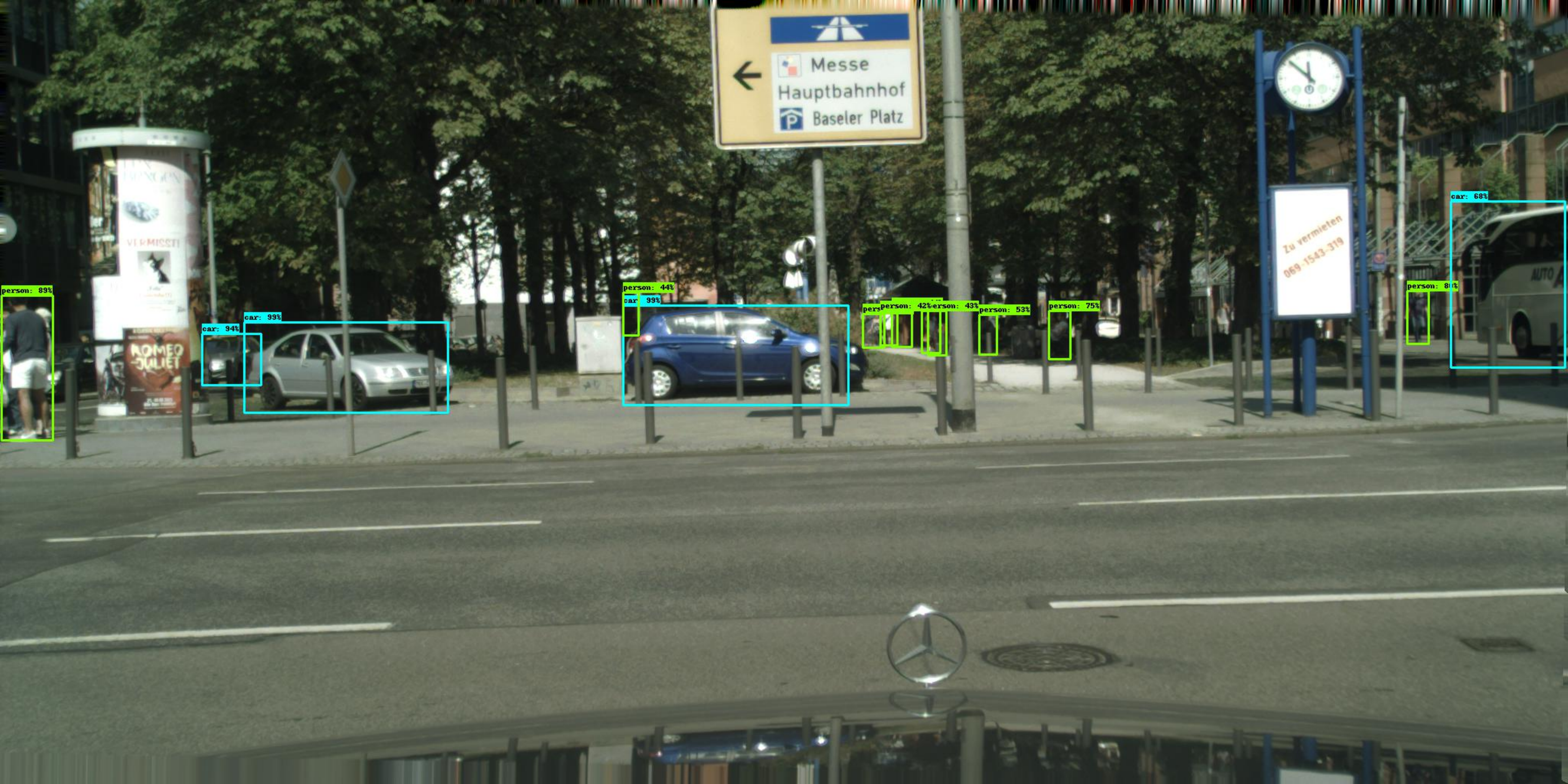}
                \caption*{VideoProp}
        \end{subfigure}%
        
        \begin{subfigure}[b]{0.49\textwidth}
                \includegraphics[width=\linewidth]{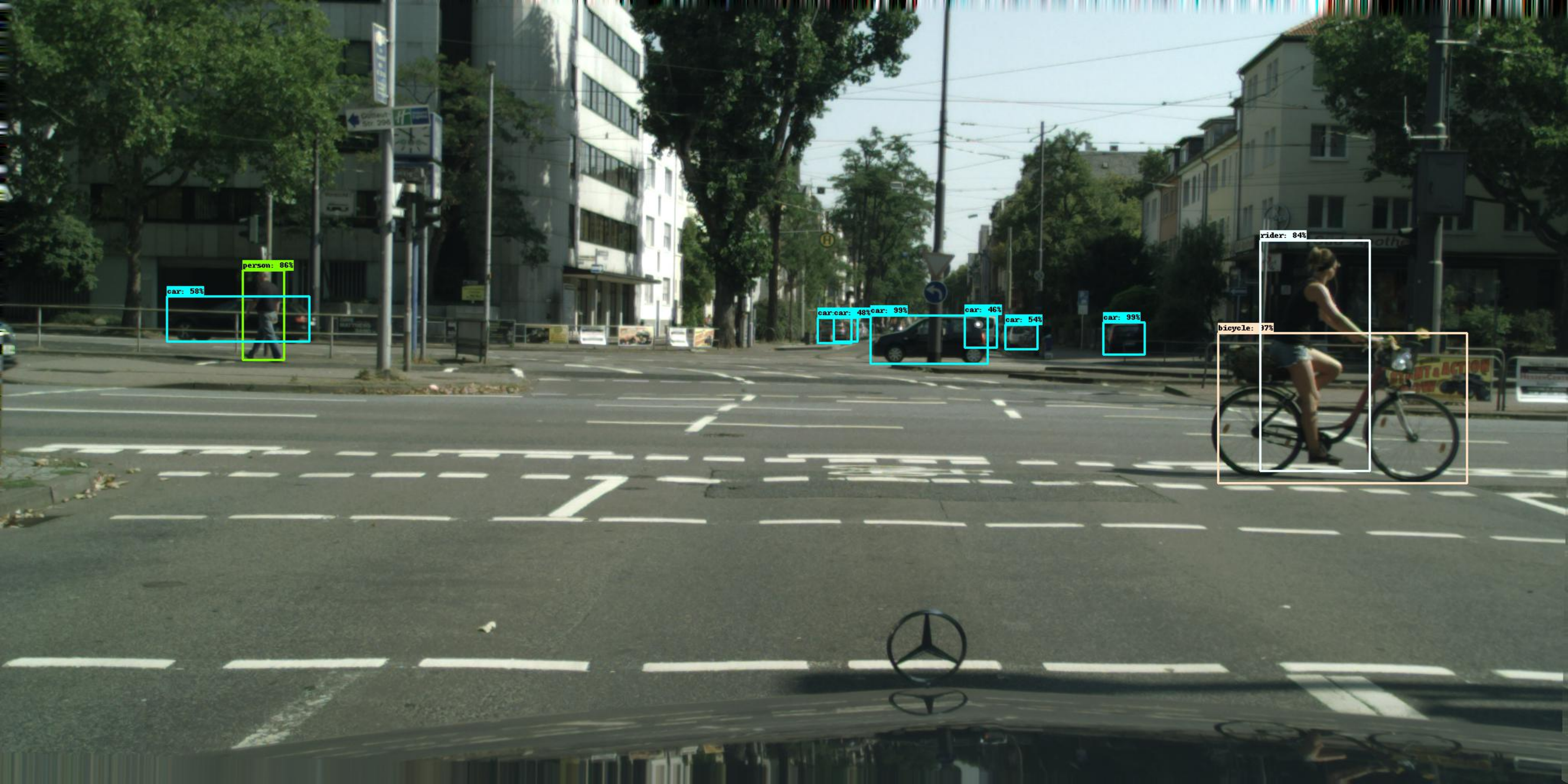}
                \caption*{PseudoProp}
        \end{subfigure}%
        \rulesep
        \begin{subfigure}[b]{0.49\textwidth}
                \includegraphics[width=\linewidth]{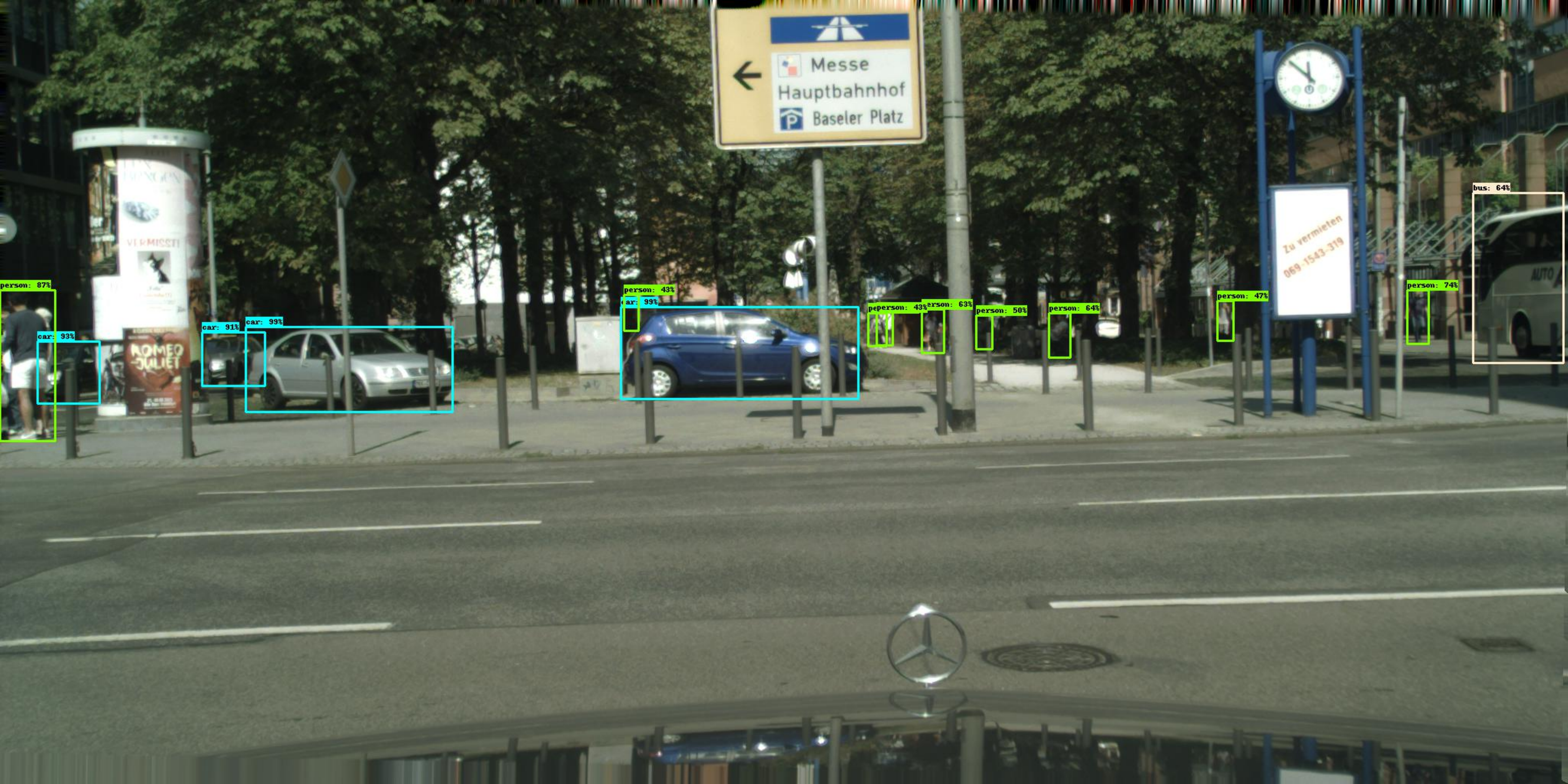}
                \caption*{PseudoProp}
        \end{subfigure}%
\caption{\small \it Visual comparison for the ground truth, Naive-Student, VideoProp, and our proposed PseudoProp on Cityscapes.}\label{fig: image_results2}
\end{figure*}

\begin{figure*}[t!]
\captionsetup[subfigure]{justification=centering}
\centering
        \begin{subfigure}[b]{0.49\textwidth}
                \includegraphics[width=\linewidth]{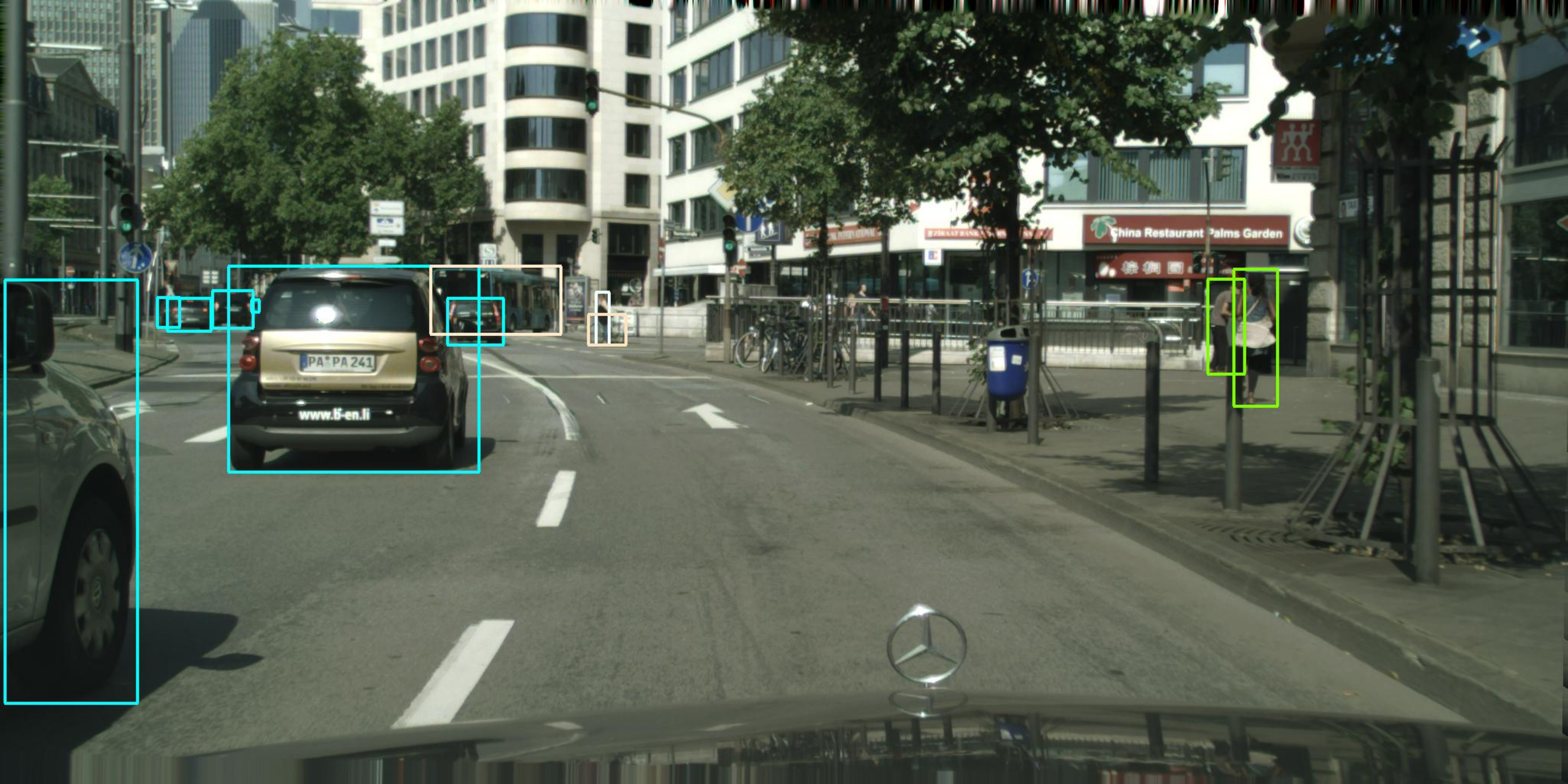}
                \caption*{Ground Truth}
        \end{subfigure}%
        \rulesep
        \begin{subfigure}[b]{0.49\textwidth}
                \includegraphics[width=\linewidth]{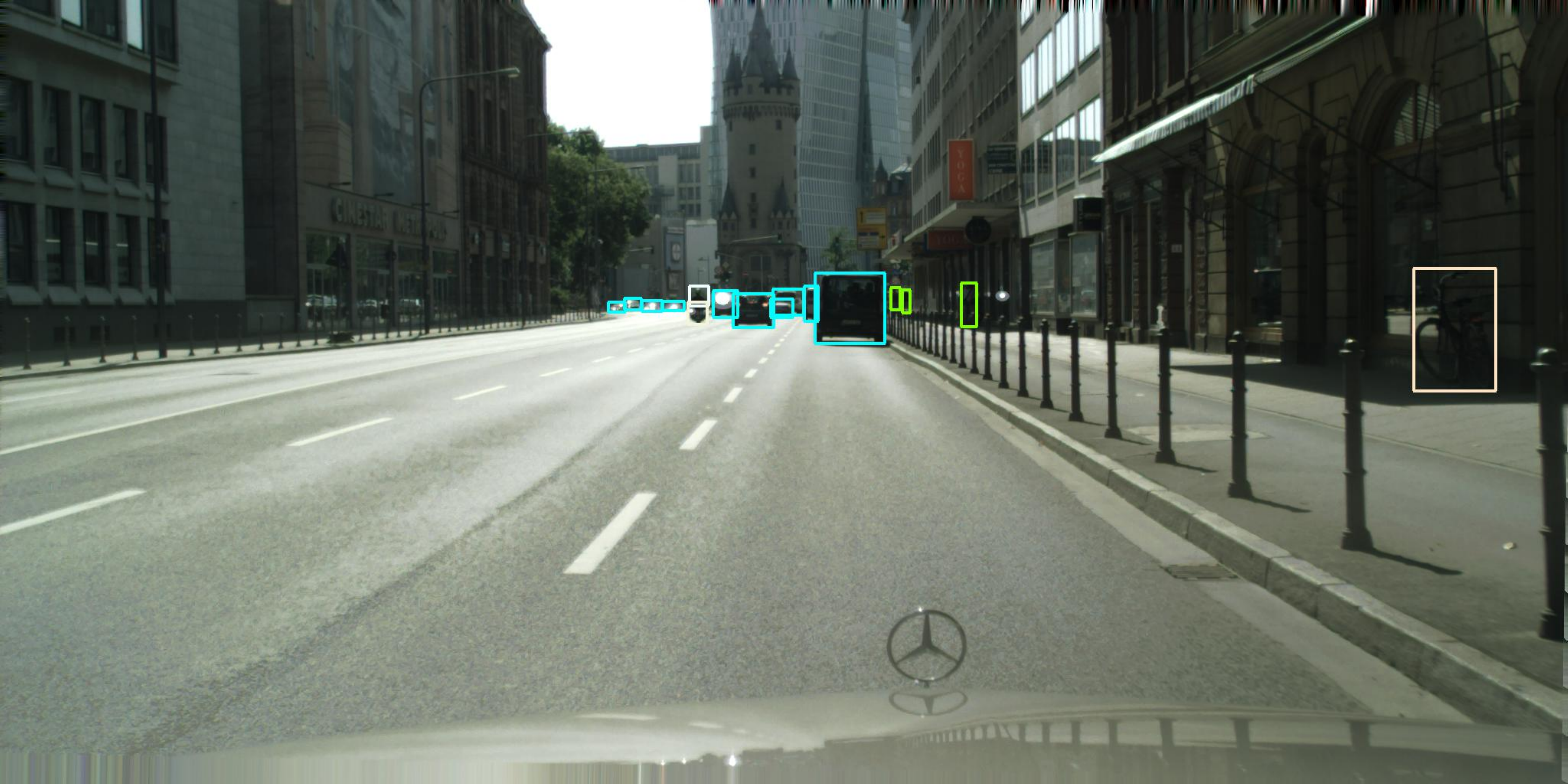}
                \caption*{Ground Truth}
        \end{subfigure}%
        
        \begin{subfigure}[b]{0.49\textwidth}
                \includegraphics[width=\linewidth]{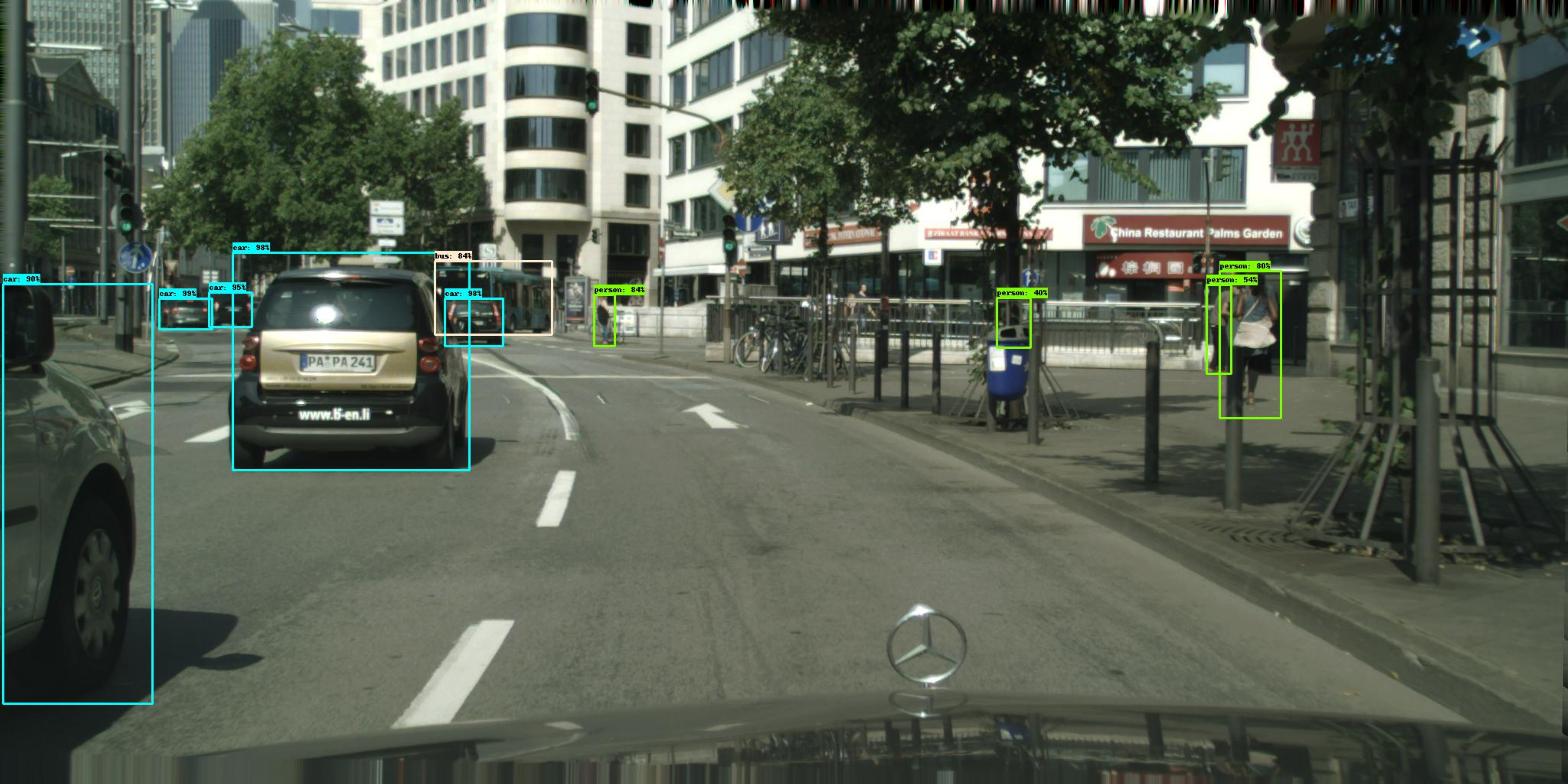}
                \caption*{Naive-Student}
        \end{subfigure}%
        \rulesep
        \begin{subfigure}[b]{0.49\textwidth}
                \includegraphics[width=\linewidth]{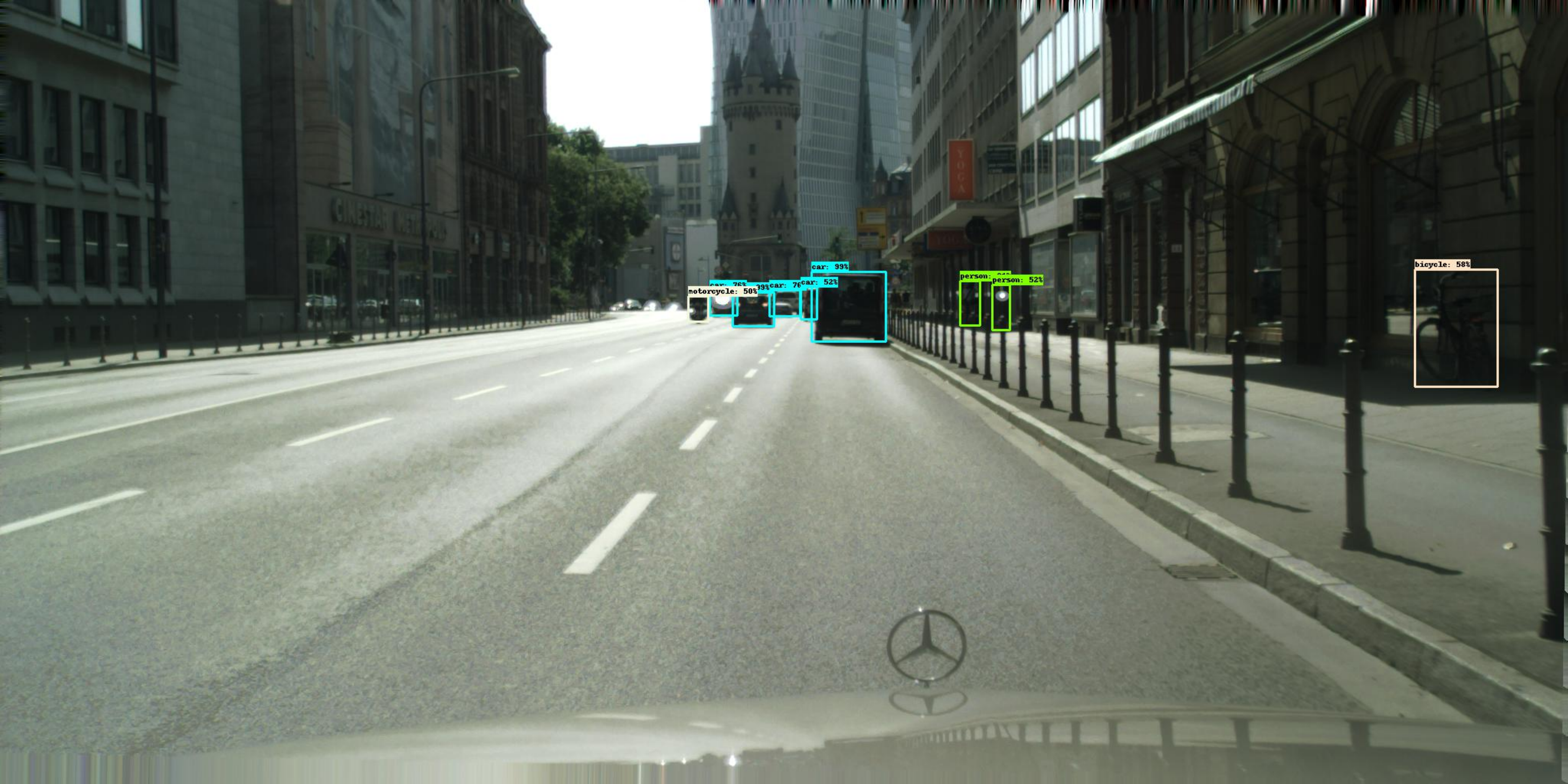}
                \caption*{Naive-Student}
        \end{subfigure}%
        
        \begin{subfigure}[b]{0.49\textwidth}
                \includegraphics[width=\linewidth]{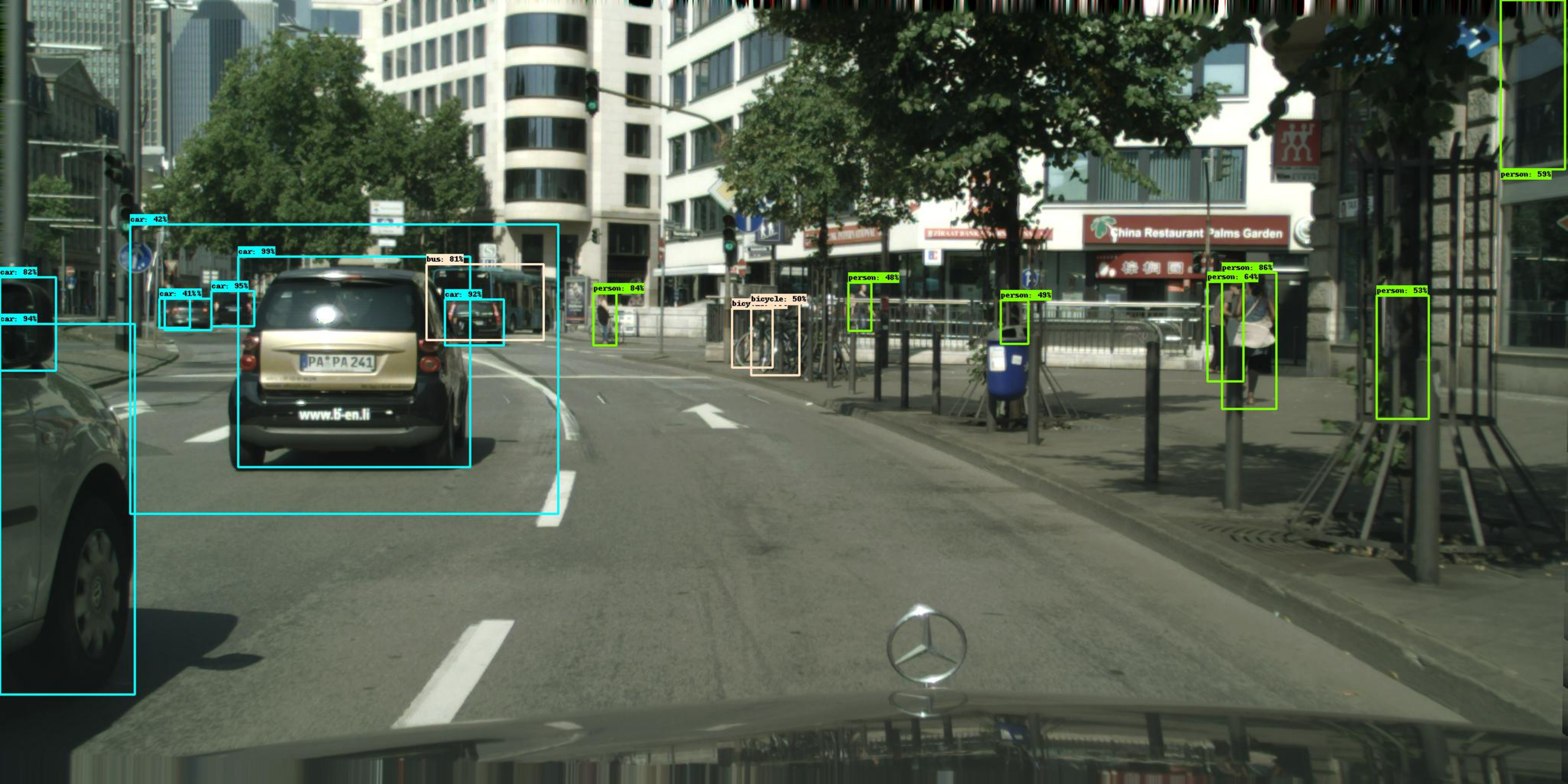}
                \caption*{VideoProp}
        \end{subfigure}%
        \rulesep
        \begin{subfigure}[b]{0.49\textwidth}
                \includegraphics[width=\linewidth]{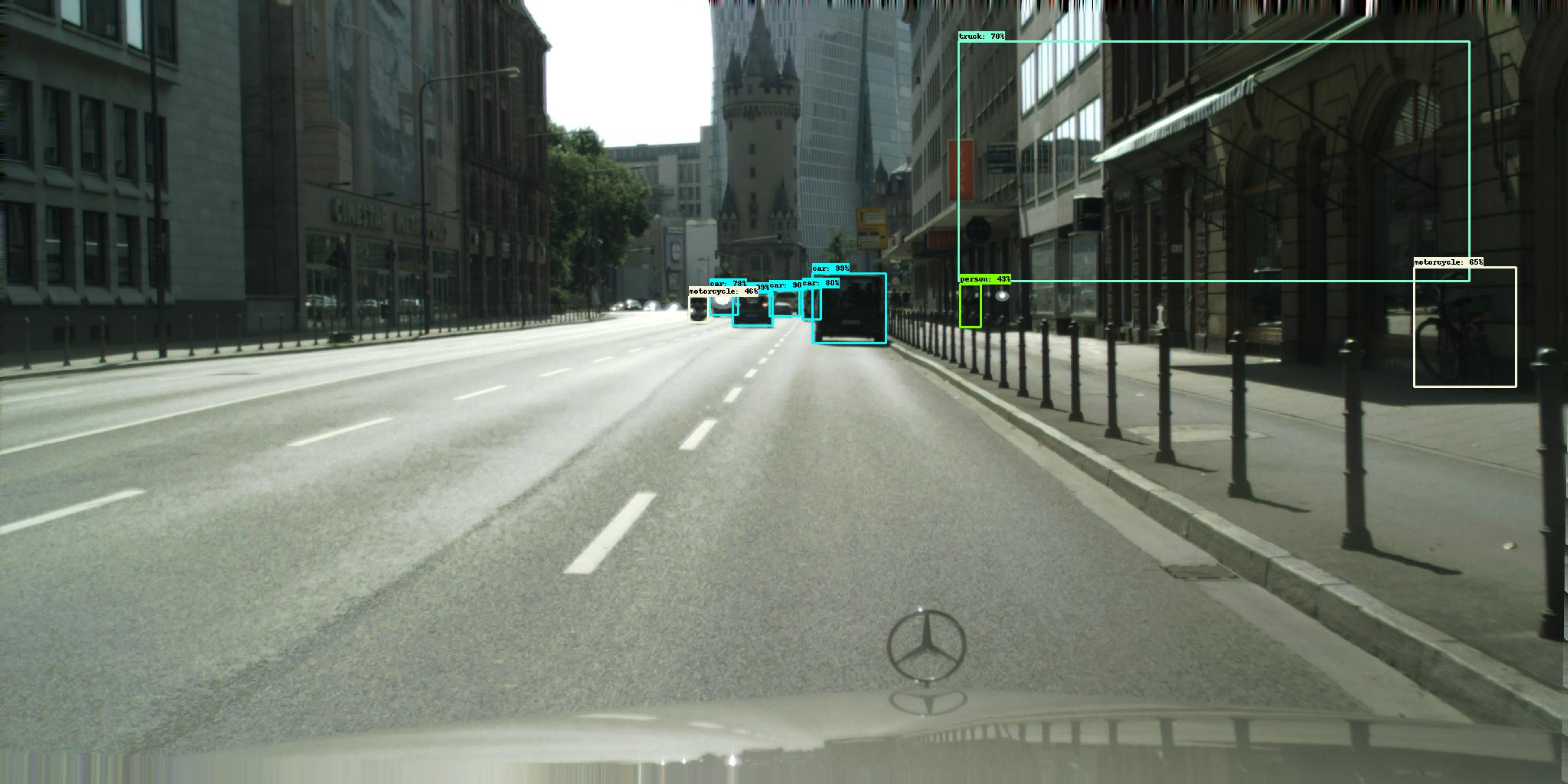}
                \caption*{VideoProp}
        \end{subfigure}%
        
        \begin{subfigure}[b]{0.49\textwidth}
                \includegraphics[width=\linewidth]{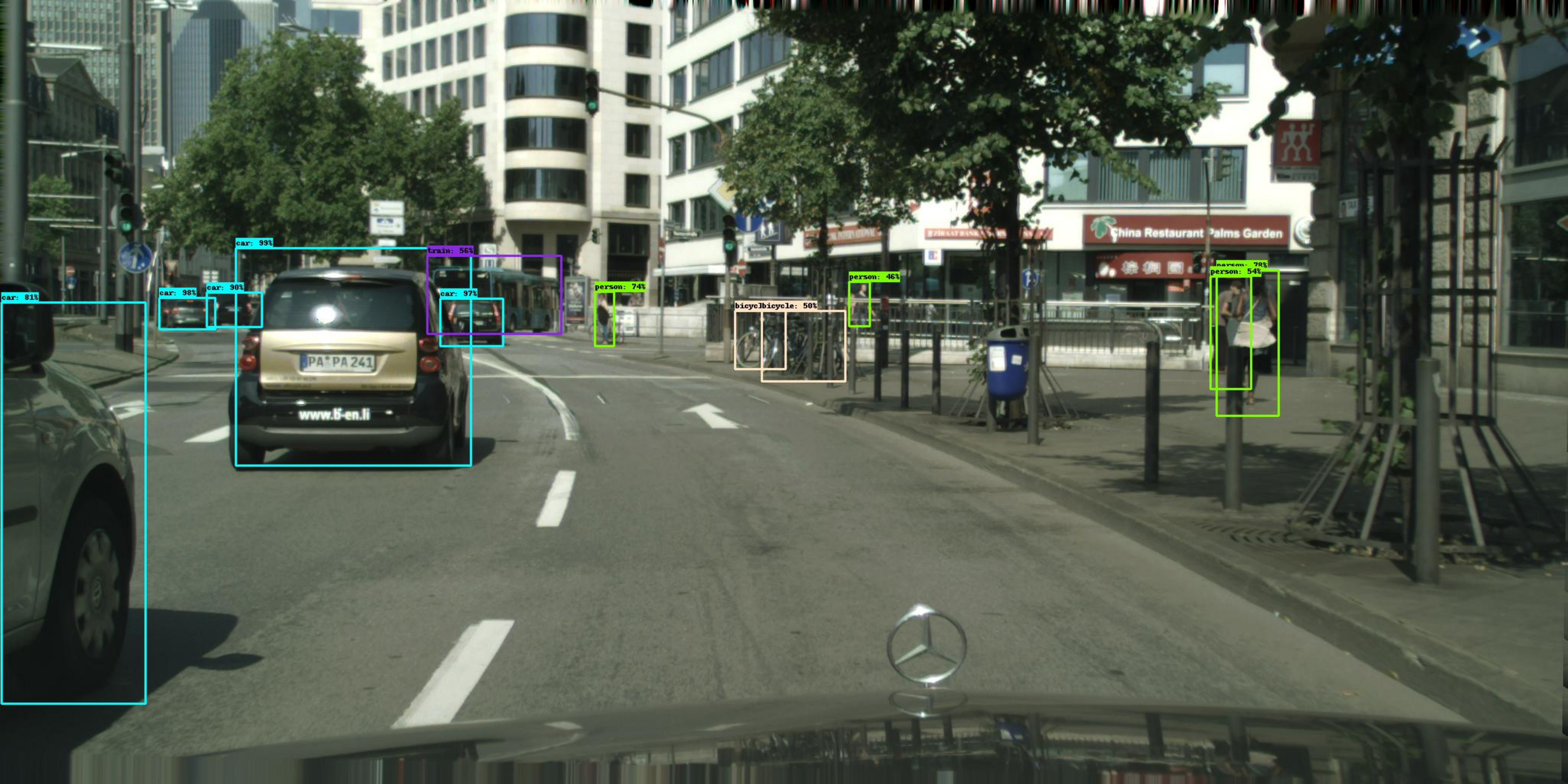}
                \caption*{PseudoProp}
        \end{subfigure}%
        \rulesep
        \begin{subfigure}[b]{0.49\textwidth}
                \includegraphics[width=\linewidth]{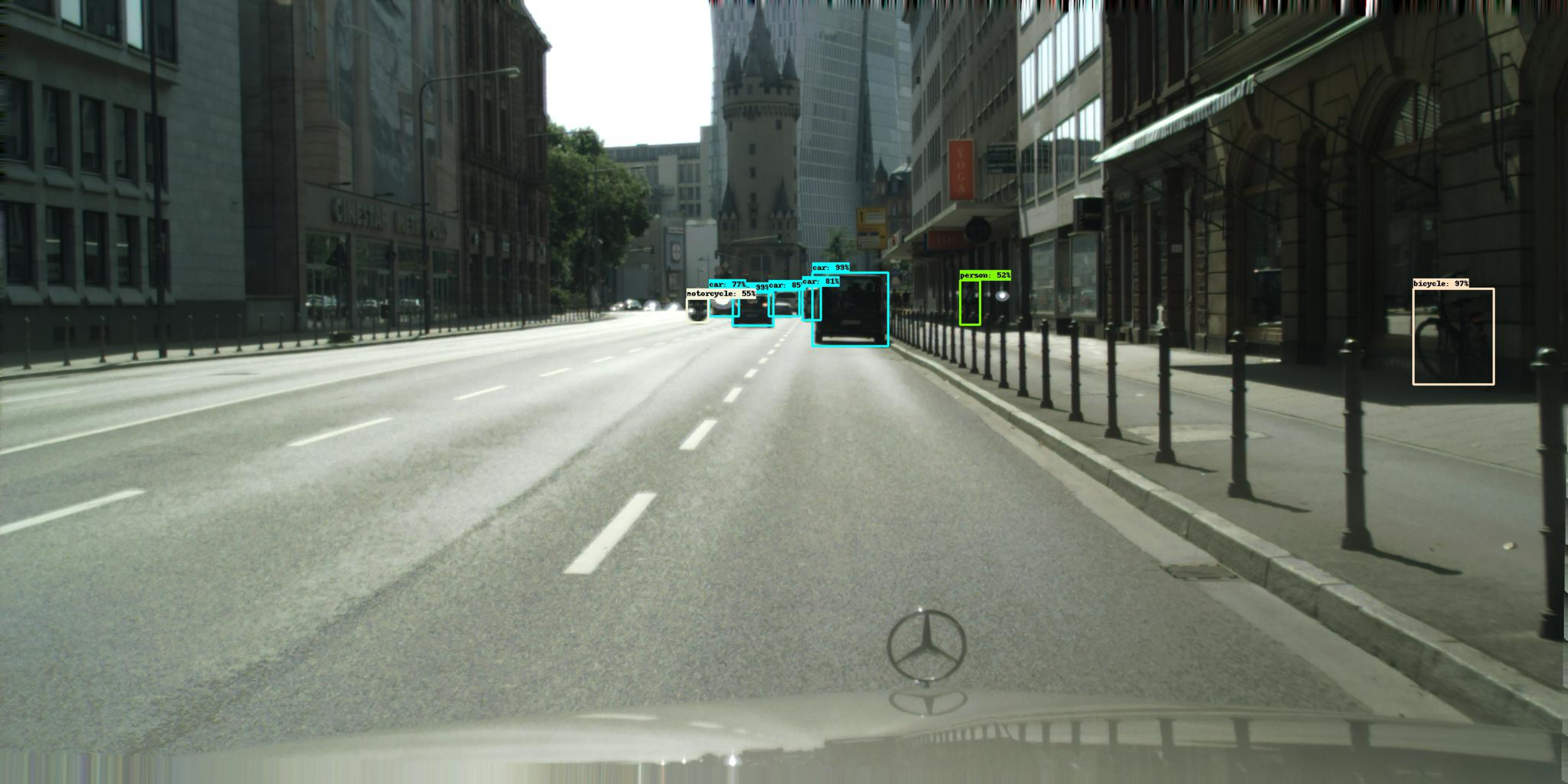}
                \caption*{PseudoProp}
        \end{subfigure}%
\caption{\small \it Visual comparison for the ground truth, Naive-Student, VideoProp, and our proposed PseudoProp on Cityscapes.}\label{fig: image_results3}
\end{figure*}

\begin{figure*}[t!]
\captionsetup[subfigure]{justification=centering}
\centering
        \begin{subfigure}[b]{0.49\textwidth}
                \includegraphics[width=\linewidth]{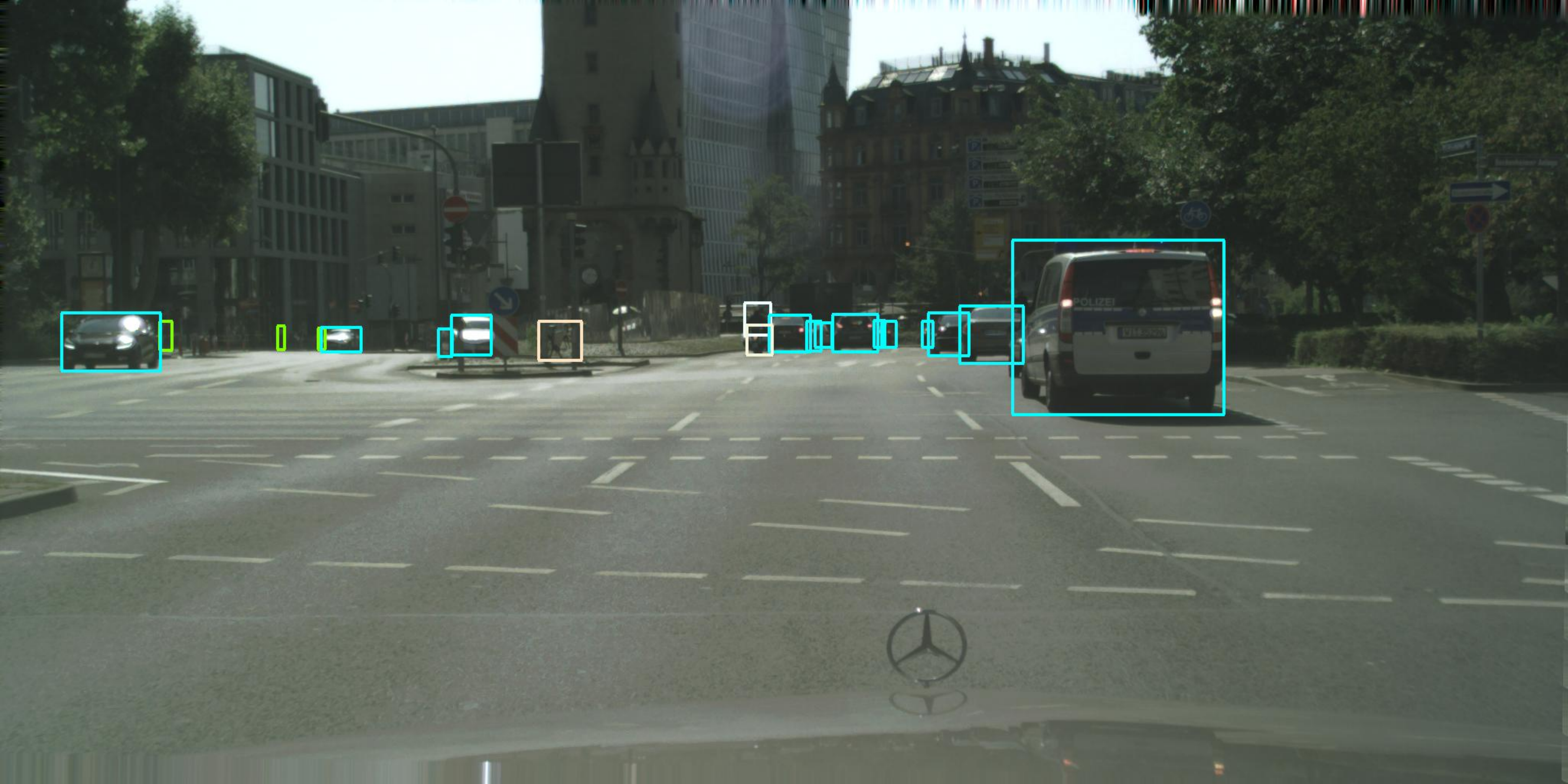}
                \caption*{Ground Truth}
        \end{subfigure}%
        \rulesep
        \begin{subfigure}[b]{0.49\textwidth}
                \includegraphics[width=\linewidth]{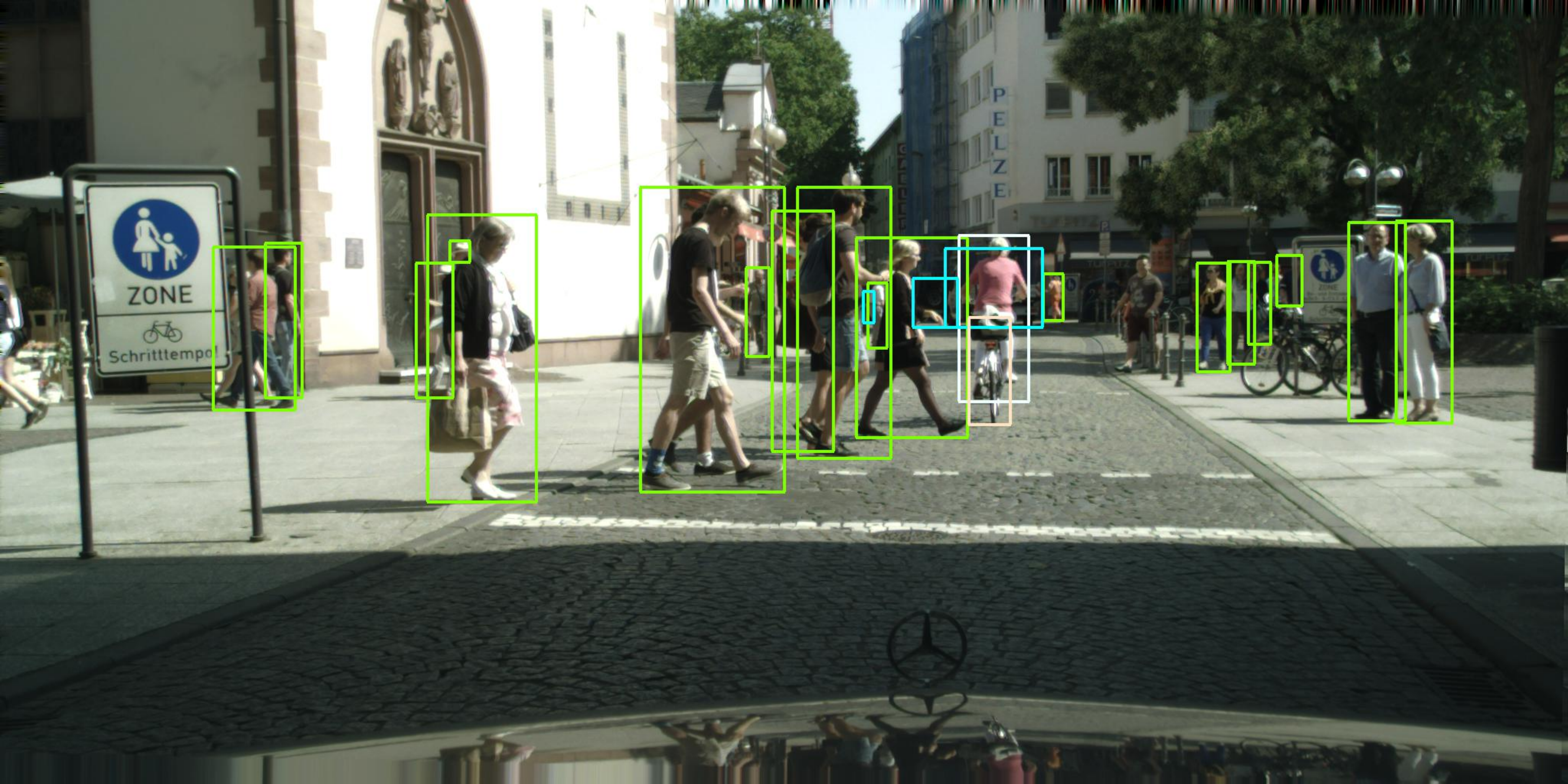}
                \caption*{Ground Truth}
        \end{subfigure}%
        
        \begin{subfigure}[b]{0.49\textwidth}
                \includegraphics[width=\linewidth]{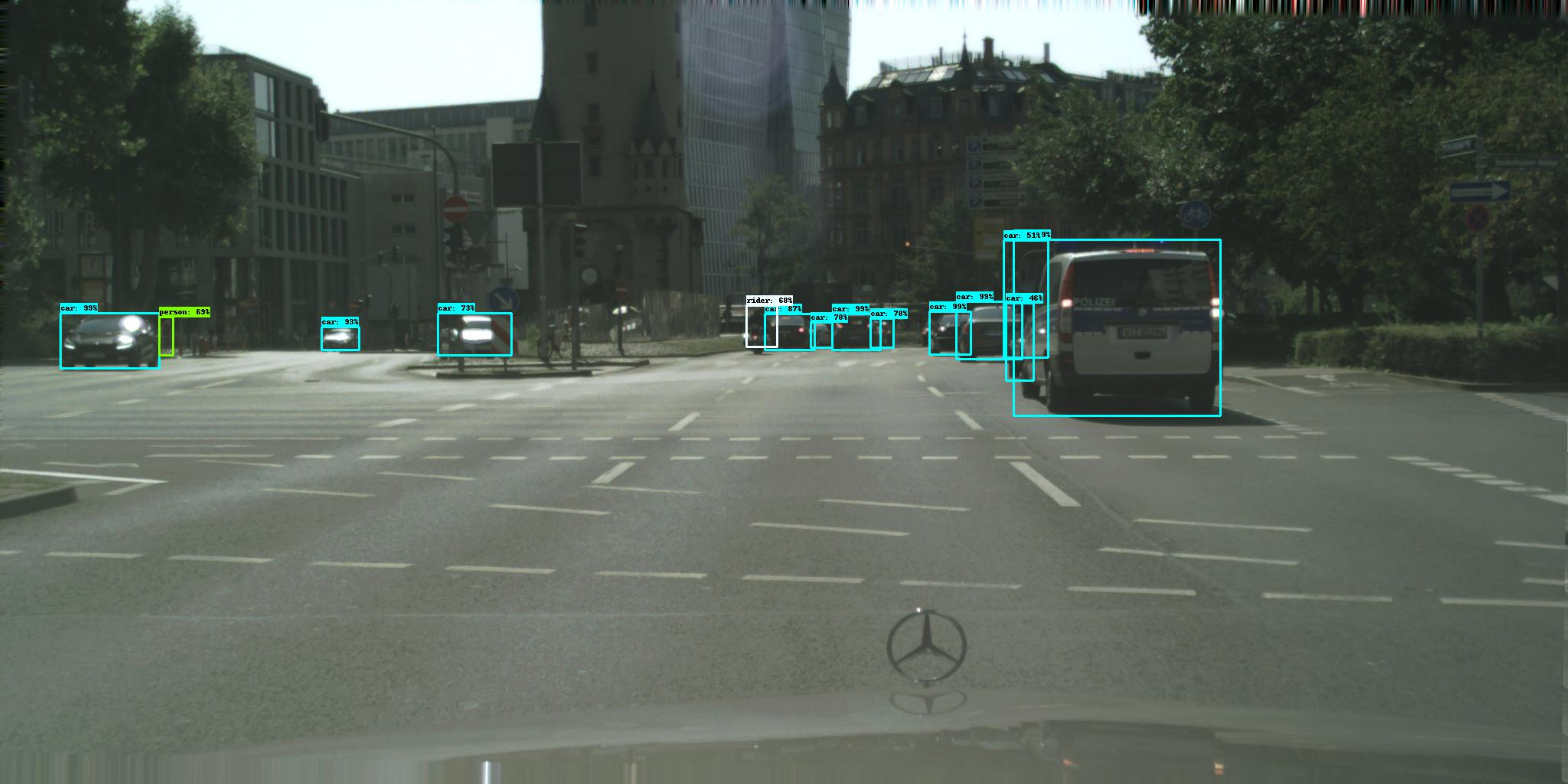}
                \caption*{Naive-Student}
        \end{subfigure}%
        \rulesep
        \begin{subfigure}[b]{0.49\textwidth}
                \includegraphics[width=\linewidth]{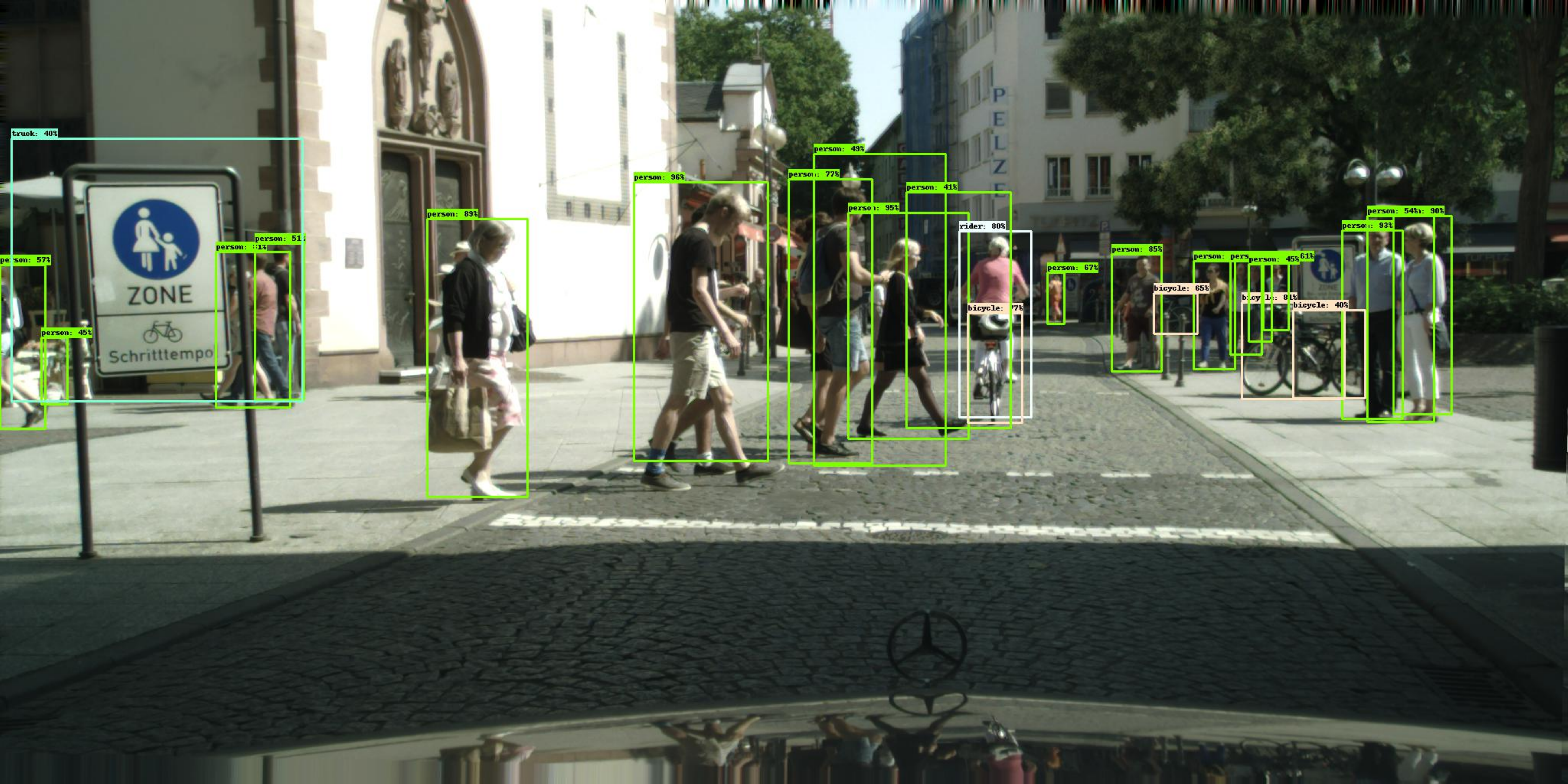}
                \caption*{Naive-Student}
        \end{subfigure}%
        
        \begin{subfigure}[b]{0.49\textwidth}
                \includegraphics[width=\linewidth]{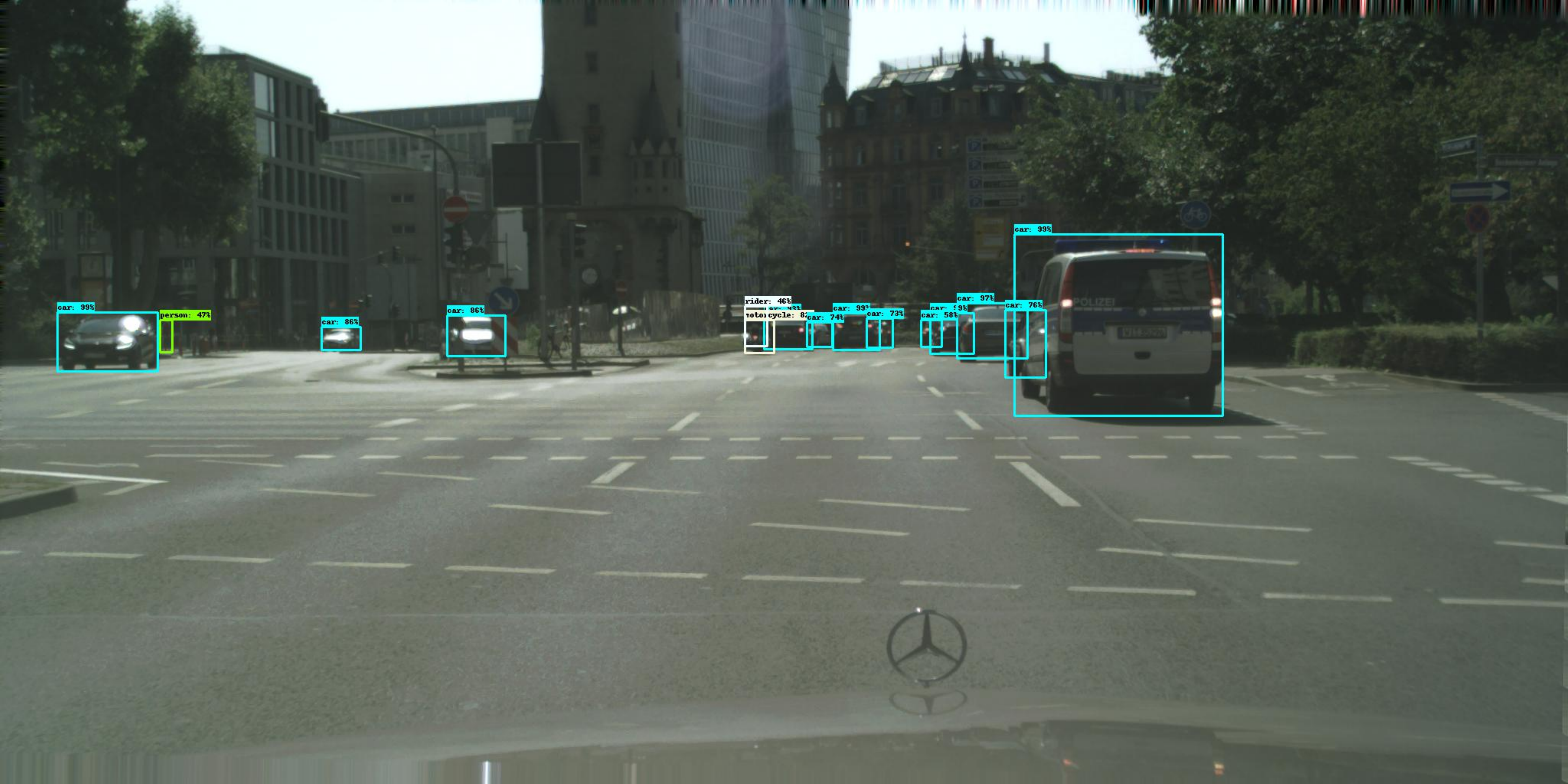}
                \caption*{VideoProp}
        \end{subfigure}%
        \rulesep
        \begin{subfigure}[b]{0.49\textwidth}
                \includegraphics[width=\linewidth]{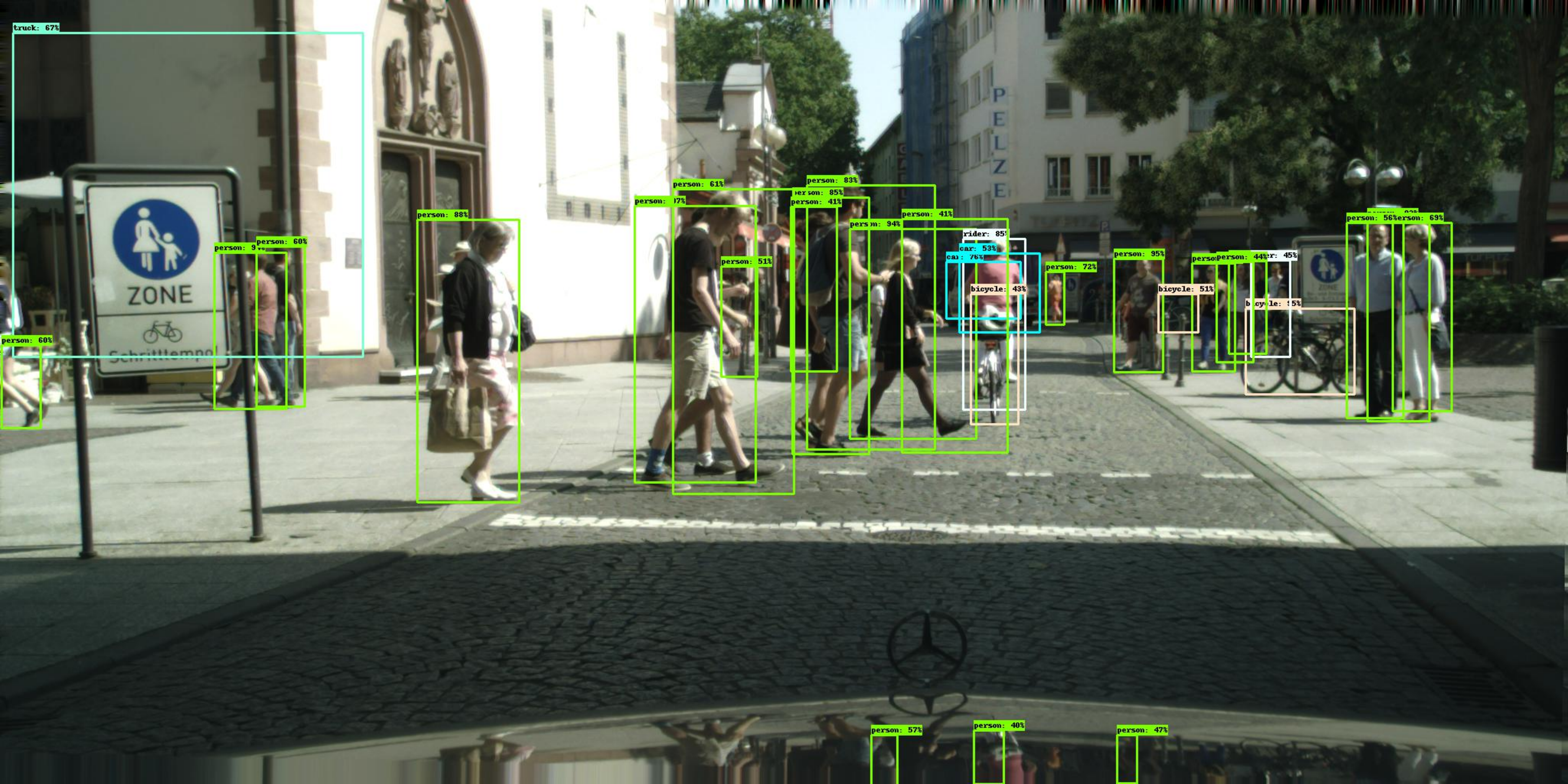}
                \caption*{VideoProp}
        \end{subfigure}%
        
        \begin{subfigure}[b]{0.49\textwidth}
                \includegraphics[width=\linewidth]{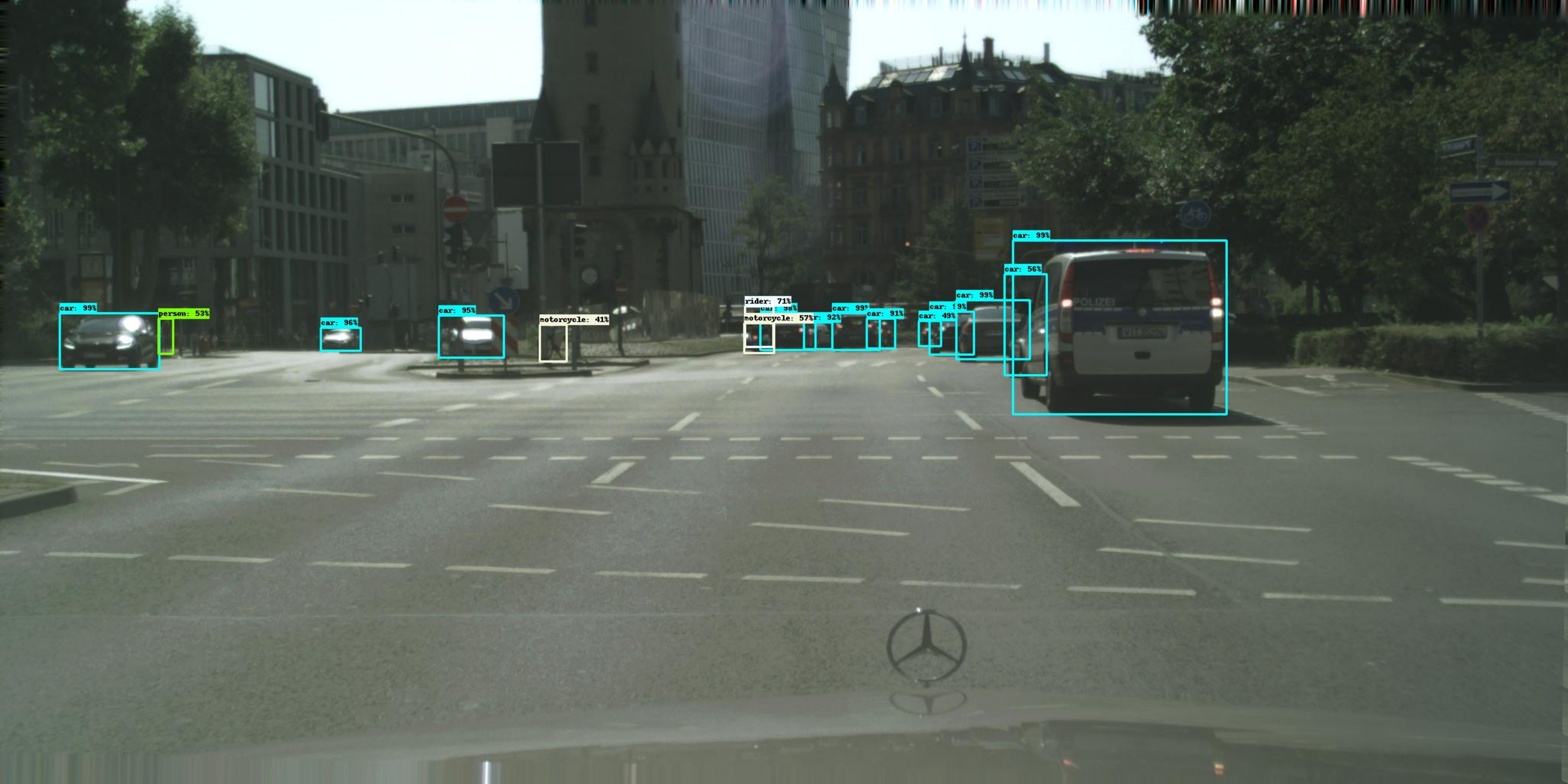}
                \caption*{PseudoProp}
        \end{subfigure}%
        \rulesep
        \begin{subfigure}[b]{0.49\textwidth}
                \includegraphics[width=\linewidth]{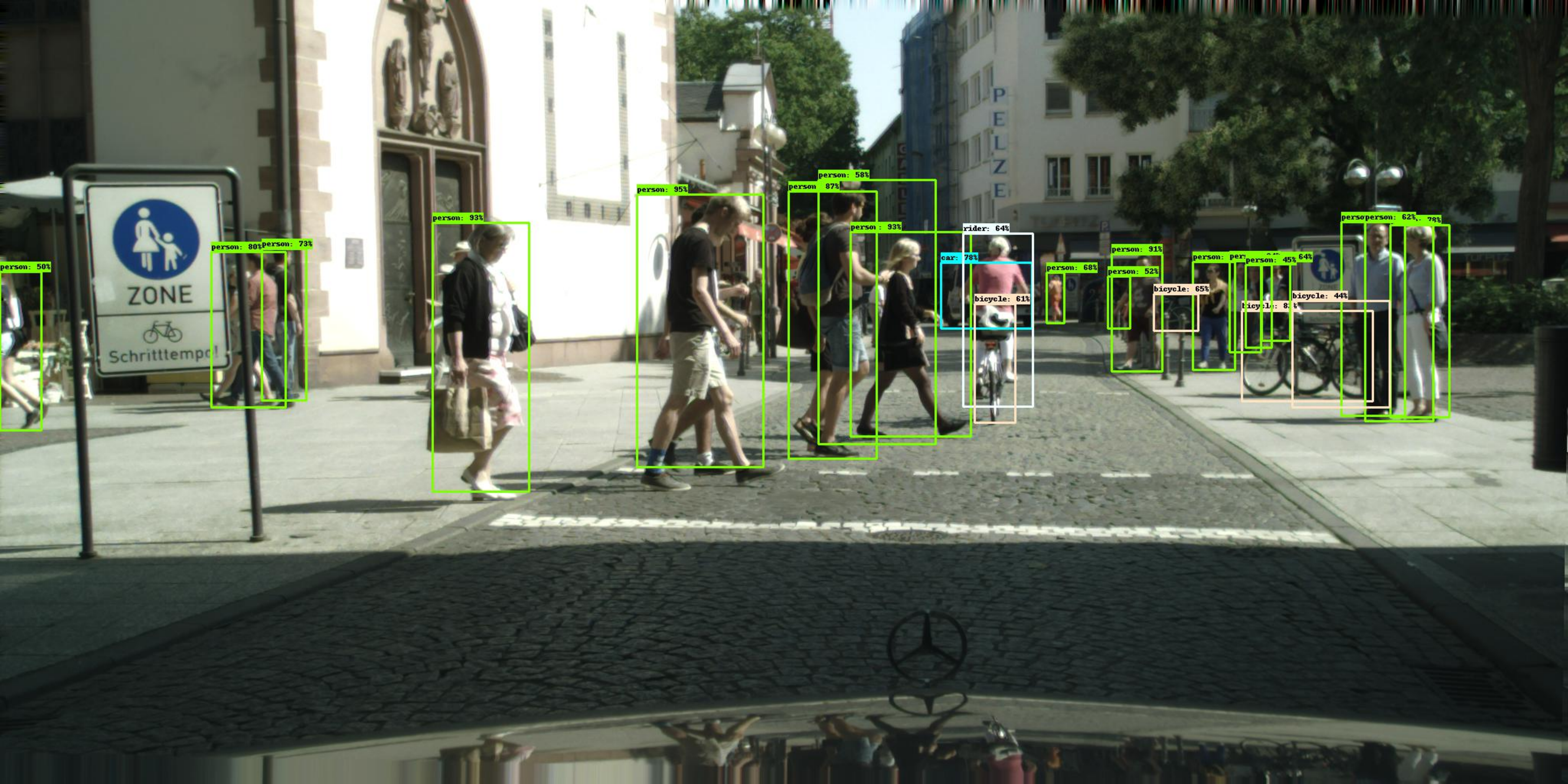}
                \caption*{PseudoProp}
        \end{subfigure}%
\caption{\small \it Visual comparison for the ground truth, Naive-Student, VideoProp, and our proposed PseudoProp on Cityscapes.}\label{fig: image_results4}
\end{figure*}

We compare the visual results in Figure \ref{fig: image_results1}, \ref{fig: image_results2}, \ref{fig: image_results3}, and \ref{fig: image_results4}, for the ground truth, Naive-Student, VideoProp, and our proposed PseudoProp respectively on the Cityscapes validation dataset. From these figures, we can see that our PseudoProp model can eliminate miss and false detections. This means the pseudo-labels generated by our model are  more robust.   

\clearpage

\clearpage

\clearpage

\clearpage




\end{document}